\documentclass[a4paper, 10pt, conference]{ieeeconf}      
\usepackage{FG2026}
\usepackage{booktabs}
\usepackage{hhline}
\usepackage{hyperref}
\usepackage{url}
\usepackage{multirow}
\usepackage{arydshln}
\usepackage{graphicx}
\usepackage{subcaption}
\usepackage{tikz}
\usepackage{amssymb} 
\usepackage{amsmath}
\newcommand*\circled[1]{\tikz[baseline=(char.base)]{
            \node[shape=circle,draw,inner sep=1pt] (char) {#1};}}

\FGfinalcopy 

\IEEEoverridecommandlockouts                              
\newcommand{\rev}[1]{\textcolor{black}{#1}}

\def\FGPaperID{150} 

\title{\LARGE \bf
Vision Transformers for Face Recognition Need More Registers
}

\author{\parbox{16cm}{\centering
    {\normalsize Tahar Chettaoui$^{1,2}$,Guray Ozgur$^{1,2}$, Eduarda Caldeira$^{1,2}$,  Naser Damer$^{1,2}$, and  Fadi Boutros$^{1}$}\\
    {\normalsize
    $^1$ Fraunhofer Institute for Computer Graphics Research IGD, Germany, $^2$ Department of Computer Science, TU Darmstadt, Germany}}
    \thanks{\scriptsize This research work has been funded by the German Federal Ministry of Education and Research and the Hessen State Ministry for Higher Education, Research and the Arts within their joint support of the National Research Center for Applied Cybersecurity ATHENE.}
}

\usepackage{fancyhdr}

\begin{document}

\ifFGfinal
\thispagestyle{empty}
\pagestyle{empty}
\else
\author{Anonymous FG2026 submission\\ Paper ID \FGPaperID \\}
\pagestyle{plain}
\fi
\maketitle
\thispagestyle{fancy}

\begin{abstract}
Recent advances in Vision Transformers (ViTs) for face recognition (FR) have moved beyond the standard \texttt{CLS}-token paradigm. In this paradigm, a special classification token (CLS) is prepended to the patch embeddings and used as a representation of the input for downstream tasks. An alternative approach, Concatenated Patch Embeddings (CPE), instead leverages all patch tokens by concatenating them into a single vector, which is then projected into a compact face representation. CPE has been shown to improve recognition performance in comparison to \texttt{CLS}-based ones, but our qualitative analysis of attention maps showed the presence of artifacts that limit their interpretability. To address this issue, we incorporate register tokens, learnable tokens concatenated to the initial patch embeddings, and processed jointly through the ViT encoder blocks. This mechanism \rev{has been shown} to produce more structured and interpretable attention maps compared to baseline ViT. We empirically demonstrate that these artifacts consistently appear across various ViT backbones, including small and large models, and that introducing register tokens effectively mitigates them. Adding four or eight registers significantly enhances interpretability, with eight registers providing the highest verification accuracies and \rev{smoothest} attention structures. Our resulting model, ViT-8R, corresponds to a CPE-based ViT-B architecture augmented with eight register tokens achieves state-of-the-art performance among ViT-based FR models on large-scale IJB-B and IJB-C benchmarks. Also, ViT-8R produces substantially clearer attention maps compared with the baseline model, which offer deeper insight into the model’s attention behavior (\url{https://github.com/TaharChettaoui/ViT-FR-Registers}). 


\end{abstract}

\vspace{-1mm}
\section{Introduction}
Vision Transformers (ViT) \cite{dosovitskiy2021imageworth16x16words} have recently gained popularity in computer vision, achieving competitive performance to traditional convolutional neural networks (CNNs) \cite{DBLP:conf/cvpr/HeZRS16} which also motivated several works \cite{DBLP:conf/bmvc/SunT22, DBLP:conf/isie/KhanSEG23, DBLP:conf/cvpr/KimS0JL24, DBLP:conf/iccv/DanLXD0XS23, DBLP:journals/tcsv/QinWDWCHD24, DBLP:journals/corr/abs-2209-08930, DBLP:journals/corr/abs-2103-14803,DBLP:journals/corr/abs-2601-05741, DBLP:journals/ivc/ChettaouiDB25, DBLP:conf/bmvc/IslamZM22} to explore their use in downstream task of face recognition (FR).  
State-of-the-art (SoTA) FRs in the literature are trained to implicitly optimize feature representations needed for the verification task by learning a multi-class classification problem with a margin-penalty softmax loss or its variants \cite{wang2018cosfacelargemargincosine, Deng_2022, DBLP:conf/cvpr/Kim0L22} and then use the embedding layer as feature representation. 
To achieve such a goal, the common architectures of ViT for learning a multi-class classification problem utilize  
a dedicated class token (CLS) \cite{dosovitskiy2021imageworth16x16words, DBLP:journals/tmlr/OquabDMVSKFHMEA24, DBLP:conf/iccv/ChenFP21}, whose final embedding serves as a global image representation. 
Early works of ViT for FR \cite{DBLP:conf/bmvc/SunT22, DBLP:conf/isie/KhanSEG23, DBLP:journals/ivc/ChettaouiDB25} rely on the \texttt{CLS} token to learn a compact global descriptor, while recent ones \cite{DBLP:conf/cvpr/KimS0JL24, DBLP:conf/iccv/DanLXD0XS23} aggregate all patch embeddings, known as Concatenated Patch Embeddings (CPE), to preserve localized cues for discrimination. 
CPE-based ViTs \cite{DBLP:conf/cvpr/KimS0JL24, DBLP:conf/iccv/DanLXD0XS23} for FR have achieved superior performances, in comparison to CNNs and CLS-based ViT architectures. However, our qualitative analysis of CPE-based ViT attention maps, as reported  later in this paper in Section \ref{sec:results}, reveals artifacts that limit their interpretability. These artifacts, identified by Darcet et al. \cite{DBLP:conf/iclr/DarcetOMB24}, tend to appear in areas corresponding to low-information background regions, suggesting that the model repurposes them for internal computations. To mitigate these artifacts, Darcet et al. \cite{DBLP:conf/iclr/DarcetOMB24}  introduced additional learnable tokens, namely register tokens, appended to the embedding sequence after the patch projection layer, which the model can learn to utilize as registers. While prior work \cite{DBLP:conf/iclr/DarcetOMB24} has established effective techniques to address such artifacts in general image classification, these approaches have not been specifically examined in the context of FR, especially when considering common CPE-based ViTs architectures for FR. This motivates our study, in which we investigate the integration of register tokens into CPE-based ViT architectures for FR, aiming to enhance interpretability and provide deeper insights into the model’s attention mechanisms.

\begin{figure}[!tp]
  \centering

\begin{minipage}{0.9\linewidth}
  \centering
    \begin{subfigure}[b]{0.15\linewidth}
    \caption*{Input}
      \includegraphics[width=\linewidth, height=1.2cm]{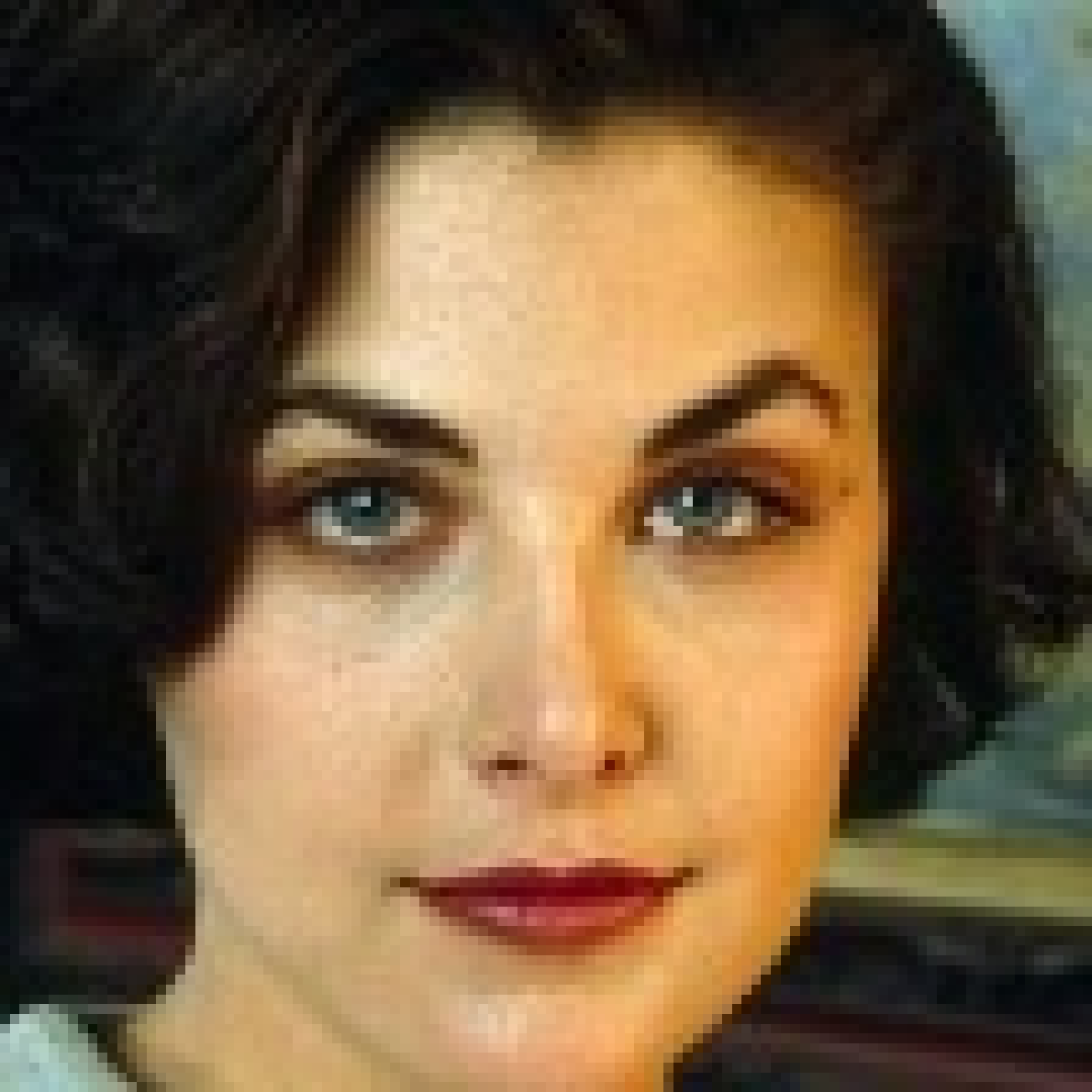}
  \end{subfigure}
  \hfill
  \begin{subfigure}[b]{0.15\linewidth}
      \caption*{0R}
      \includegraphics[width=\linewidth, height=1.2cm]{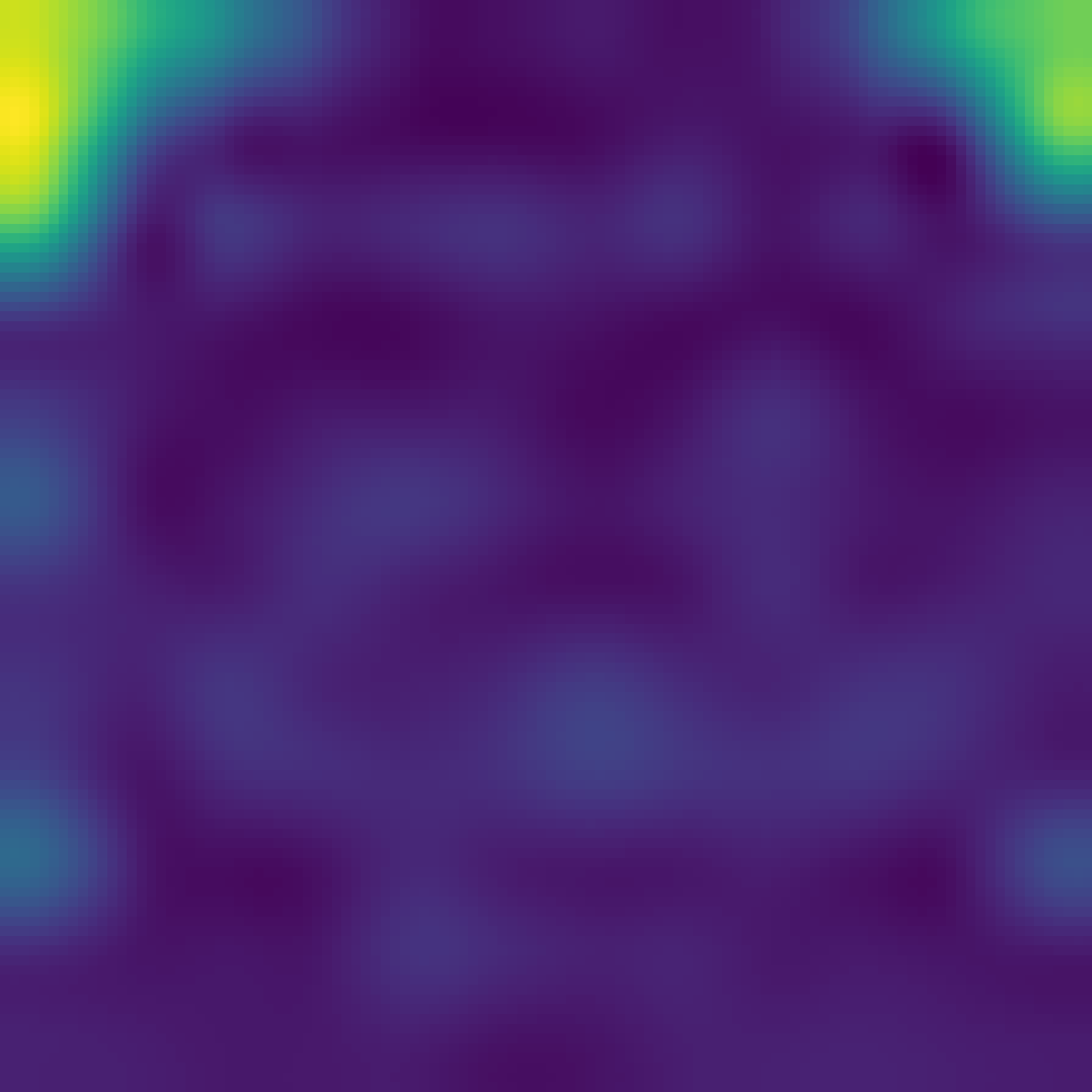}
  \end{subfigure}
  \hfill
    \begin{subfigure}[b]{0.15\linewidth}
    \caption*{1R}
      \includegraphics[width=\linewidth, height=1.2cm]{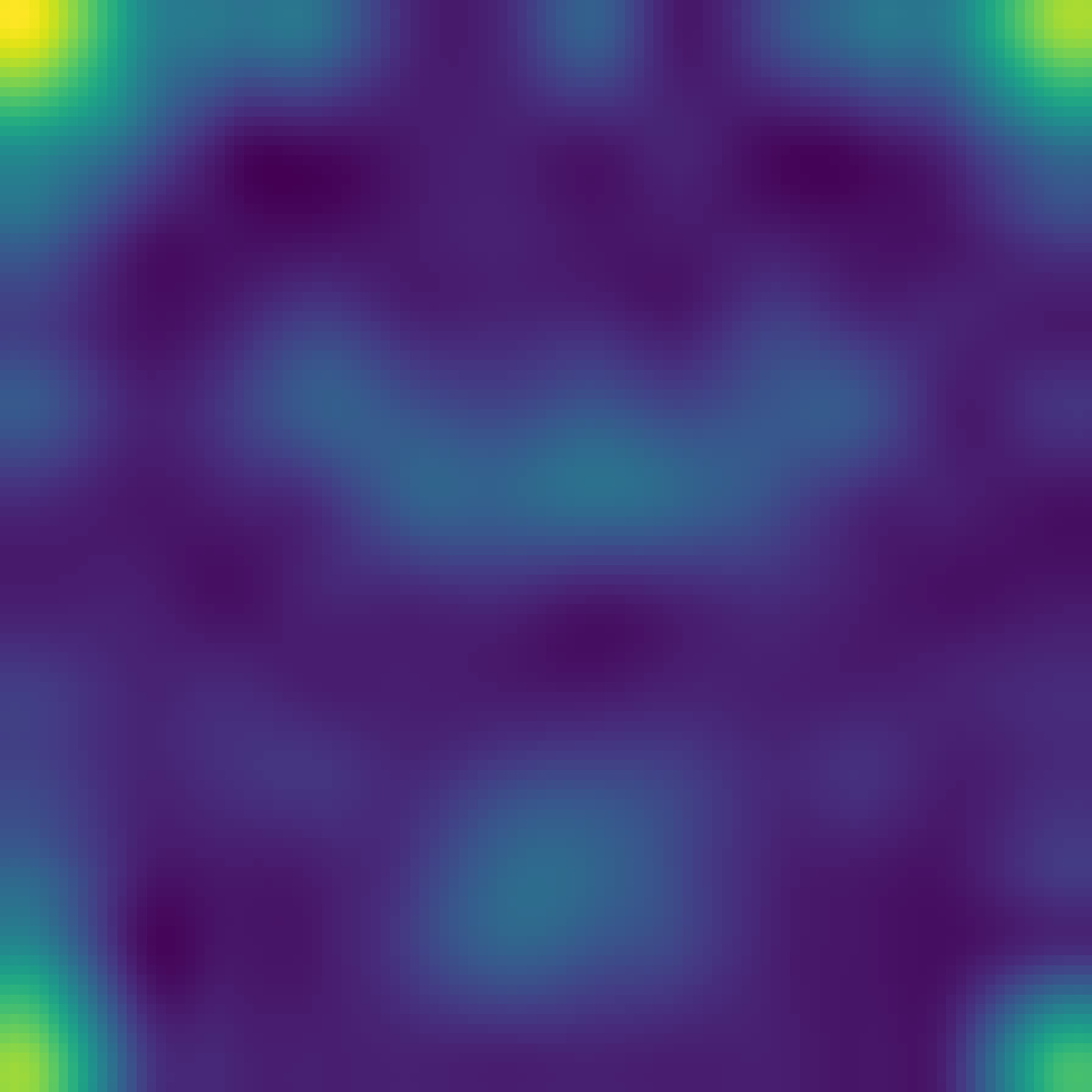}
  \end{subfigure}
  \hfill
    \begin{subfigure}[b]{0.15\linewidth}
          \caption*{2R}
      \includegraphics[width=\linewidth, height=1.2cm]{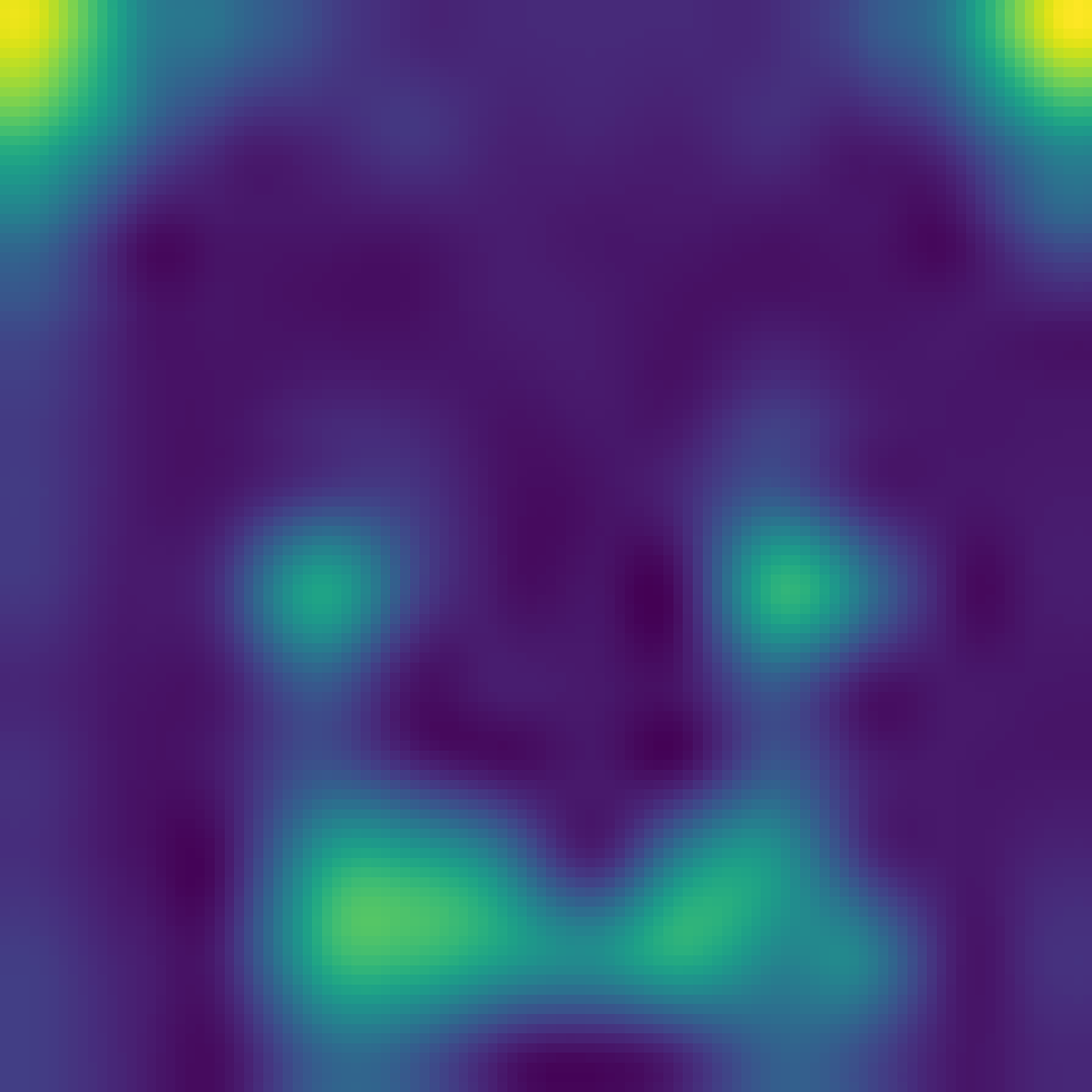}
  \end{subfigure}
  \hfill
  \begin{subfigure}[b]{0.15\linewidth}
  \caption*{4R}
      \includegraphics[width=\linewidth, height=1.2cm]{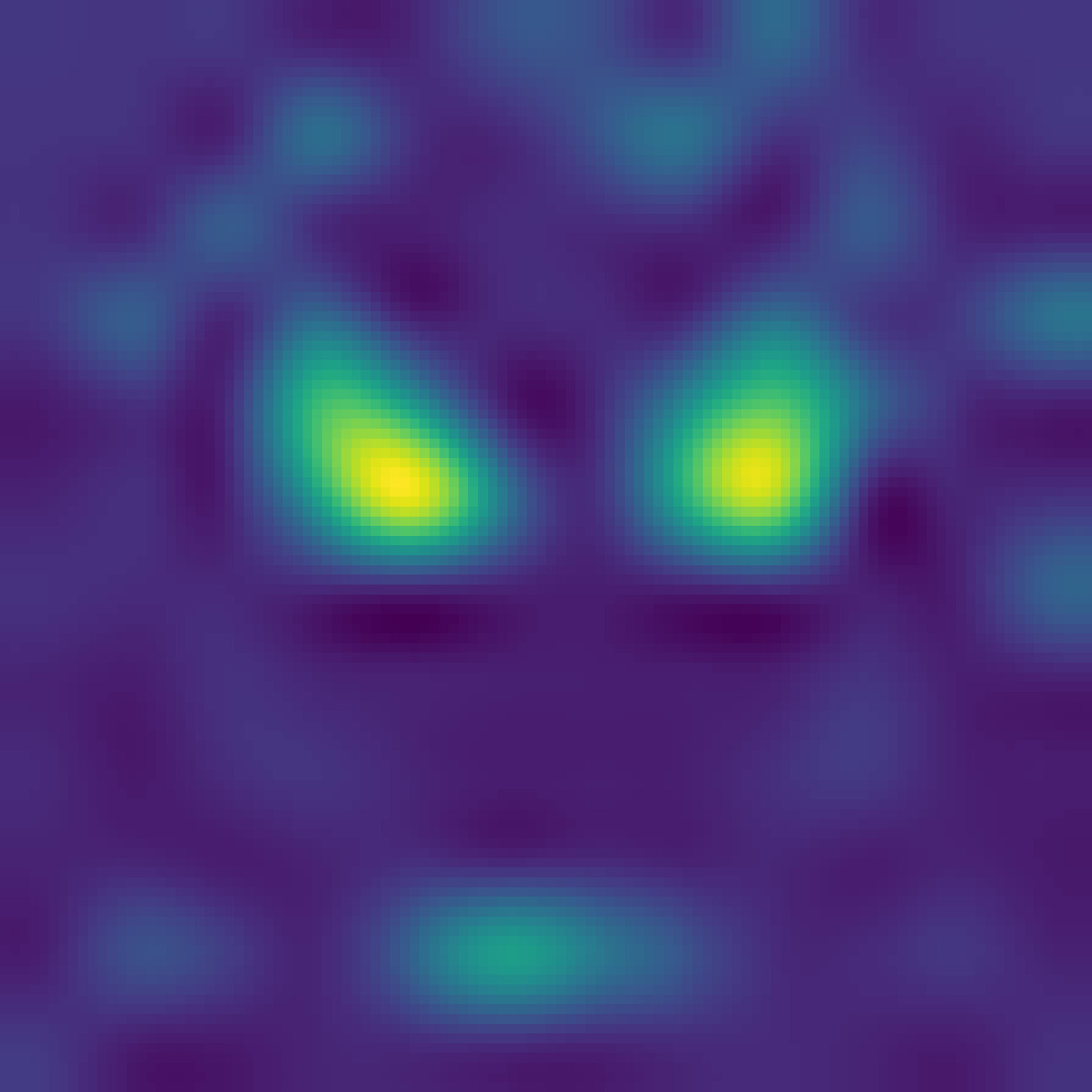}
  \end{subfigure}
  \hfill
    \begin{subfigure}[b]{0.15\linewidth}
    \caption*{8R}
      \includegraphics[width=\linewidth, height=1.2cm]{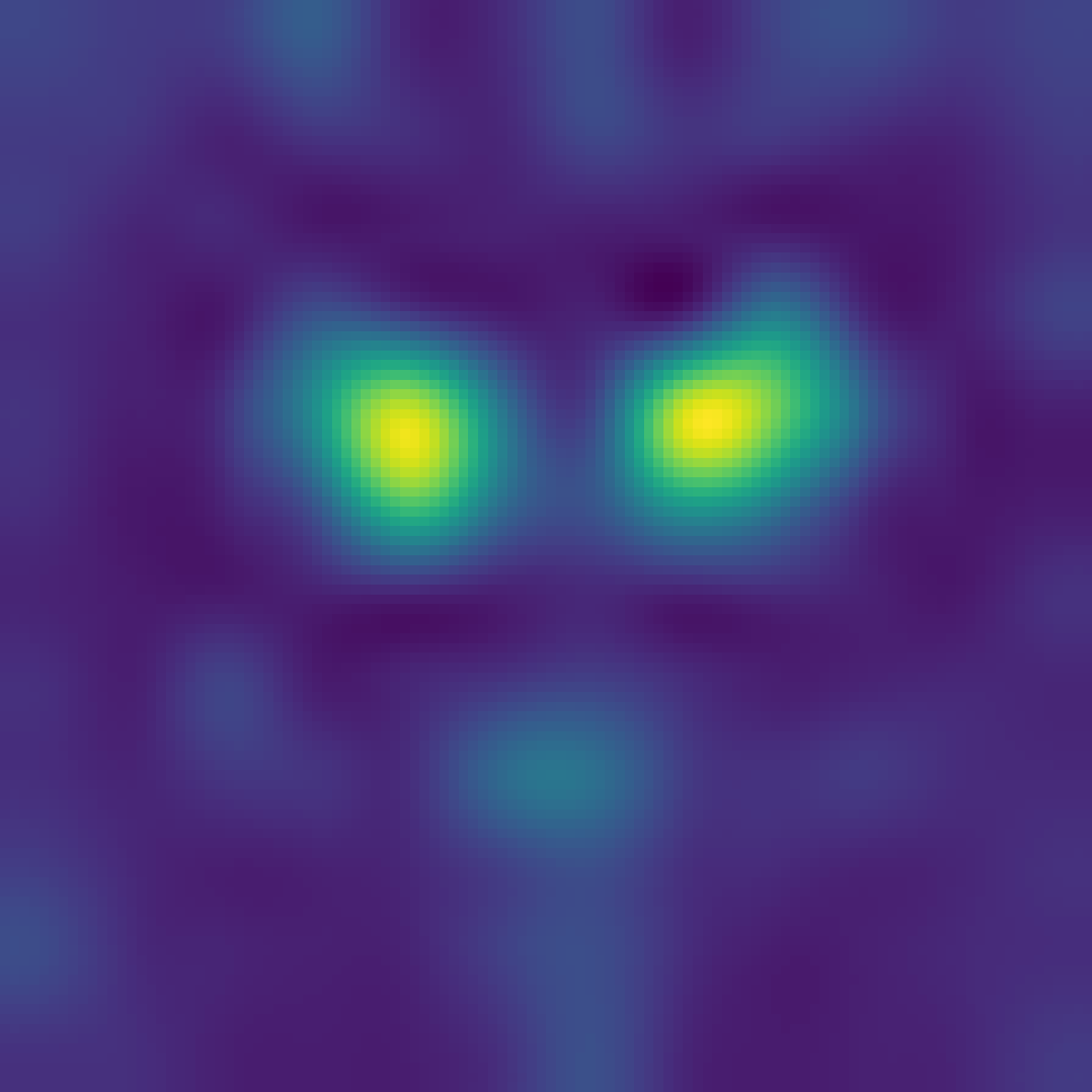}
  \end{subfigure}
\end{minipage}
  
  \vspace{0.1cm} 

\begin{minipage}{0.9\linewidth}
  \centering
    \begin{subfigure}[b]{0.15\linewidth}
      \includegraphics[width=\linewidth, height=1.2cm]{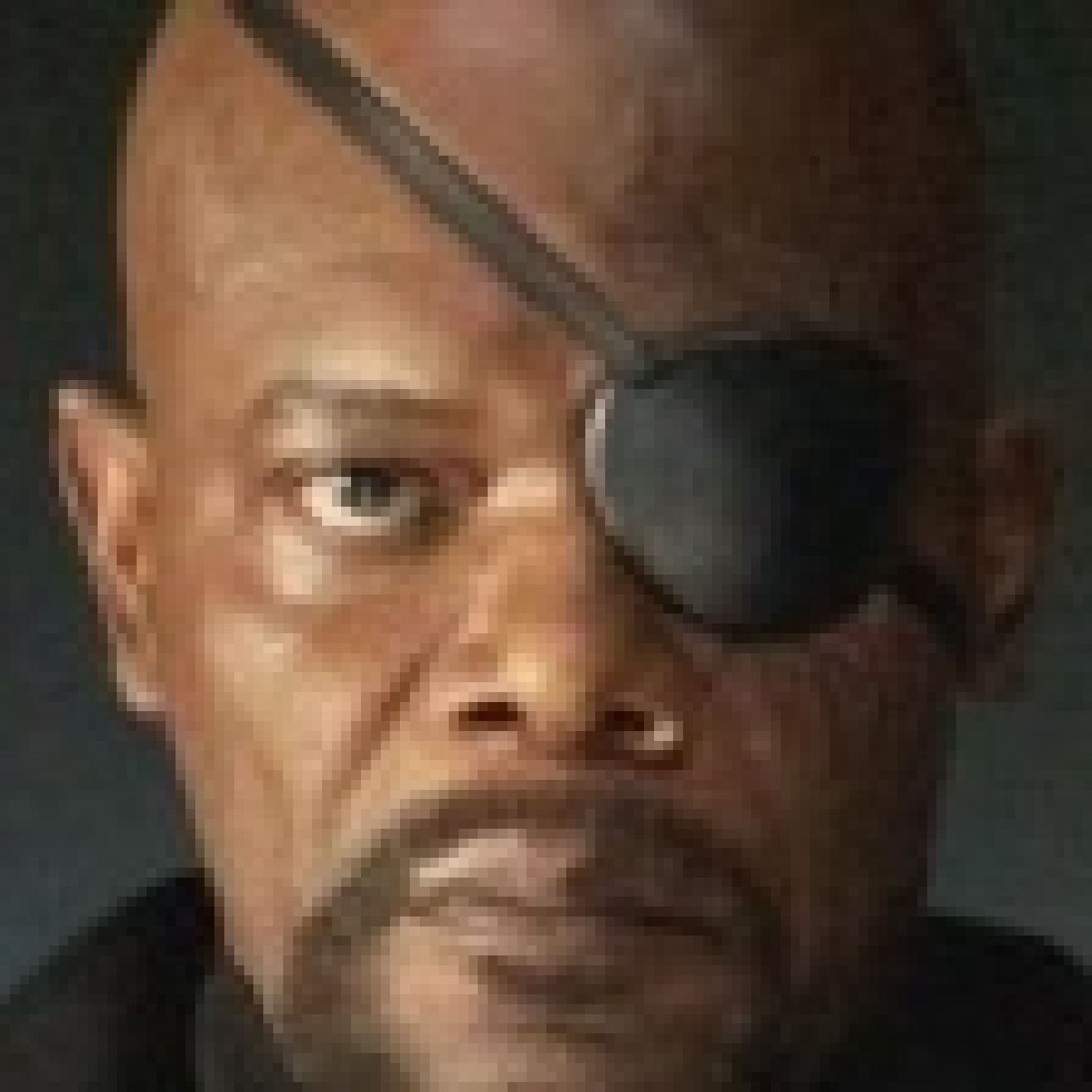}
  \end{subfigure}
  \hfill
  \begin{subfigure}[b]{0.15\linewidth}
      \includegraphics[width=\linewidth, height=1.2cm]{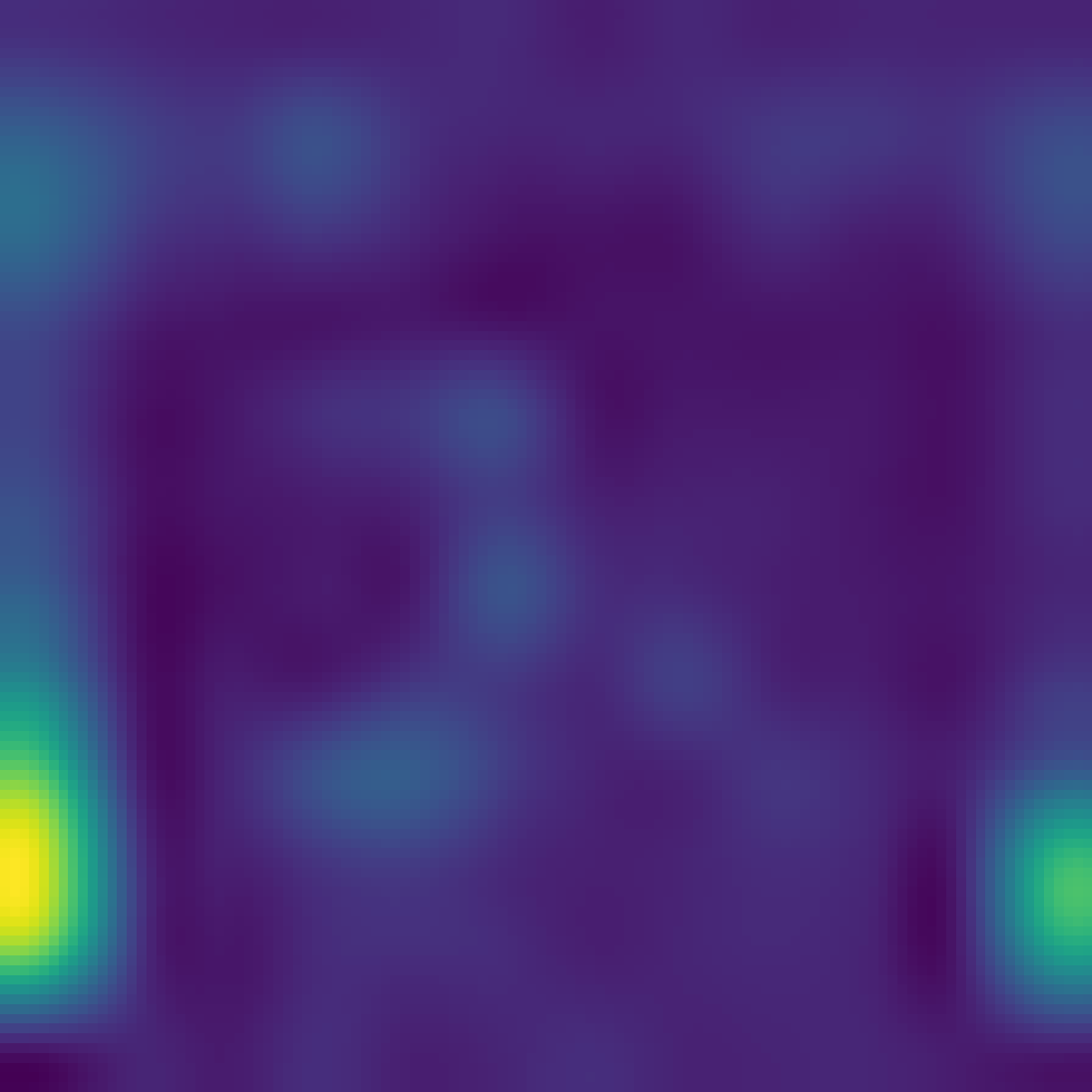}
  \end{subfigure}
  \hfill
    \begin{subfigure}[b]{0.15\linewidth}
      \includegraphics[width=\linewidth, height=1.2cm]{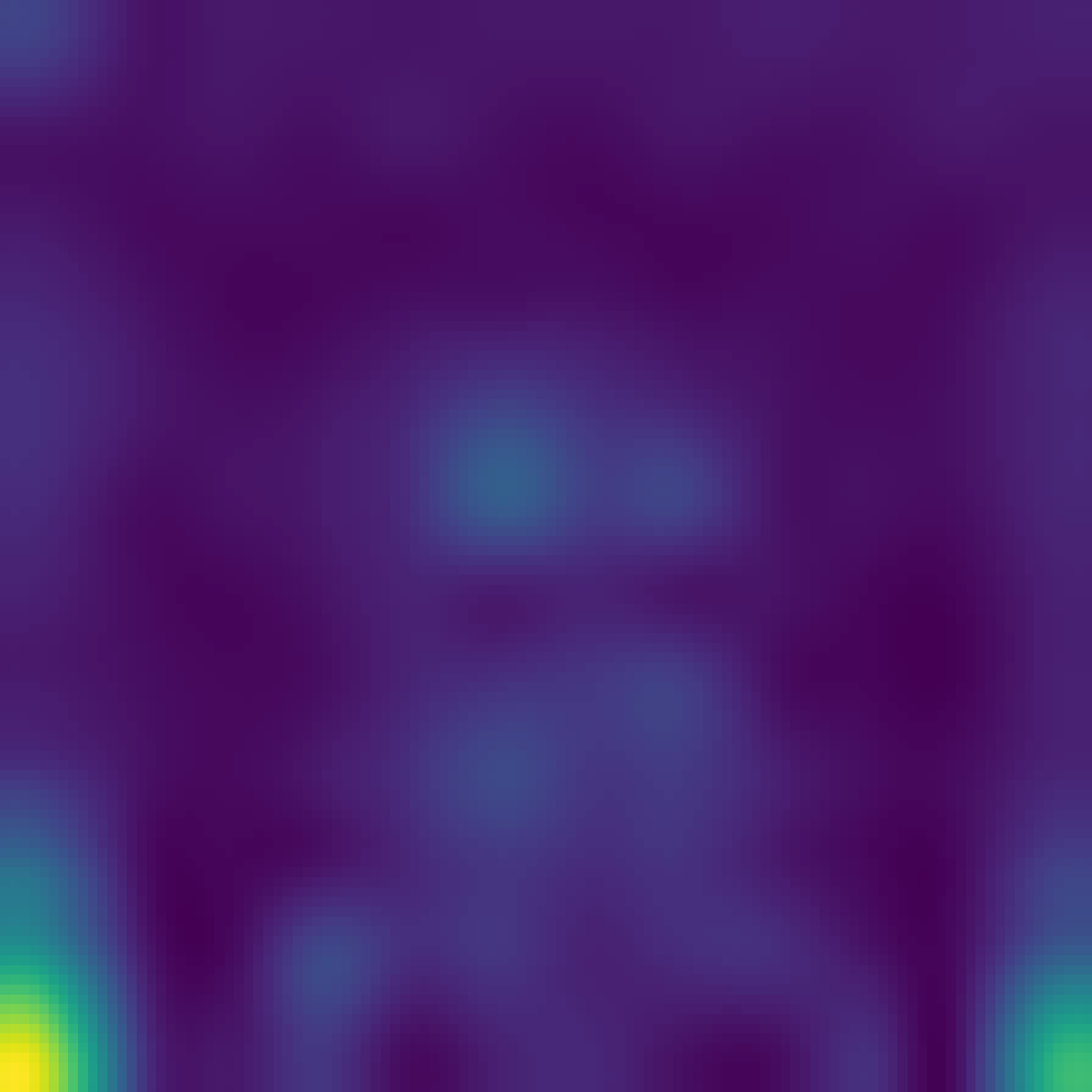}

  \end{subfigure}
  \hfill
    \begin{subfigure}[b]{0.15\linewidth}
      \includegraphics[width=\linewidth, height=1.2cm]{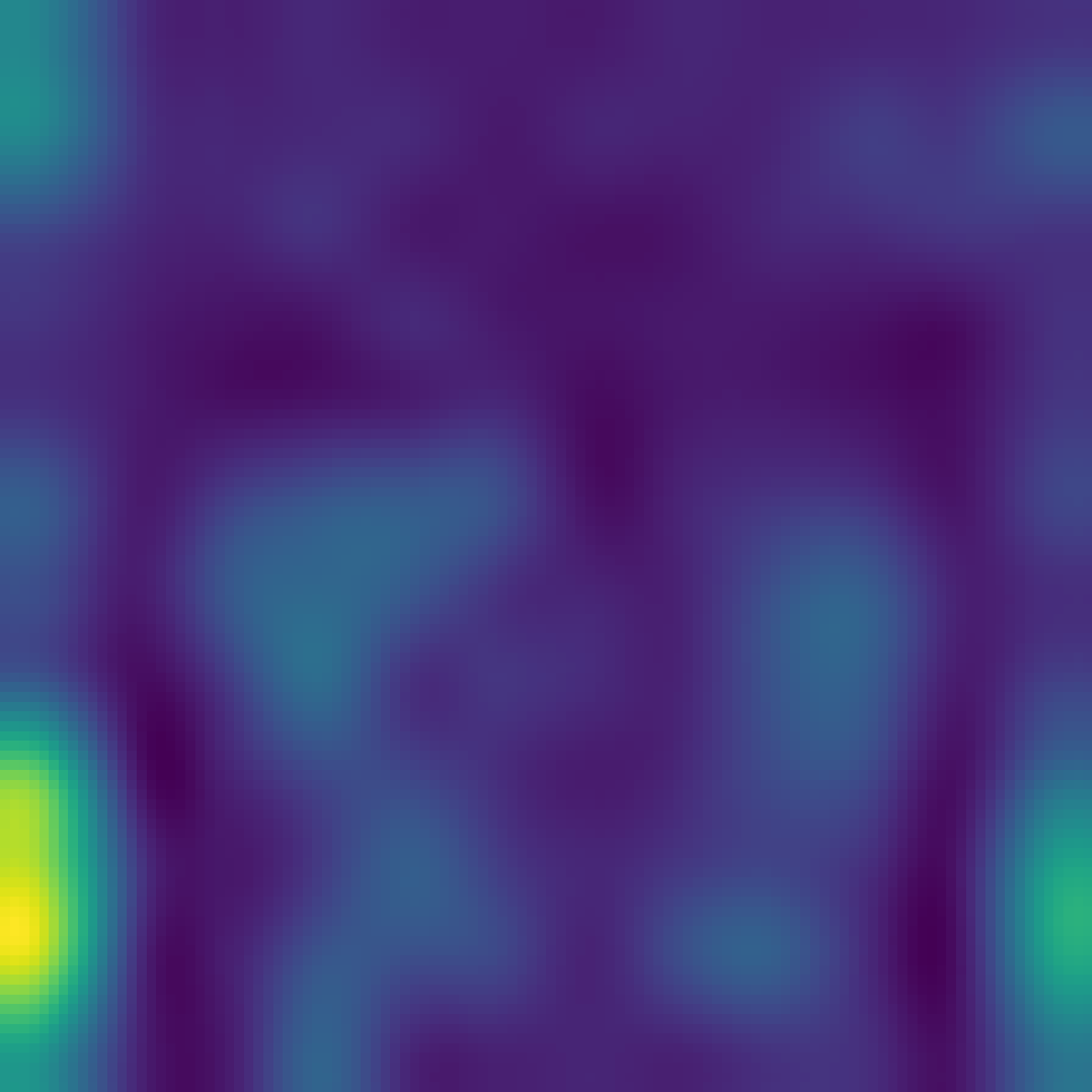}

  \end{subfigure}
  \hfill
  \begin{subfigure}[b]{0.15\linewidth}
      \includegraphics[width=\linewidth, height=1.2cm]{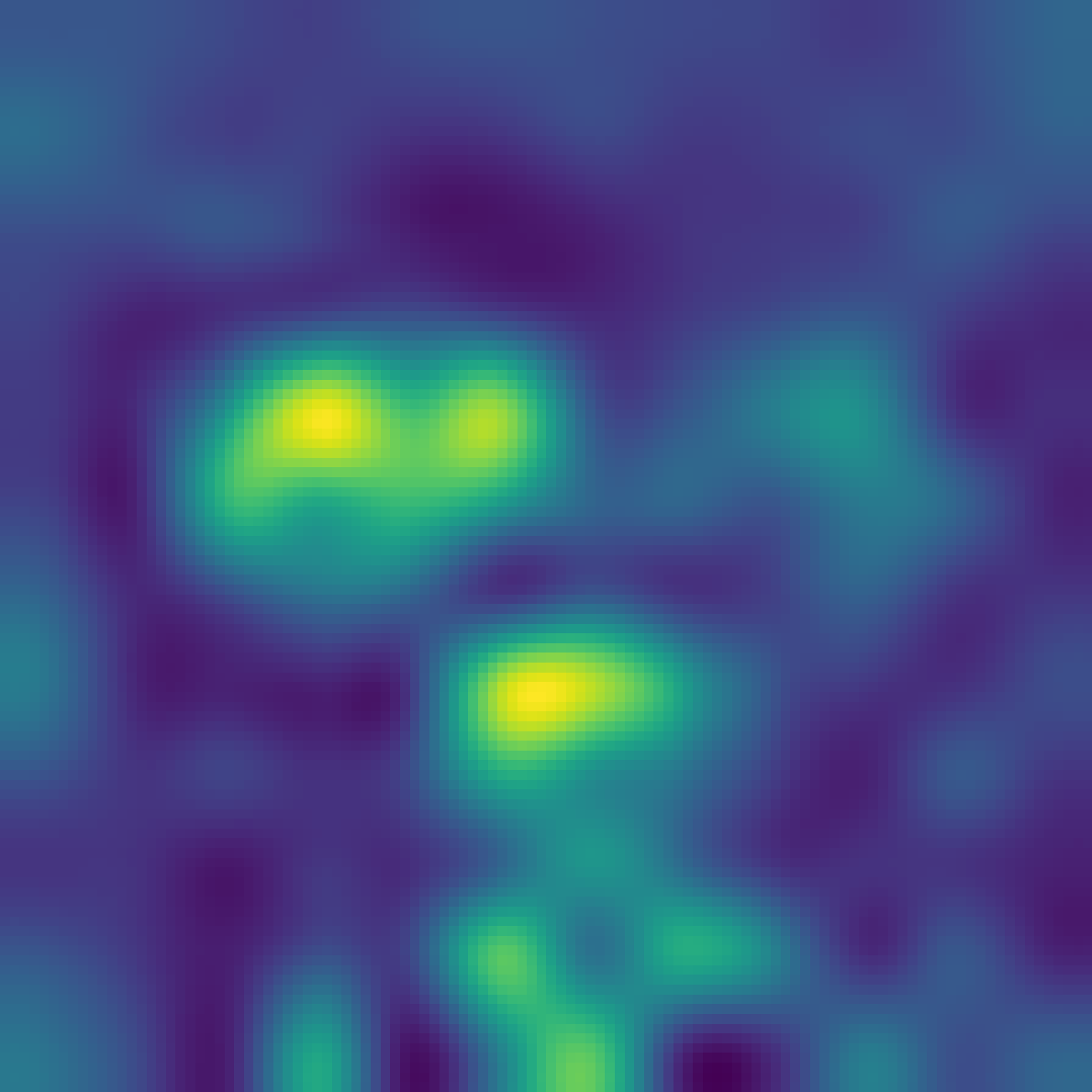}

  \end{subfigure}
  \hfill
    \begin{subfigure}[b]{0.15\linewidth}
      \includegraphics[width=\linewidth, height=1.2cm]{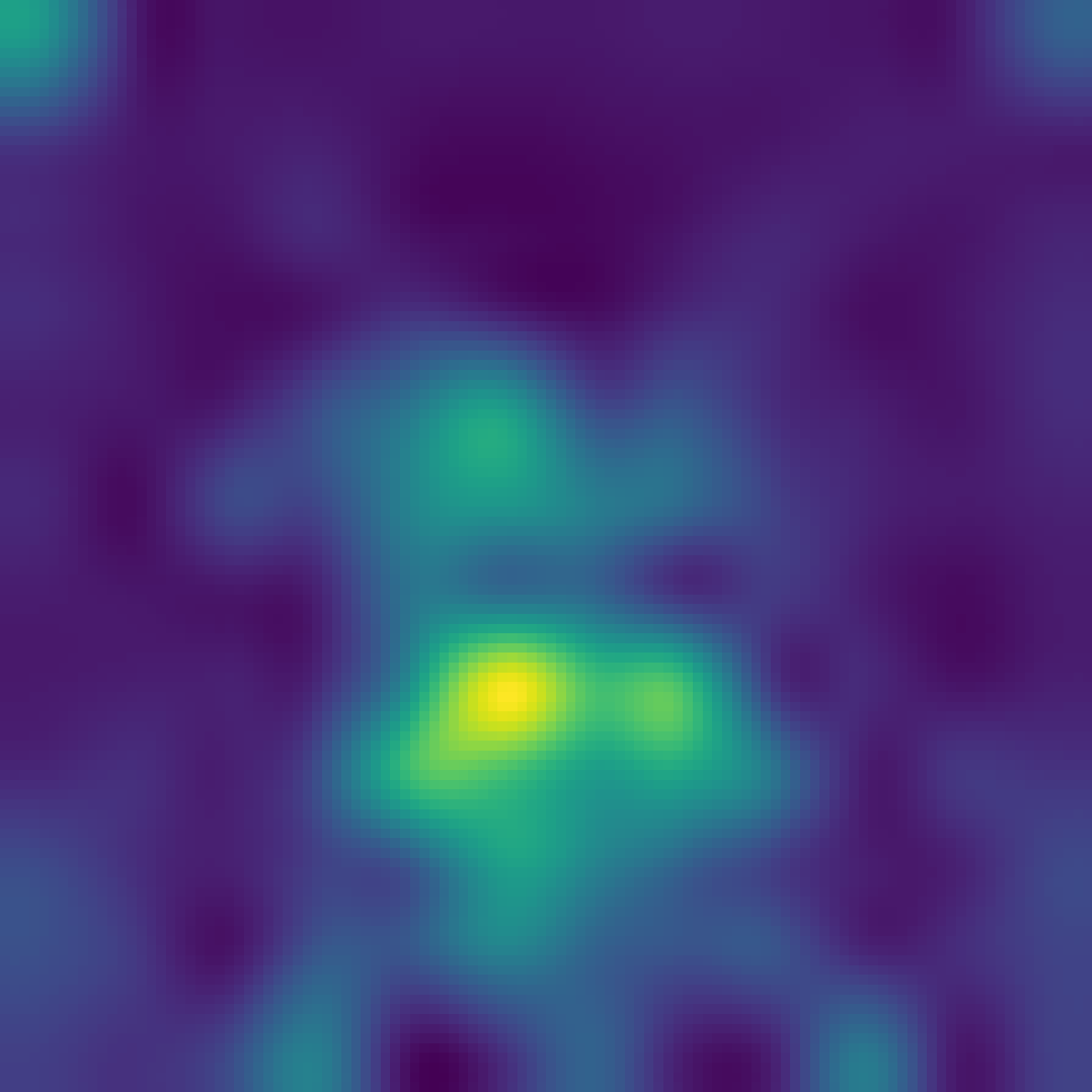}

  \end{subfigure}
\end{minipage}

  \vspace{0.1cm} 

\begin{minipage}{0.9\linewidth}
  \centering
    \begin{subfigure}[b]{0.15\linewidth}
      \includegraphics[width=\linewidth, height=1.2cm]{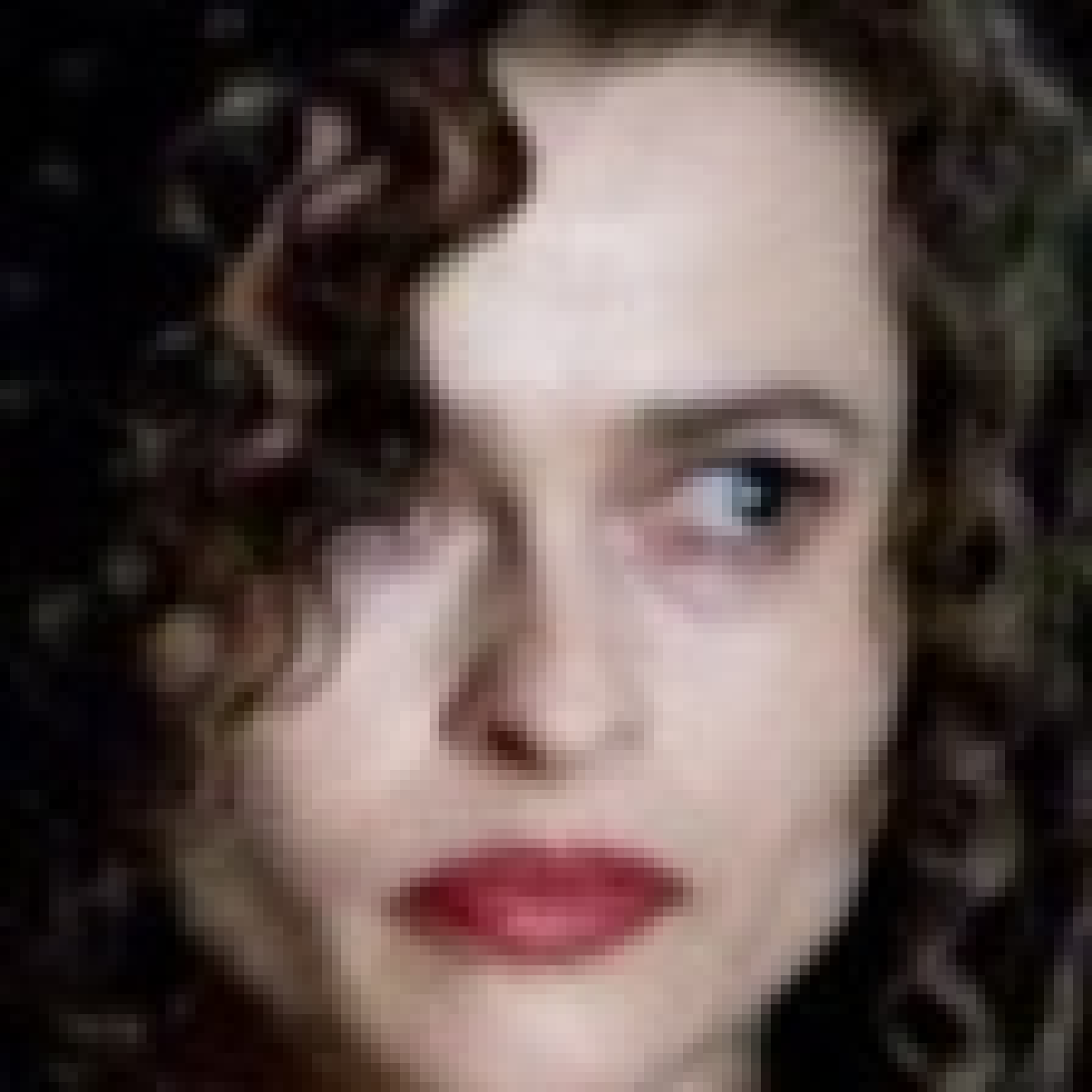}

  \end{subfigure}
  \hfill
  \begin{subfigure}[b]{0.15\linewidth}
      \includegraphics[width=\linewidth, height=1.2cm]{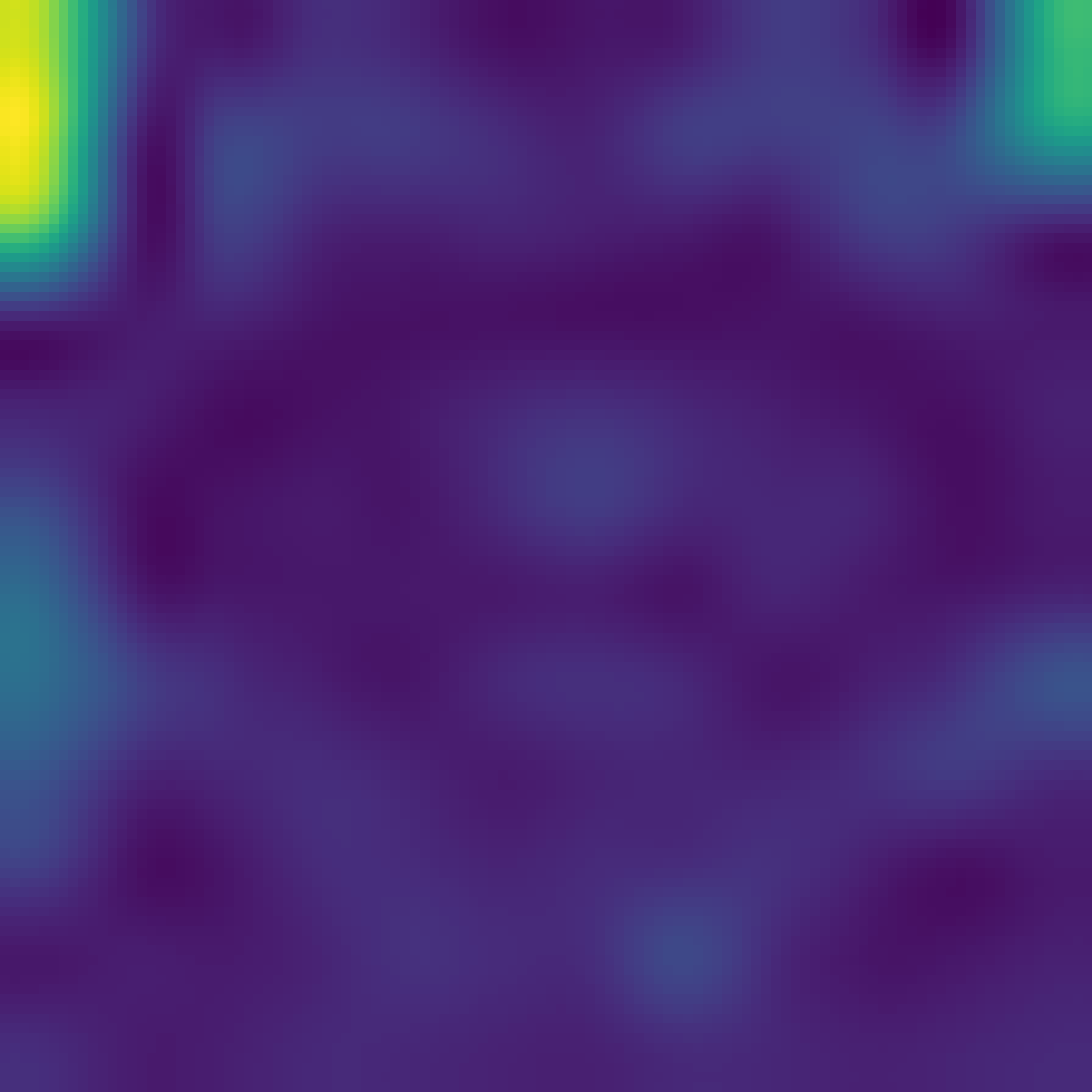}

  \end{subfigure}
  \hfill
    \begin{subfigure}[b]{0.15\linewidth}
      \includegraphics[width=\linewidth, height=1.2cm]{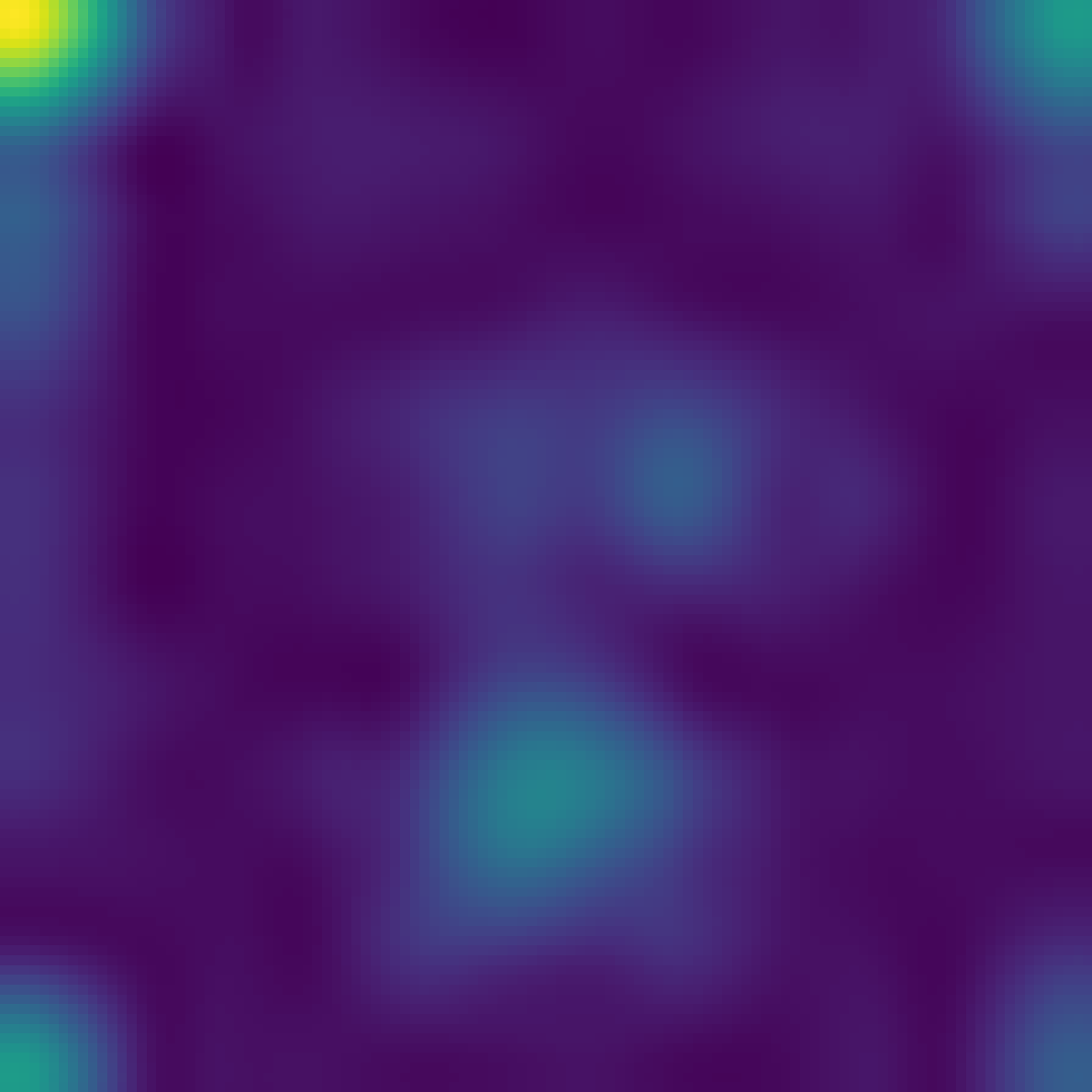}

  \end{subfigure}
  \hfill
    \begin{subfigure}[b]{0.15\linewidth}
      \includegraphics[width=\linewidth, height=1.2cm]{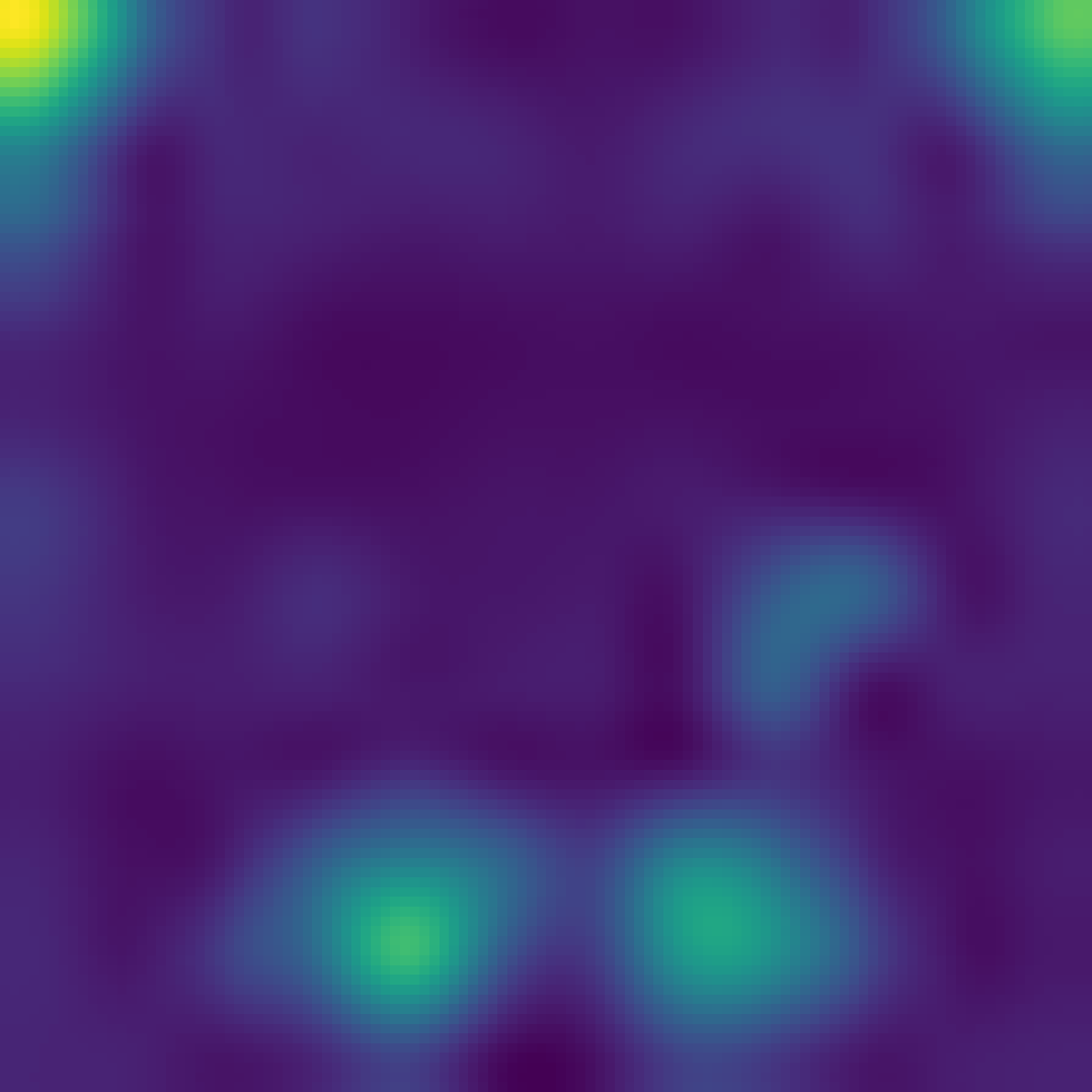}

  \end{subfigure}
  \hfill
  \begin{subfigure}[b]{0.15\linewidth}
      \includegraphics[width=\linewidth, height=1.2cm]{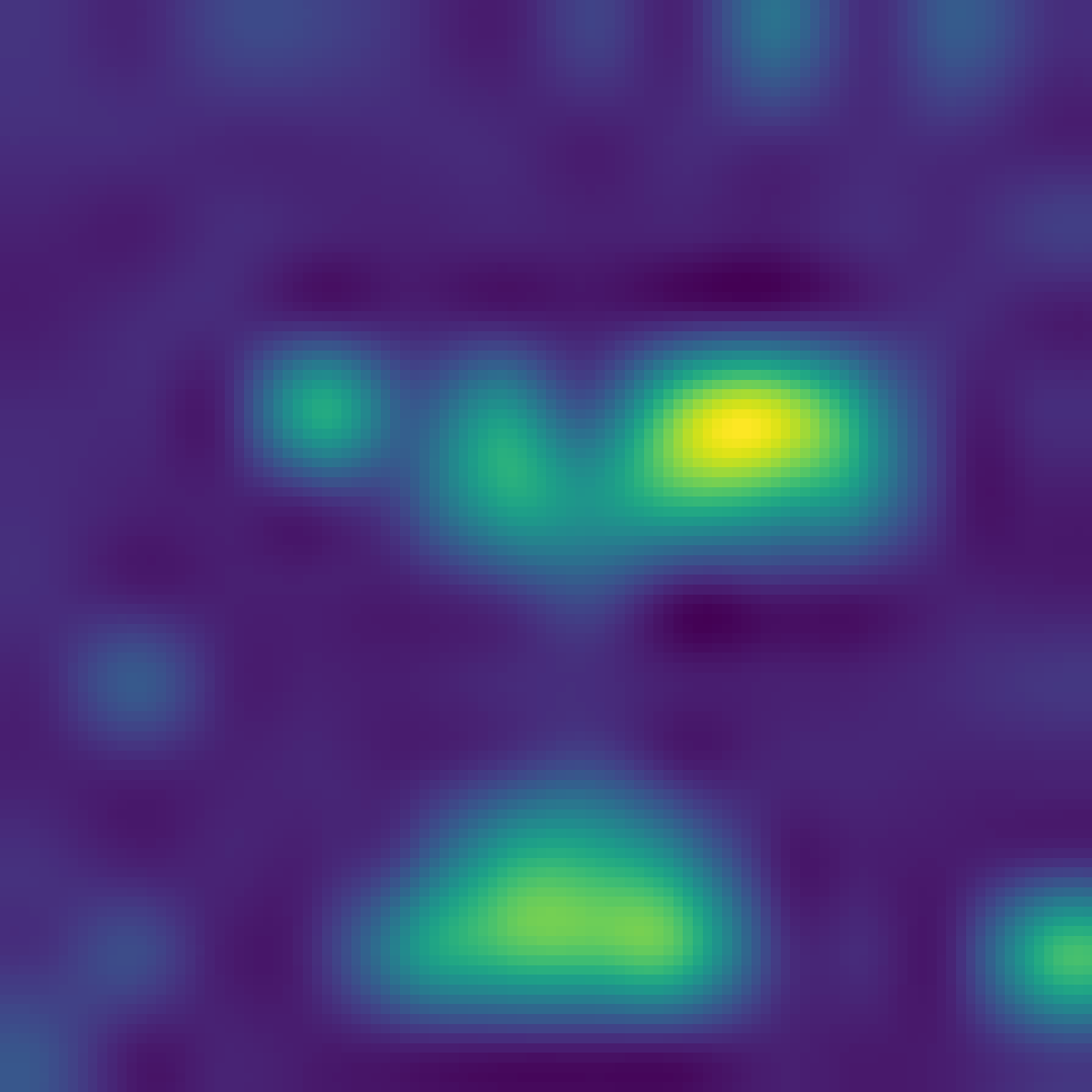}

  \end{subfigure}
  \hfill
    \begin{subfigure}[b]{0.15\linewidth}
      \includegraphics[width=\linewidth, height=1.2cm]{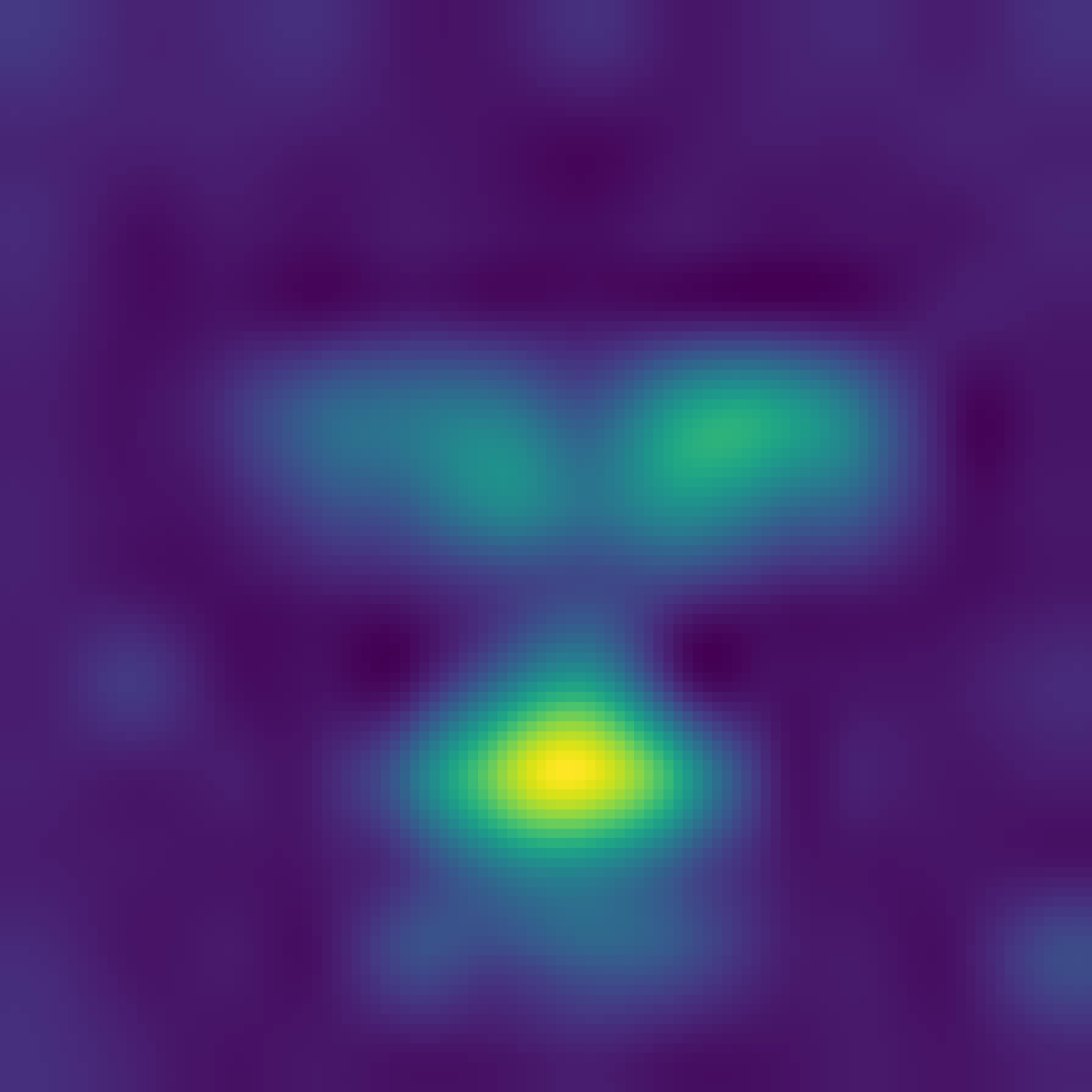}

  \end{subfigure}
\end{minipage}

  \vspace{0.1cm} 

\begin{minipage}{0.9\linewidth}
  \centering
    \begin{subfigure}[b]{0.15\linewidth}
      \includegraphics[width=\linewidth, height=1.2cm]{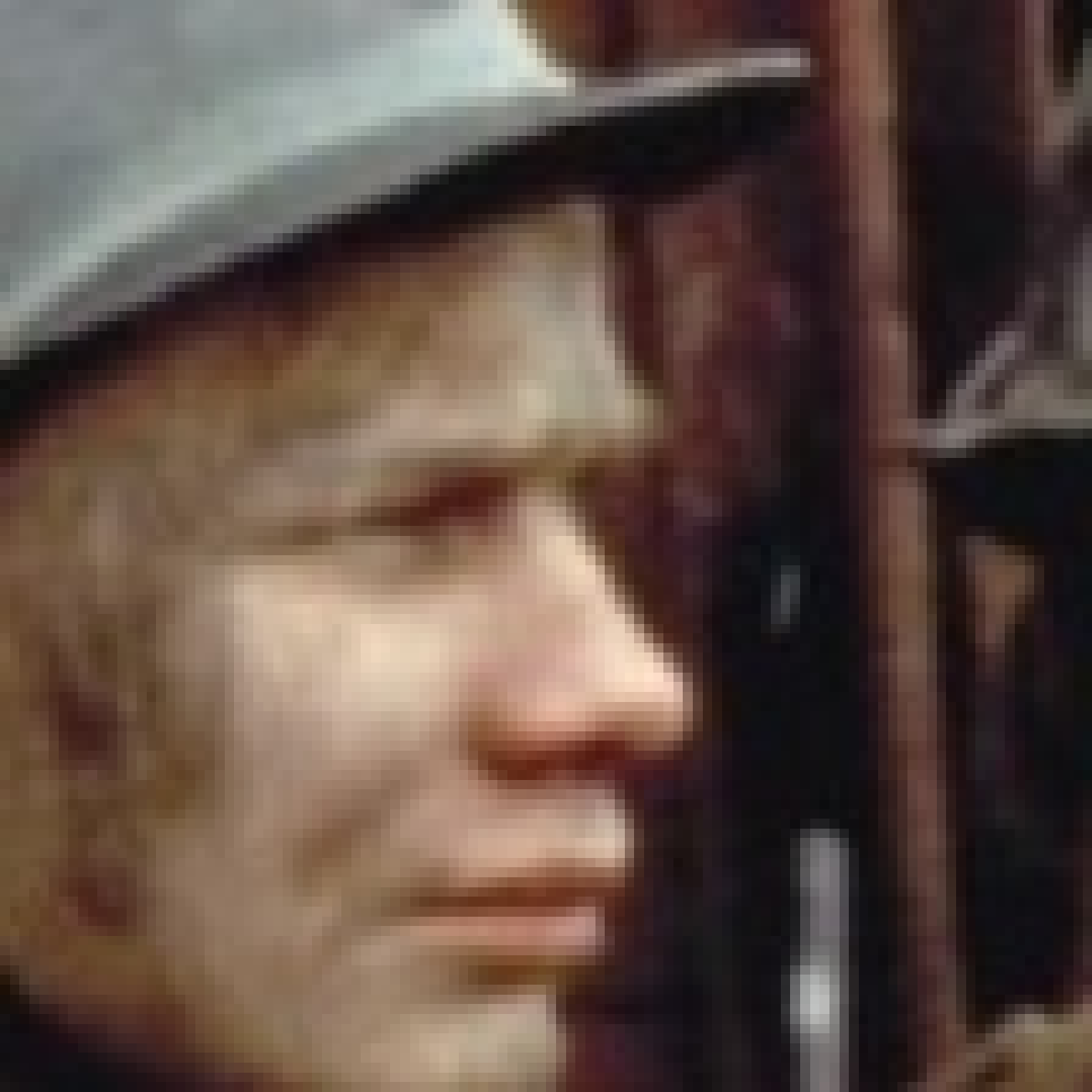}

  \end{subfigure}
  \hfill
  \begin{subfigure}[b]{0.15\linewidth}
      \includegraphics[width=\linewidth, height=1.2cm]{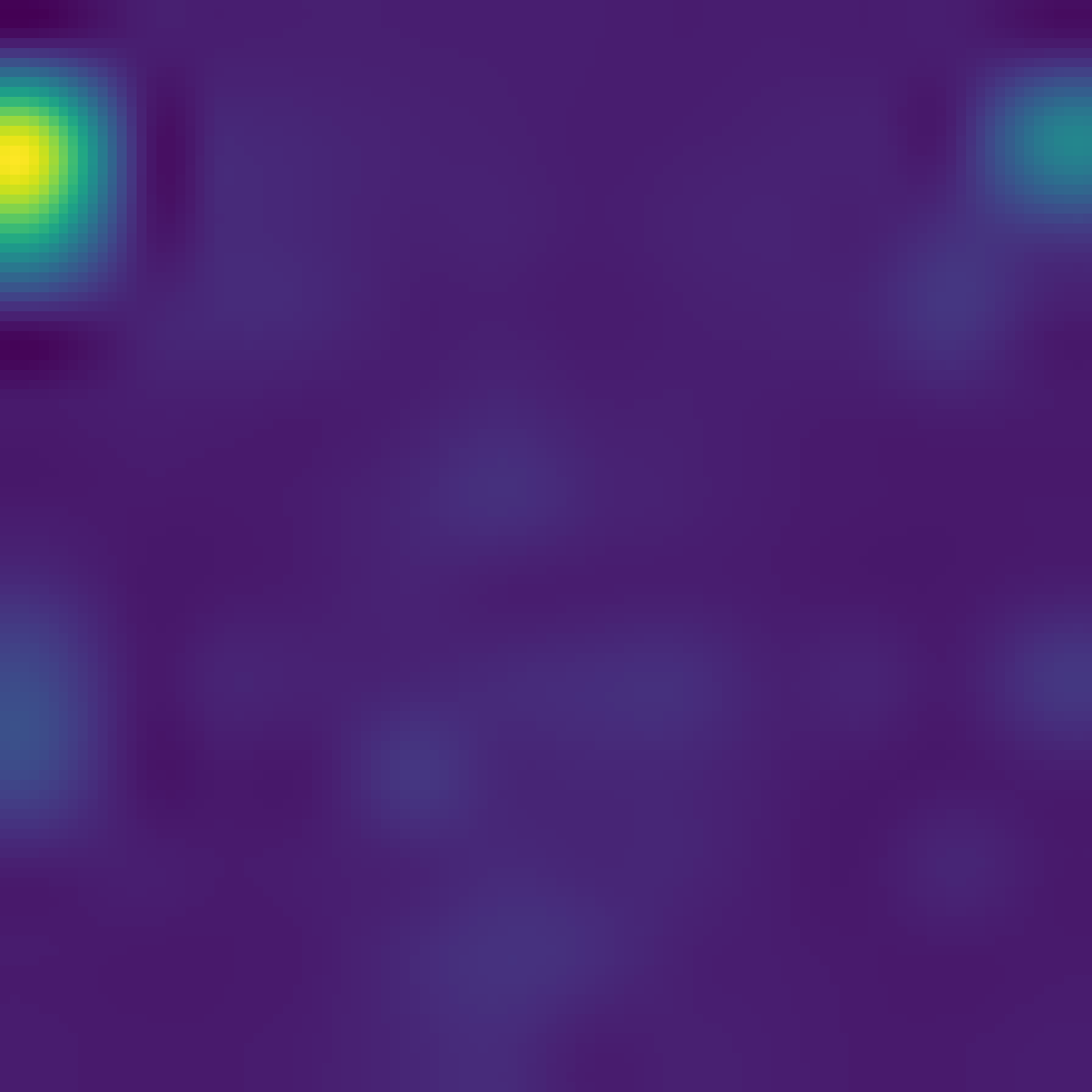}

  \end{subfigure}
  \hfill
    \begin{subfigure}[b]{0.15\linewidth}
      \includegraphics[width=\linewidth, height=1.2cm]{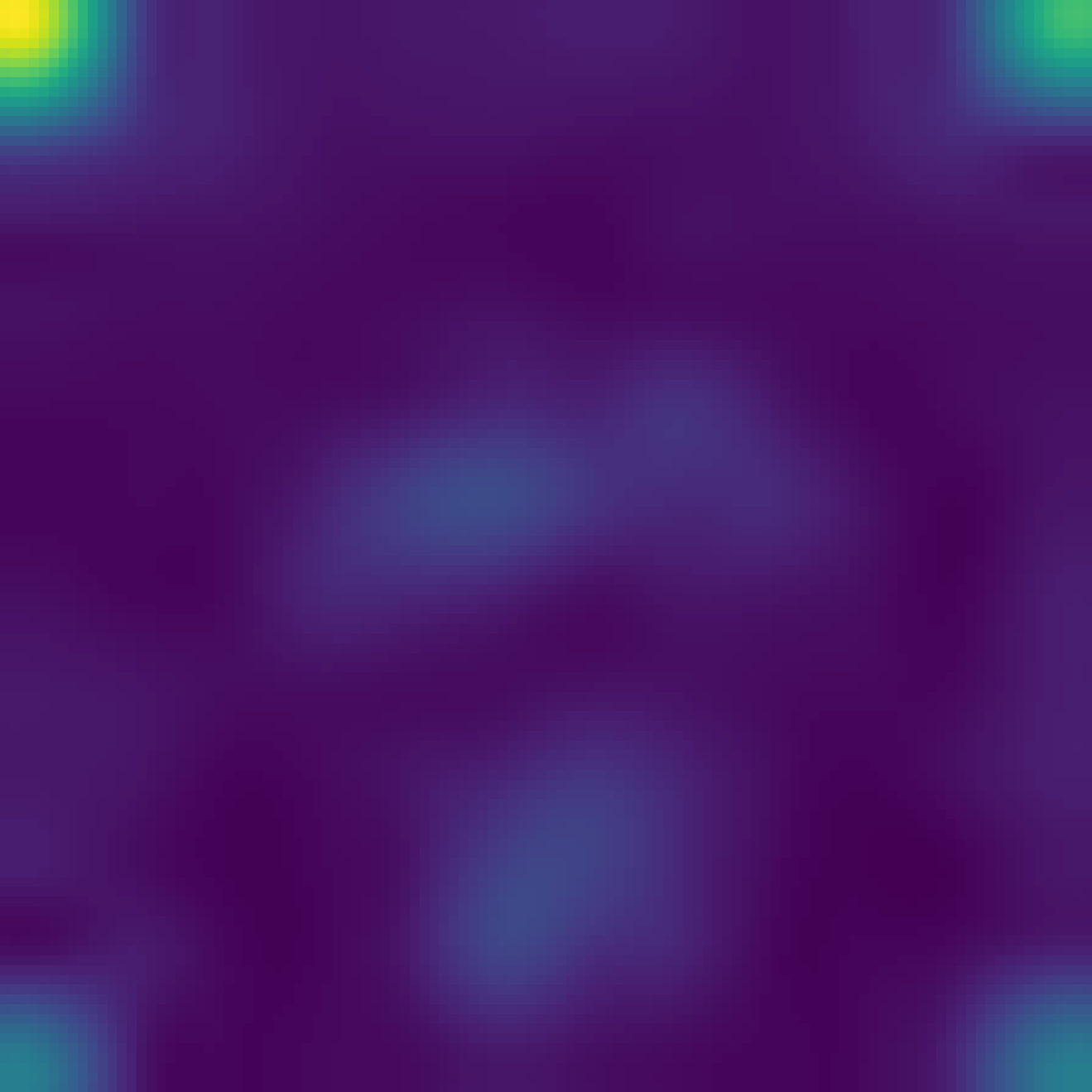}

  \end{subfigure}
  \hfill
    \begin{subfigure}[b]{0.15\linewidth}
      \includegraphics[width=\linewidth, height=1.2cm]{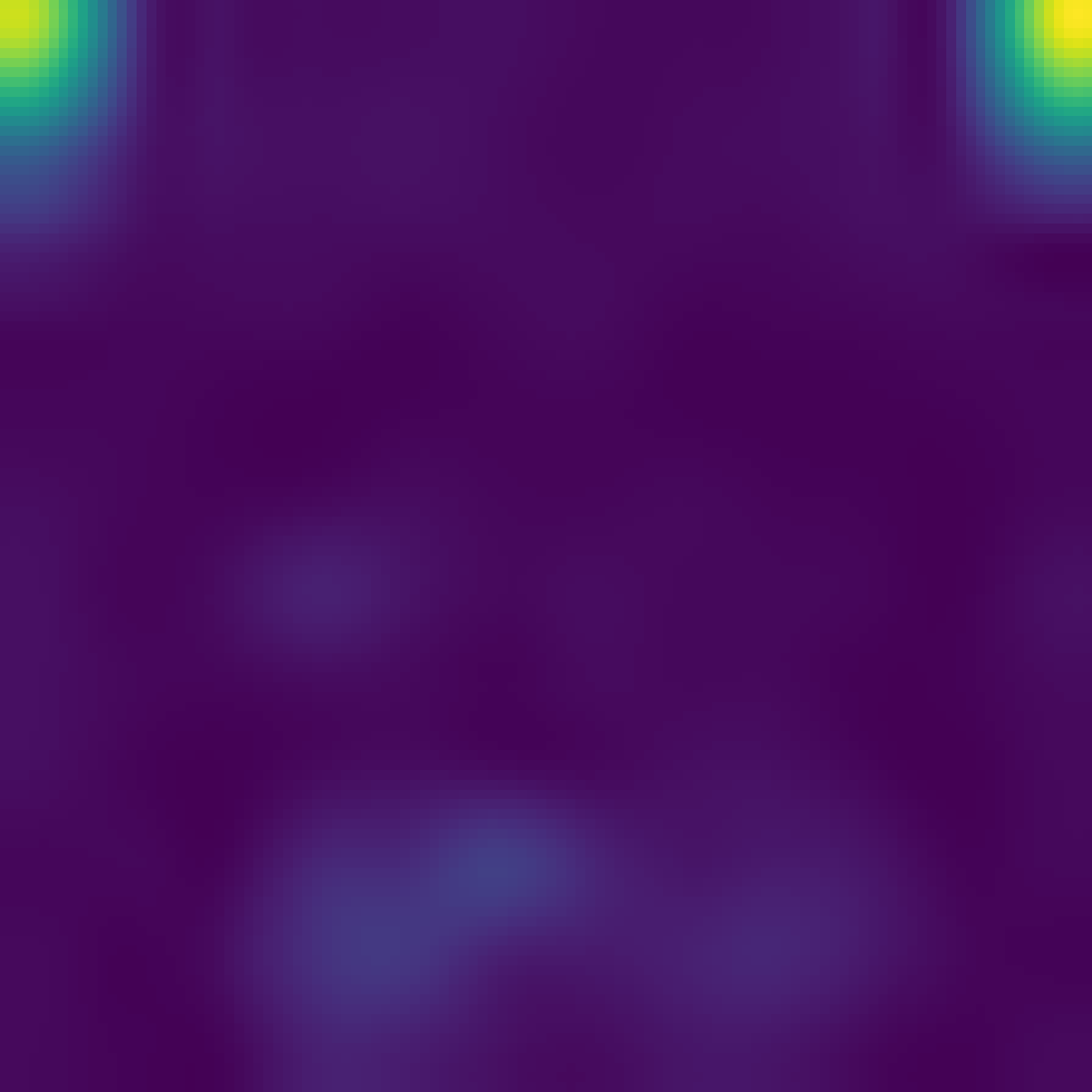}

  \end{subfigure}
  \hfill
  \begin{subfigure}[b]{0.15\linewidth}
      \includegraphics[width=\linewidth, height=1.2cm]{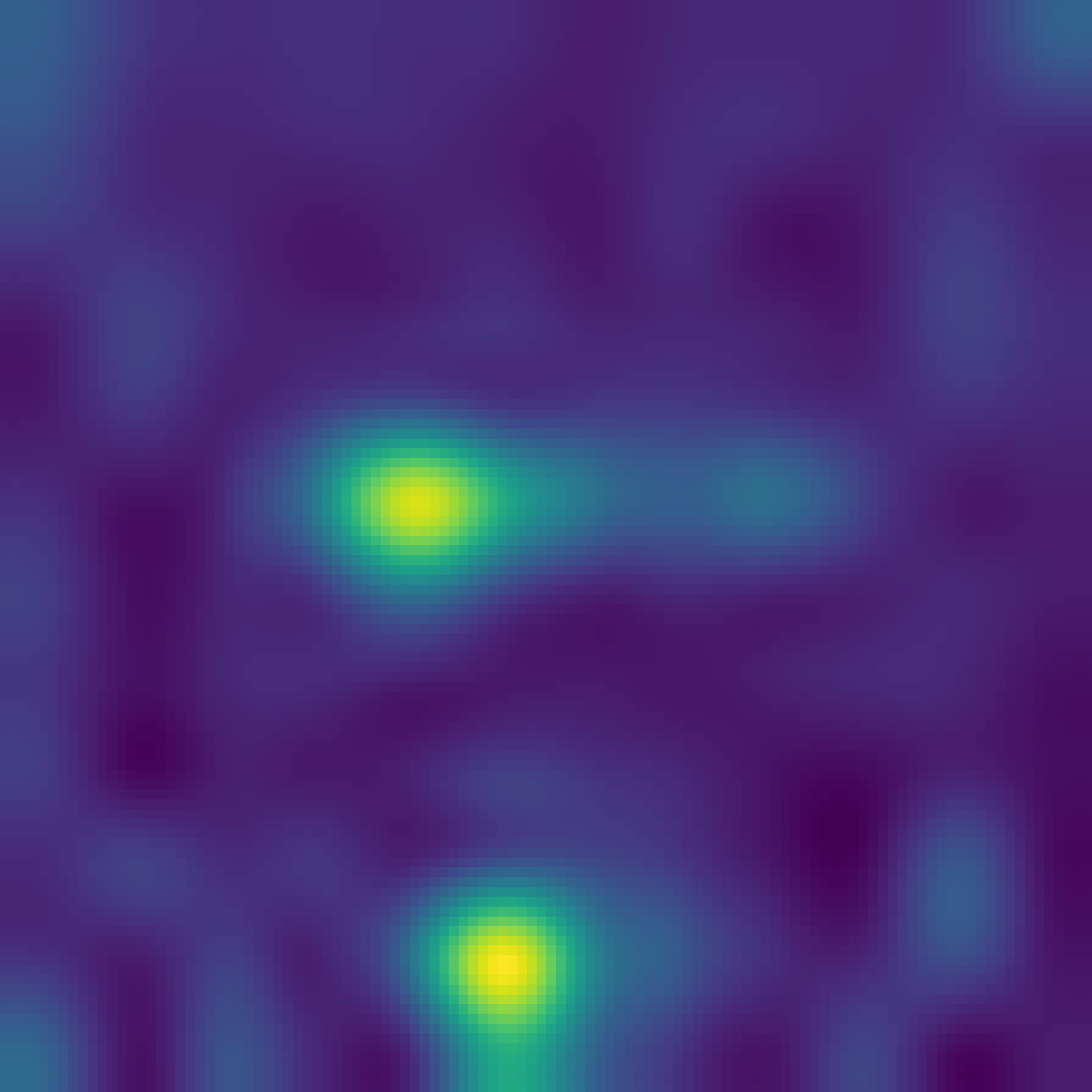}
 
  \end{subfigure}
  \hfill
    \begin{subfigure}[b]{0.15\linewidth}
      \includegraphics[width=\linewidth, height=1.2cm]{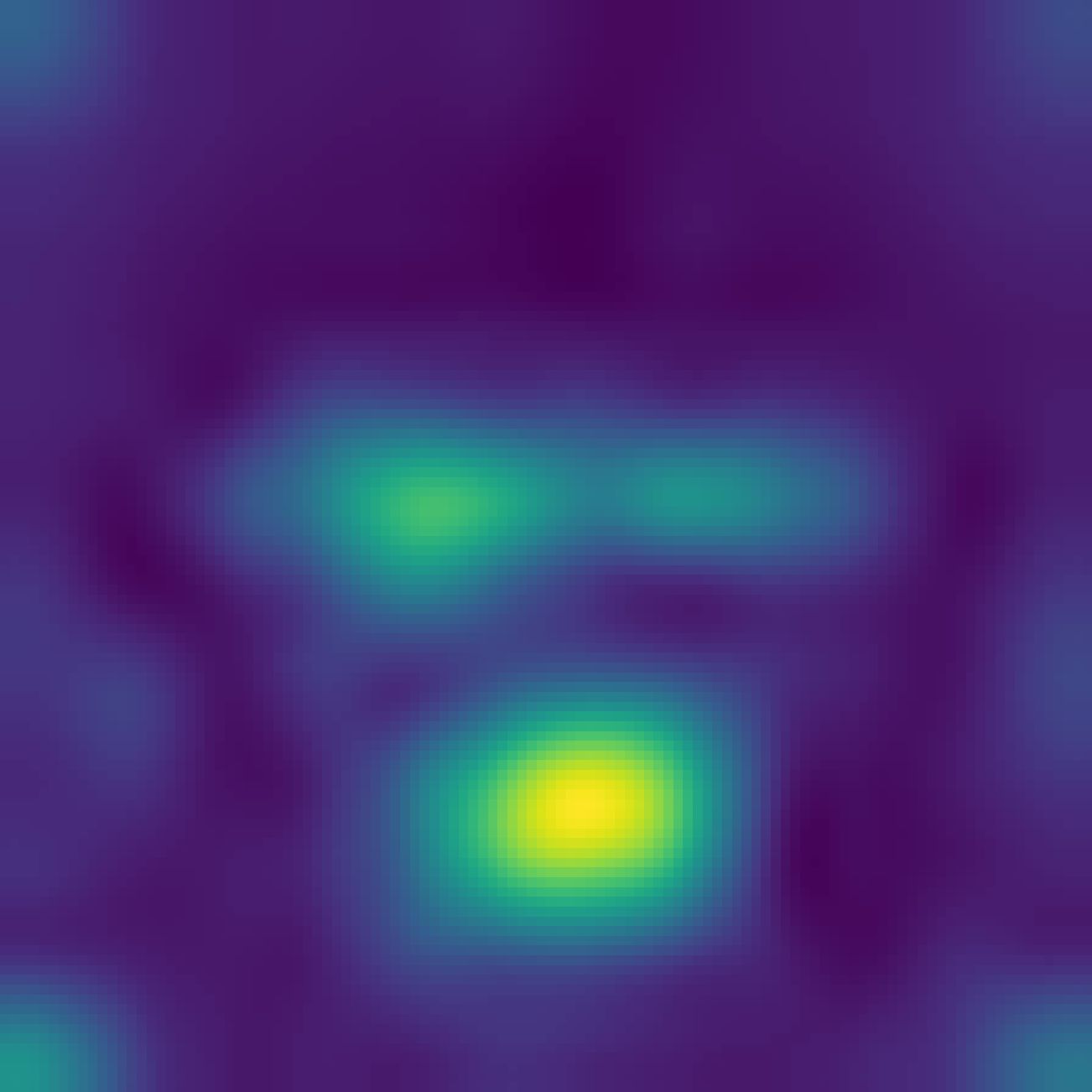}

  \end{subfigure}
\end{minipage}
\caption{Attention maps of CPE-based ViT-B for FR, \rev{shown for the sample image in the first column of each row}, comparing the models trained with 0 (baseline), 1, 2, 4, and 8 Registers (R). Interpretability improves with 4 or 8 registers.} 
\label{fig:attention_sample}
\vspace{-7mm}
\end{figure}

In this work, we first investigate whether the attention maps produced by CLS-based and CPE-based ViTs exhibit artifacts. To this end, we train ViT-L, ViT-B, ViT-S, and ViT-T models without registers and visualize their attention maps from the last attention layer, averaged across five benchmarks, as described in Section \ref{sec:experimentalsetup}. While the \texttt{CLS}-based approach does not exhibit such artifacts, as shown in Section \ref{sec:resultscpe_attention}, \rev{the CPE-based approach does exhibit these artifacts}. Despite this difference, the \texttt{CLS}-based approach achieves lower recognition performance on various FR benchmarks compared to CPE, as reported in Table \ref{tab:cls_cpe}. 
These observations motivate mitigating artifacts in CPE-based models to improve interpretability and \rev{potentially} performance, \rev{while acknowledging that clearer visualizations do not necessarily reflect or guarantee better model behavior \cite{DBLP:conf/naacl/JainW19}.} To this end, we examine next the effect of introducing register tokens to CPE-based ViTs with the aim of mitigating the observed artifacts. For this purpose, we train ViT-B and ViT-S models with 0, 1, 2, 4, 8, and 16 register tokens on MS1MV2 \cite{DBLP:conf/cvpr/DengGXZ19}. Finally, we evaluate our enhanced CPE-based ViT model with eight register tokens (ViT-8R) and compare its performance against SoTA ViT-based FR methods trained on MS1MV2 \cite{DBLP:conf/cvpr/DengGXZ19}, MS1MV3 \cite{DBLP:conf/iccvw/DengGZDLS19}, and WebFace4M \cite{zhu2021webface260mbenchmarkunveilingpower}.
Our results reveal the following key findings:
\begin{itemize}
    \item 
    CPE-based ViTs for FR consistently exhibit artifacts in background patches, a phenomenon observed across various backbones, both small and large.

    \item Introducing register tokens effectively mitigates artifacts. Adding 4 or 8 register tokens substantially enhances interpretability, with 8 tokens providing the best overall face verification accuracies on evaluation benchmarks, in comparison to the baseline model and ones with less registers, and the clearest attention structures.   

    \item Our resulting model ViT-8R, a CPE-based ViT-B model augmented with eight register tokens, achieves SoTA performance among ViT-based FR models trained on MS1MV2, MS1MV3, and WebFace4M, on the large-scale IJB-B and IJB-C benchmarks, while achieving very competitive results on small benchmarks.
\end{itemize}

These findings highlight that register tokens provide a simple yet effective mechanism to improve both the interpretability and performance of ViT-based solutions for FR.
 \rev{While consistent with prior observations that adding register tokens reduces artifacts \cite{DBLP:conf/iclr/DarcetOMB24}, our study reveals several task-specific characteristics for CPE-based ViTs in FR. Specifically, whereas outliers in the original work occur only in the largest CLS-based models (Large, Huge, and Giant) \cite{DBLP:conf/iclr/DarcetOMB24}, we observe artifacts across all model sizes, including Tiny, Small, Base, and Large. In addition, while a single register token suffices in their setup, CPE-based ViTs for FR require four or eight tokens to obtain clear and interpretable attention maps.} 
Figure \ref{fig:attention_sample} presents sample images with attention maps extracted from models trained with 0, 1, 2, 4, 8, and 16 registers, respectively. The model without registers shows pronounced artifacts, whereas the model with 8 registers exhibits minimal artifacts.

\vspace{-1mm}
\section{Related Work}
\vspace{-1mm}
\begin{figure}[!tp]
  \centering
   \includegraphics[width=1\linewidth]{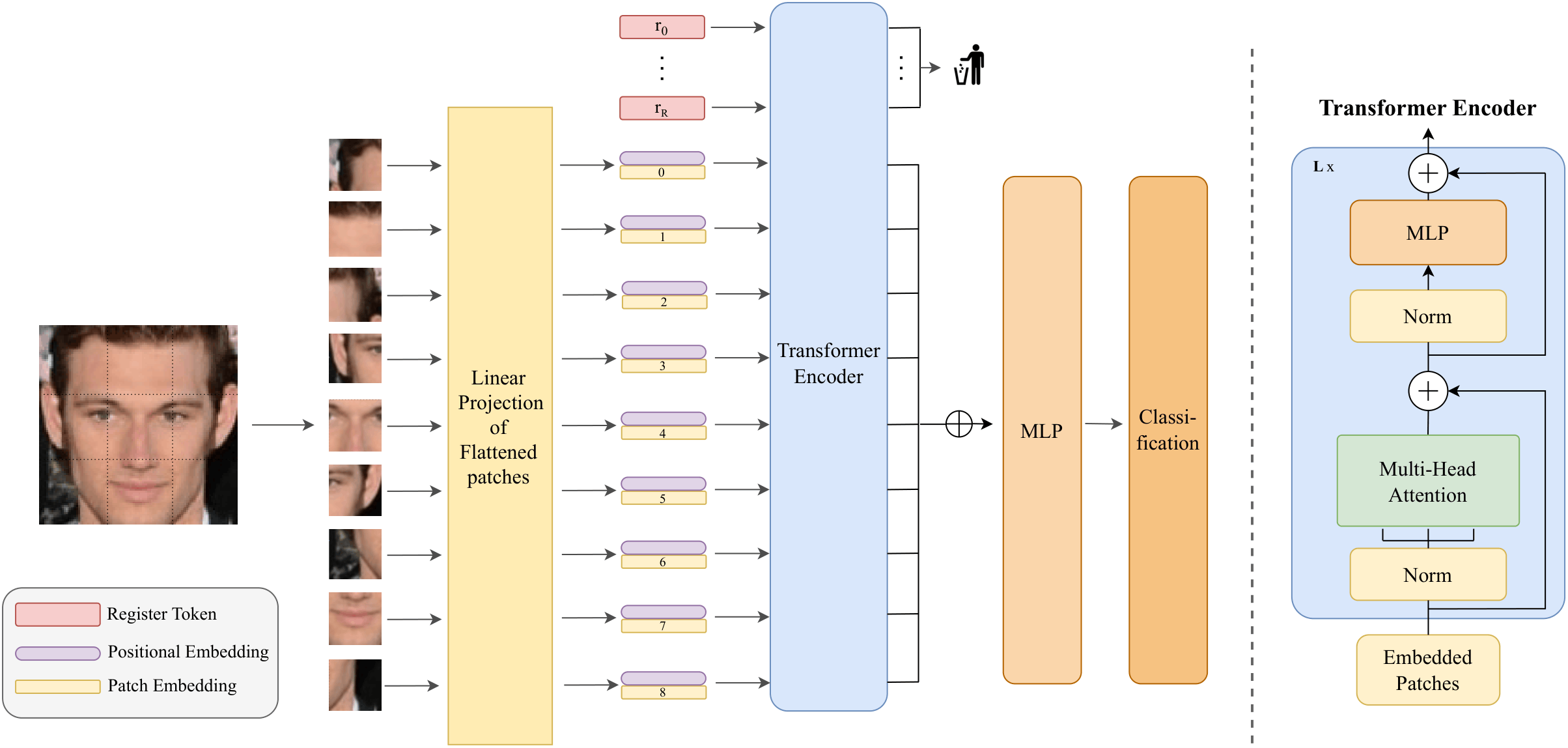}
   \caption{CPE-based ViT with Registers pipeline. The image is divided into patches and projected into the embedding space. Positional embeddings are added to preserve spatial information, then learnable register tokens are appended to the patch embeddings. The full sequence is passed through the transformer encoder. At the output, all patch embeddings are concatenated to form the face representation, while registers are discarded. $\oplus$ denotes the concatenation operator.}
   \label{fig:vit_registers}
   \vspace{-5mm}
\end{figure}

FR has experienced rapid progress over the last decade, primarily driven by advancements in deep learning and the widespread availability of large-scale annotated datasets \cite{guo2016ms, DBLP:conf/iccvw/AnZGXZFWQZZF21, DBLP:conf/cvpr/DengGXZ19, DBLP:conf/iccvw/DengGZDLS19, zhu2021webface260mbenchmarkunveilingpower}. Convolutional Neural Networks (CNNs) such as ResNet \cite{DBLP:conf/cvpr/HeZRS16}, MobileFaceNet \cite{DBLP:conf/ccbr/ChenLGH18}, PocketNet \cite{DBLP:journals/access/BoutrosSKDKK22}, and GhostFaceNet \cite{DBLP:journals/access/AlansariHJSZW23} have dominated the field, achieving SoTA performance on several FR benchmarks \cite{huang:inria-00321923, c3517bca662f4193a58fd8f9e3145c8f, moschoglou2017agedb, DBLP:journals/corr/abs-1708-08197, CPLFWTech, inproceedingsijbb, DBLP:conf/icb/MazeADKMO0NACG18}. However, CNNs inherently suffer from limitations in modeling long-range dependencies due to their localized receptive fields. To address this, ViTs, originally proposed for image classification \cite{dosovitskiy2021imageworth16x16words}, have recently emerged as a compelling alternative, offering greater flexibility and the ability to model global contextual relationships, encouraging a range of solutions \cite{DBLP:conf/bmvc/SunT22, DBLP:conf/isie/KhanSEG23, DBLP:conf/cvpr/KimS0JL24, DBLP:conf/iccv/DanLXD0XS23, DBLP:journals/tcsv/QinWDWCHD24, DBLP:journals/corr/abs-2209-08930, DBLP:journals/corr/abs-2103-14803, DBLP:journals/ivc/ChettaouiDB25, DBLP:conf/bmvc/IslamZM22}  focused on their use in FR.

Typically, ViT performs image classification by introducing a dedicated class token (\texttt{CLS}) \cite{dosovitskiy2021imageworth16x16words, DBLP:journals/tmlr/OquabDMVSKFHMEA24, DBLP:conf/iccv/ChenFP21}, whose final embedding serves as a global image representation. Building on this token configuration, several works \cite{DBLP:conf/bmvc/SunT22, DBLP:conf/isie/KhanSEG23, DBLP:journals/ivc/ChettaouiDB25} have adapted ViT for FR. Sun \cite{DBLP:conf/bmvc/SunT22} proposed a part-based pipeline where a lightweight CNN predicts facial landmarks, and local patches around them are fed to a ViT for part-aware FR. FRoundation \cite{DBLP:journals/ivc/ChettaouiDB25} adapted CLIP \cite{radford2021learningtransferablevisualmodels} and DINOv2 \cite{DBLP:journals/tmlr/OquabDMVSKFHMEA24} foundation models for FR using Low-Rank Adaptation (LoRA) \cite{DBLP:conf/iclr/HuSWALWWC22}, demonstrating the potential of leveraging the inherent generalizability of foundation models in low-data availability scenarios. ARTriViT \cite{DBLP:conf/isie/KhanSEG23} is a triplet loss-based Siamese network with a ViT as a feature extractor. The Siamese network analyzes a pair of face images as input, extracts the characteristics from these pairs, and uses similarity indexes to evaluate them for FR. While the \texttt{CLS}-based token configuration follows the original ViT setup \cite{dosovitskiy2021imageworth16x16words}, other approaches \cite{DBLP:conf/cvpr/KimS0JL24, DBLP:conf/iccv/DanLXD0XS23} discard the \texttt{CLS} token and instead aggregate all patch embeddings. This design, commonly referred to as the concatenated patch embedding (CPE) approach, allows the model to leverage information from all patches rather than relying solely on the \texttt{CLS} token. To make ViT more resilient to scale, translation, and pose variations, Keypoint-Relative Position Encoding (KP-RPE) \cite{DBLP:conf/cvpr/KimS0JL24} builds on Relative Position Encoding (RPE) in ViTs by introducing Keypoint RPE (KP-RPE), which assigns pixel importance based not only on proximity but also on their positions relative to keypoints, enhancing the model’s ability to preserve spatial relationships under affine transformations. TransFace \cite{DBLP:conf/iccv/DanLXD0XS23} addresses overfitting and training instability issues often encountered when training ViTs on large-scale datasets such as MS-Celeb-1M \cite{guo2016ms} and Glint360K\cite{DBLP:conf/iccvw/AnZGXZFWQZZF21}. To address these challenges, TransFace introduced Dominant Patch Amplitude Perturbation, which perturbs dominant patches to enhance generalization, and Entropy-based Hard Sample Mining, which weights samples by uncertainty, boosting accuracy and training efficiency. 

CPE-based ViTs for FR \cite{DBLP:conf/cvpr/KimS0JL24, DBLP:conf/iccv/DanLXD0XS23} have demonstrated superior results in FR, in comparison to the CLS-based ViTs. However, our qualitative analysis of CPE-based ViT attention maps, as shown in Section \ref{sec:results}, exhibits artifacts that make it unclear which regions the model focuses on. While previous work \cite{DBLP:conf/iclr/DarcetOMB24} has proposed well-established techniques to address similar issues in general image classification, none have specifically tackled these attention-map artifacts in CPE-based or CLS-based ViT architectures for FR.  
Following \cite{DBLP:conf/iclr/DarcetOMB24}, we investigated in this work the artifacts that appeared in the attention maps of ViTs for FR. We also evaluated the use of register tokens \cite{DBLP:conf/iclr/DarcetOMB24} as a potential solution to reduce such artifacts, producing visually coherent attention maps.  

\section{Methodology}

This section presents ViT with registers for FR. We first outline the original ViT architecture, then describe register tokens and their interaction with patch embeddings, highlighting their role in mitigating artifacts. The final subsection details how registers are integrated into ViT for FR.

\subsection{ViT (preliminary)} \label{sec:VIT}
Given a facial image \( X \in \mathbb{R}^{H \times W \times C} \), where \( H \), \( W \), and \( C \) represent the height, width, and number of channels, respectively, the image is divided into a sequence of patches, each of size \( P \times P \). These patches are then flattened into vectors, resulting in a patch matrix \( \mathbf{x}_p \in \mathbb{R}^{N \times (P^2 C)} \), where \( N = \frac{HW}{P^2} \) is the total number of patches. These patch vectors are then linearly projected into a latent embedding space of dimension \( D \) using a trainable projection matrix \( \mathbf{E} \in \mathbb{R}^{(P^2C) \times D} \):

\vspace{-2mm}
\begin{equation}
\mathbf{z}_0^i = \mathbf{x}_p^i \mathbf{E}, \quad \text{for } i = 1, \dots, N.
\tag{1}
\end{equation}

Additionally, in vanilla ViTs, a learnable embedding known as the class (\texttt{CLS}) token \cite{dosovitskiy2021imageworth16x16words} is prepended to the sequence of embedded patches, denoted as $\mathbf{z}_0^{\text{\texttt{CLS}}} \in \mathbb{R}^D$. The \texttt{CLS} token serves as a global image representation and is typically used to extract high-level features from the input image \cite{dosovitskiy2021imageworth16x16words, DBLP:journals/tmlr/OquabDMVSKFHMEA24, DBLP:conf/iccv/ChenFP21}. To retain positional information, learnable positional embeddings \( \mathbf{E}_{\text{pos}} \in \mathbb{R}^{(N+1) \times D} \) are added to the input embeddings \( \mathbf{Z}_0 \):

\vspace{-2mm}
\begin{equation}
\mathbf{Z}_0 = [\mathbf{z}_0^{\text{\texttt{CLS}}}; \mathbf{z}_0^1; \dots; \mathbf{z}_0^N], 
\tag{2}
\end{equation}
\begin{equation}
\mathbf{Z}_0' = \mathbf{Z}_0 + \mathbf{E}_{\text{pos}}.
\tag{3}
\end{equation}

The resulting sequence of embedding vectors \( \mathbf{Z}_0' \), comprising patch embeddings and the additional \texttt{CLS} Token, is then fed into the transformer encoder. The transformer maintains this constant latent vector size \( D \) throughout all of its layers. The ViT follows the original transformer encoder design \cite{DBLP:journals/corr/VaswaniSPUJGKP17}, consisting of alternating layers of Multiheaded Self-Attention (MSA) and Multilayer Perceptron (MLP) blocks, each preceded by Layer Normalization (LN) and followed by residual connections. In MSA, we run $k$ heads in parallel, each with its own set of $q$, $k$, and $v$. For an embedding $x$, the computation of the $q$, $k$, and $v$ projection layers in attention head $i$ are:

\vspace{-2mm}
\begin{equation}
Q_i = \mathbf{x} W_i^Q, \quad K_i = \mathbf{x} W_i^K, \quad V_i = \mathbf{x} W_i^V,
\tag{4}
\end{equation}
where \( W_i^Q, W_i^K, W_i^V \in \mathbb{R}^{D \times d} \) are learnable weight matrices, and \( d = D / h \) represents the dimensionality of each attention head given  \( h \) heads. Each head computes scaled dot-product attention:

\vspace{-3mm}
\begin{equation}
\text{Attention}(Q, K, V) = \text{softmax}\left( \frac{QK^\top}{\sqrt{d}} \right)V.
\tag{5}
\end{equation}

The outputs of all heads are concatenated and projected through a linear transformation to form the final attention output. This mechanism allows the model to jointly attend to information from different representation subspaces, enhancing its ability to capture complex relationships across patches \cite{dosovitskiy2021imageworth16x16words}. The output is then passed to the MLP, completing the execution of a single transformer block. These repeated attention and MLP operations refine the token representations throughout the encoder depth, ultimately producing contextualized outputs used for downstream tasks \cite{dosovitskiy2021imageworth16x16words}. In this case, the model processes patch embeddings, generating feature embeddings that capture complex image patterns, with the hidden state of the \texttt{CLS} token serving as the global feature representation of the image.

\subsection{Registers for ViT} \label{sec:registers}
Darcet et al. \cite{DBLP:conf/iclr/DarcetOMB24} identified and characterized artifacts in the feature maps of ViT networks. They showed that artifacts arise from high-norm tokens generated during inference. These tokens occur mainly in low-information background regions and are repurposed by the model for internal computations. To mitigate artifacts in the attention maps, additional learnable tokens, called register tokens, are appended to the sequence of embeddings after the patch projection layer \cite{DBLP:conf/iclr/DarcetOMB24}. The motivation behind introducing these additional tokens is to provide the ViT with learnable tokens that it can use as internal registers, as the model otherwise repurposes tokens from low-informative areas for internal computations. Formally, let $R$ denote the number of register tokens, each with a learnable embedding $\mathbf{r}_0^j \in \mathbb{R}^D$ for $j = 1, \dots, R$. These tokens are concatenated to the patch and \texttt{CLS} embeddings to form the input to the transformer:

\vspace{-2mm}
\begin{equation}
\mathbf{Z}_0^{\text{reg}} = [\mathbf{z}_0^{\text{\texttt{CLS}}}; \mathbf{z}_0^1; \dots; \mathbf{z}_0^N; \mathbf{r}_0^1; \dots; \mathbf{r}_0^R] \in \mathbb{R}^{(N+1+R) \times D}.
\tag{6}
\end{equation}

This sequence is then processed by the standard transformer encoder, with the registers participating in MSA and MLP computations alongside the patch and \texttt{CLS} tokens. Unlike patch embeddings, the register tokens do not receive positional encodings. At the output of the transformer, only the patch and \texttt{CLS} tokens are retained for downstream tasks, while the register tokens are discarded. By isolating internal computations in the registers, this mechanism improves the interpretability of the attention maps. 


\subsection{ViT with Registers for FR} \label{sec:VITFR}

\textbf{CPE-based ViT:}
In original ViT architectures, image classification is performed by introducing a dedicated (\texttt{CLS}) token \cite{dosovitskiy2021imageworth16x16words, DBLP:journals/tmlr/OquabDMVSKFHMEA24, DBLP:conf/iccv/ChenFP21}, whose final embedding serves as a global image representation. In the context of FR, there are two common approaches to extract image-level features. Some works \cite{DBLP:conf/bmvc/SunT22, DBLP:conf/isie/KhanSEG23, DBLP:journals/ivc/ChettaouiDB25} rely on the \texttt{CLS} token to obtain a compact global descriptor, while others \cite{DBLP:conf/cvpr/KimS0JL24, DBLP:conf/iccv/DanLXD0XS23} aggregate all patch embeddings, also referred to as Concatenated Patch Embeddings (CPE), to retain more localized information useful for discrimination. The CPE approach discards the \texttt{CLS} token and instead leverages all output patch embeddings

\vspace{-3mm}
\begin{equation}
\mathbf{Z}_L = [\mathbf{z}_L^1; \dots; \mathbf{z}_L^N] \in \mathbb{R}^{N \times D},
\tag{7}
\end{equation}

produced by the transformer. This method concatenates all patch embeddings into a single vector of dimension \( N \cdot D \), which is then passed through a projection layer to obtain a final feature representation of dimension \( D \). The resulting vector serves as the final image representation. 
As shown in Table \ref{tab:cls_cpe}, the CPE-based approach outperforms the \texttt{CLS}-only configuration. However, unlike the \texttt{CLS}-based approach, the CPE attention maps exhibit artifacts that reduce interpretability, as discussed in Section \ref{sec:resultscpe_attention}, motivating the introduction of register tokens to mitigate these effects.

\textbf{CPE-based ViT with Registers for FR:}
In this token configuration, illustrated in Figure \ref{fig:vit_registers}, the \texttt{CLS} token is discarded and learnable register tokens $\mathbf{r}_0^j$ for $j = 1, \dots, R$ are appended to the sequence of patch embeddings, forming 

\vspace{-4mm}
\begin{equation}
\mathbf{Z}_0^{\text{reg}} = [\mathbf{z}_0^1; \dots; \mathbf{z}_0^N; \mathbf{r}_0^1; \dots; \mathbf{r}_0^R] \in \mathbb{R}^{(N+R) \times D}.
\tag{8}
\end{equation}

This enriched sequence is then passed through a stack of transformer encoder layers, where MSA captures interactions among all patches and registers. The register tokens serve as dedicated placeholders for internal computations, helping to mitigate spurious activations in low-information regions without directly contributing to the final representation. At the output, all patch embeddings are concatenated and passed through an MLP to form the global feature representation for FR, while the register tokens are discarded as implemented in this work. To optimize the ViT for FR, following prior works \cite{DBLP:conf/cvpr/KimS0JL24, DBLP:journals/ivc/ChettaouiDB25, DBLP:conf/bmvc/SunT22}, we enhance the model by incorporating a margin-penalty softmax loss \cite{wang2018cosfacelargemargincosine, Deng_2022}. This is achieved by adding a multi-class classification layer and adjusting the loss function to enforce larger margins between classes, which improves the model's discriminative ability for FR tasks \cite{wang2018cosfacelargemargincosine, Deng_2022, DBLP:conf/cvpr/Kim0L22, DBLP:conf/cvpr/DengGXZ19}.

\vspace{-2mm}
\section{Experimental Setup:} \label{sec:experimentalsetup}
 
\begin{figure}[!t]
  \centering
  \begin{minipage}{\linewidth}
      \centering
        \begin{subfigure}[b]{0.18\linewidth}
          \includegraphics[width=\linewidth, height=1.6cm]{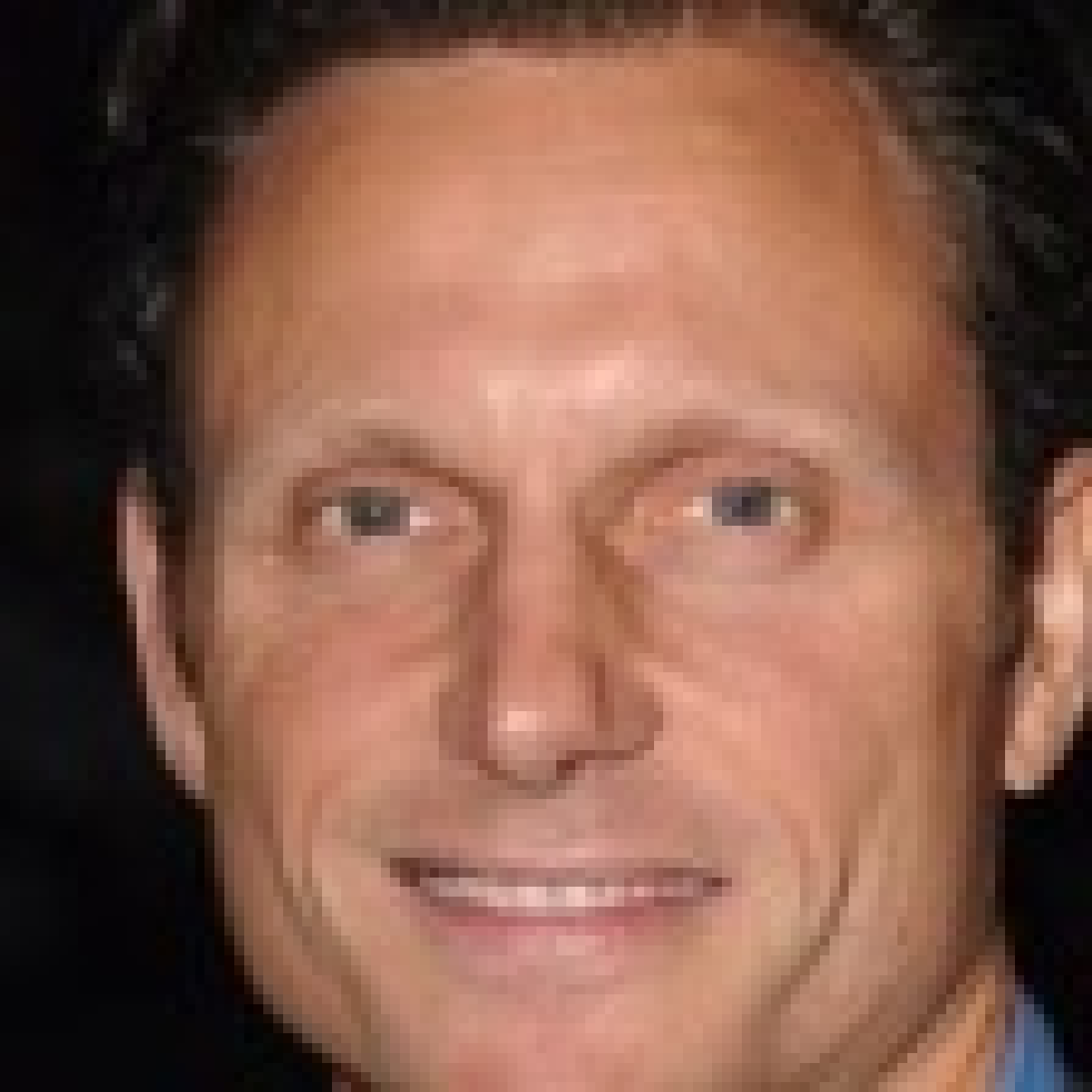}
          \caption{Sample}
      \end{subfigure}
      \hfill
     \begin{subfigure}[b]{0.18\linewidth}
          \includegraphics[width=\linewidth, height=1.6cm]{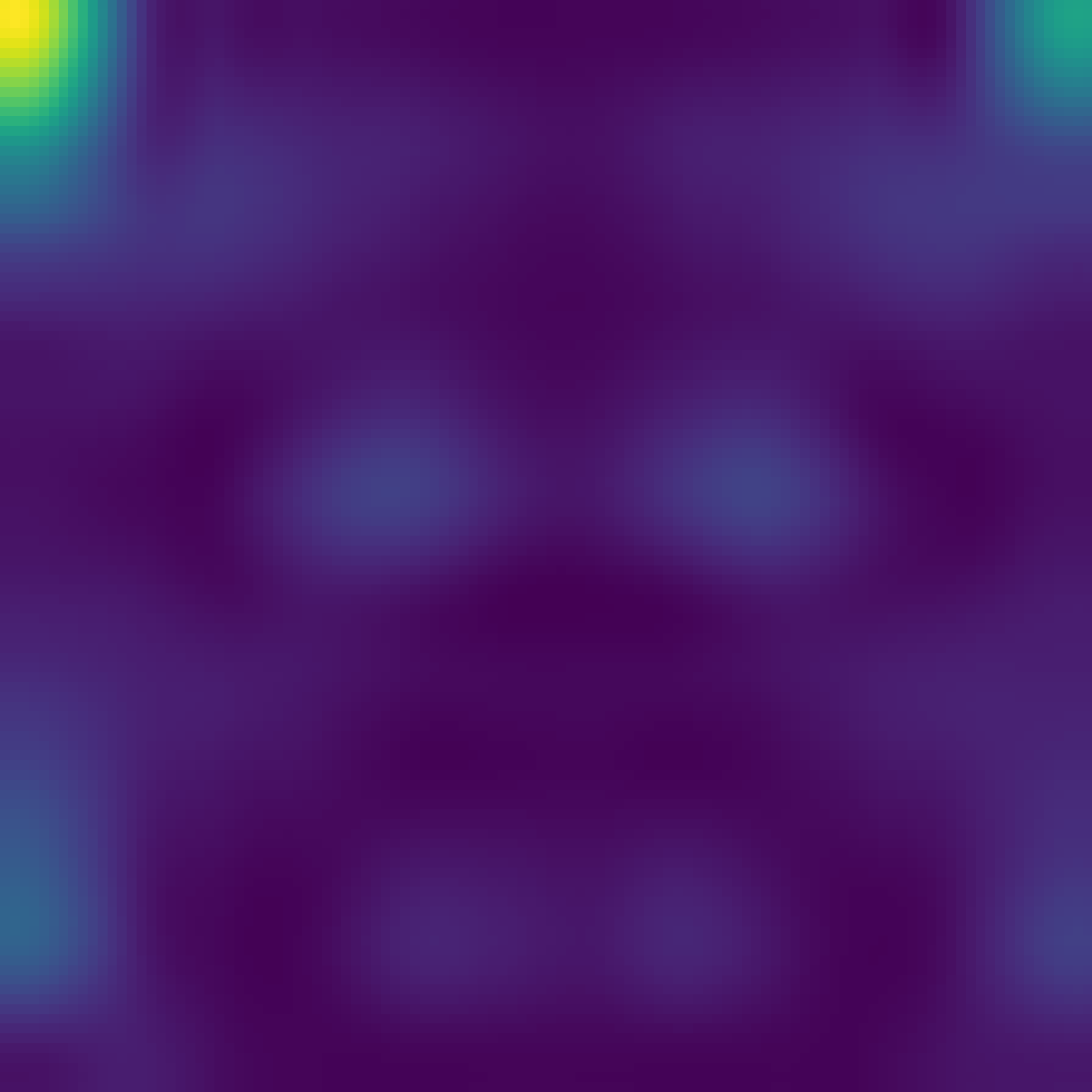}
          \caption{ViT-L}
      \end{subfigure}
      \hfill
      \begin{subfigure}[b]{0.18\linewidth}
          \includegraphics[width=\linewidth, height=1.6cm]{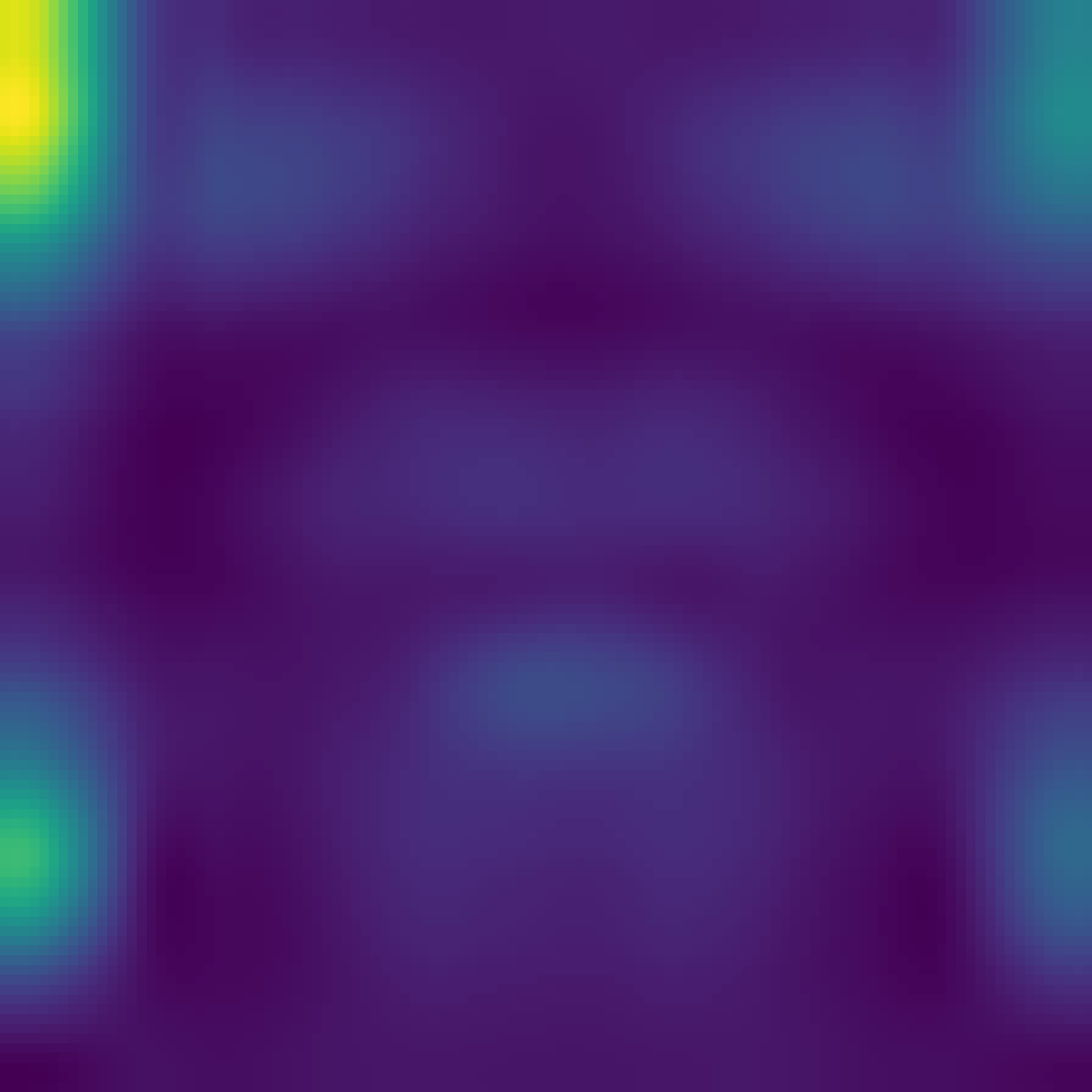}
          \caption{ViT-B}
      \end{subfigure}
      \hfill
      \begin{subfigure}[b]{0.18\linewidth}
          \includegraphics[width=\linewidth, height=1.6cm]{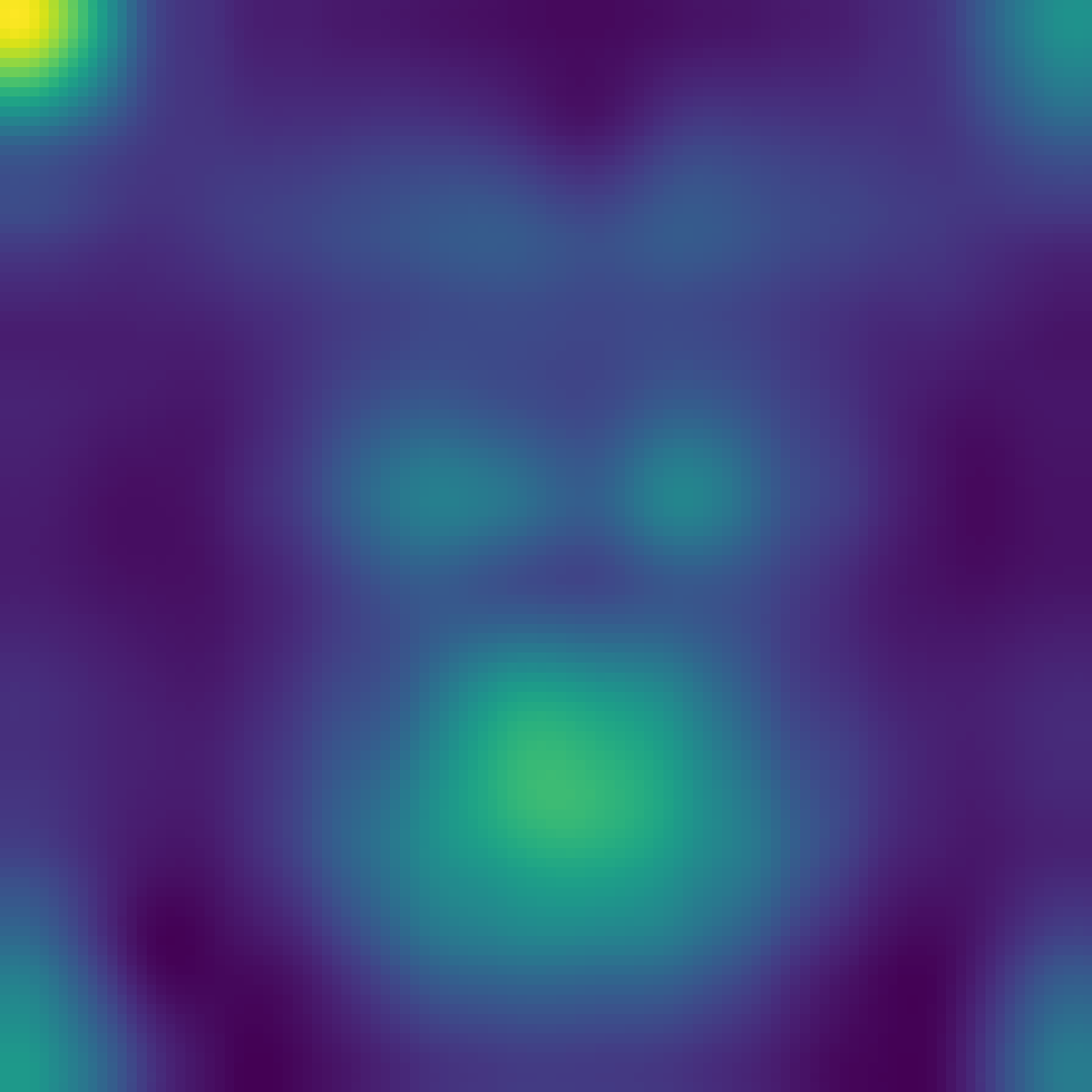}
          \caption{ViT-S}
      \end{subfigure}
      \hfill
        \begin{subfigure}[b]{0.18\linewidth}
          \includegraphics[width=\linewidth, height=1.6cm]{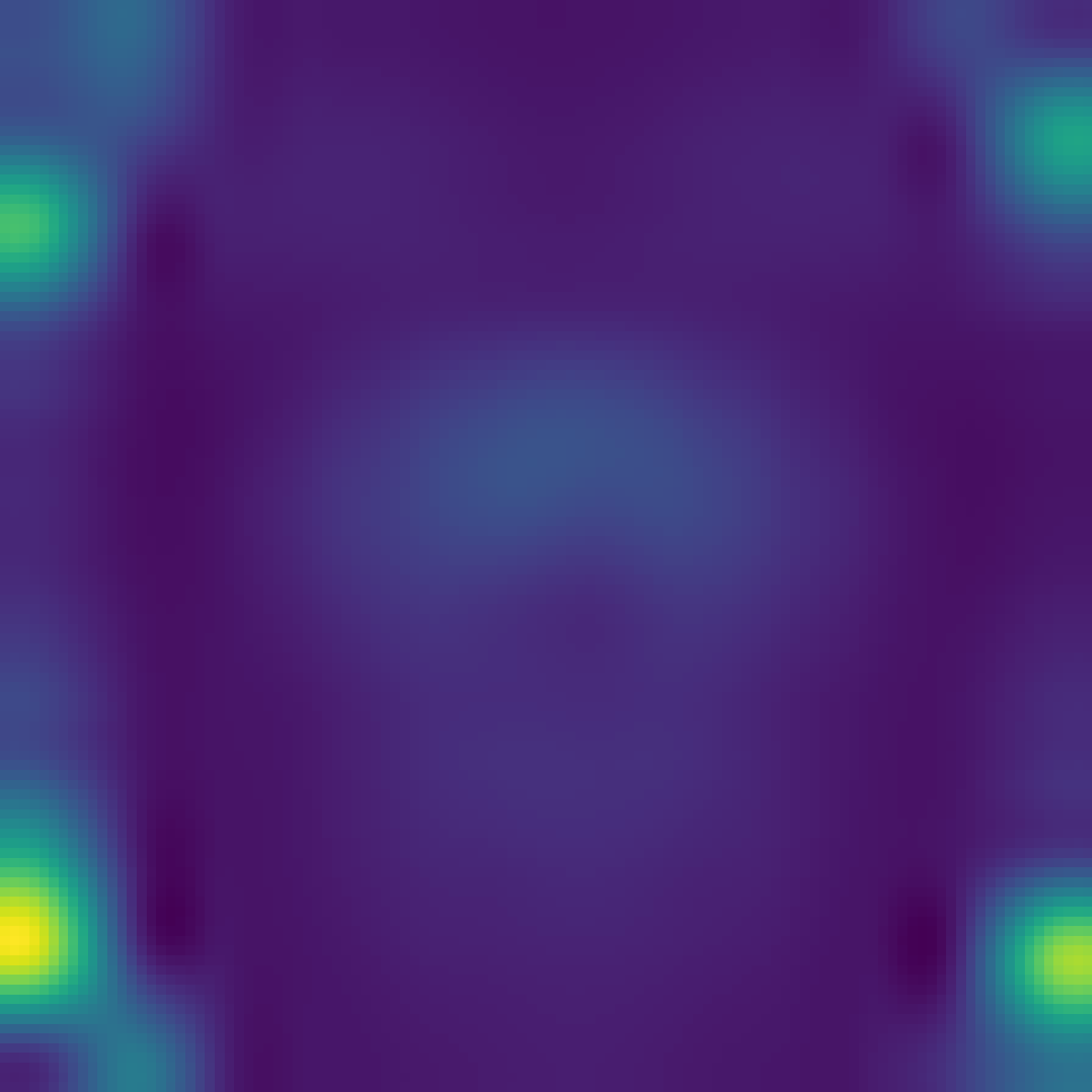}
          \caption{ViT-T}
      \end{subfigure}
  \end{minipage}

  \caption{CPE-based attention maps, \rev{averaged across five benchmarks (see Section \ref{sec:experimentalsetup})}, of ViT-L, ViT-B, ViT-S, and ViT-T backbones for FR. All architectures display artifacts.}
  \label{fig:attention:cpe_vitb_vits}
  \vspace{-2mm}
\end{figure}

\begin{figure}[!t]
  \centering
   \includegraphics[width=0.18\linewidth]{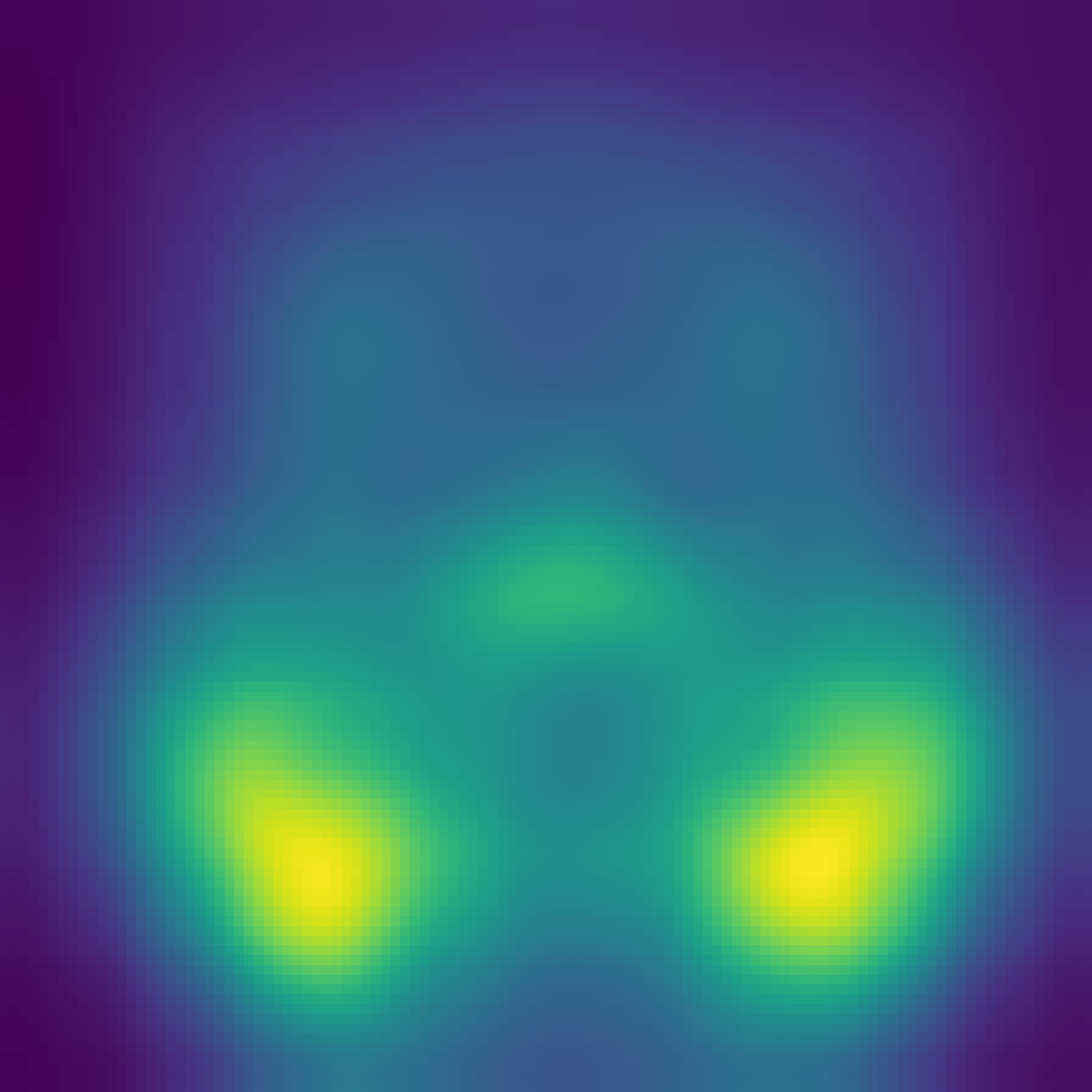}
   \caption{Attention map of the \texttt{CLS} token in ViT-B \rev{averaged across five benchmarks (see Section \ref{sec:experimentalsetup})}. The CLS-based ViT-B does not suffer from background artifacts.}
   \label{fig:cls_attention}
   \vspace{-5mm}
\end{figure}


\begin{figure*}[!t]
  \centering

\begin{minipage}{1\linewidth}
  \centering
    \begin{subfigure}[b]{0.11\linewidth}
      \includegraphics[width=\linewidth, height=2cm]{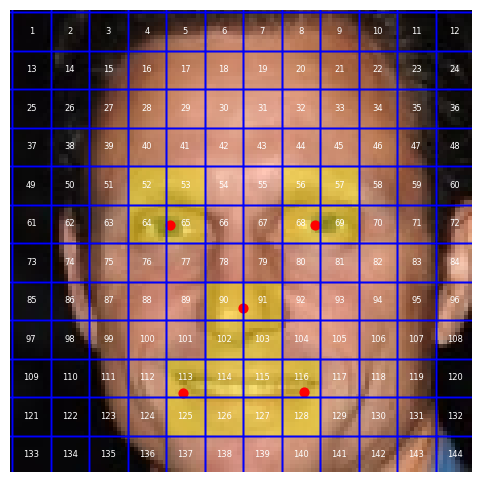}
      \caption{Sample}
  \end{subfigure}
  \hfill
  \begin{subfigure}[b]{0.11\linewidth}
      \includegraphics[width=\linewidth, height=2cm]{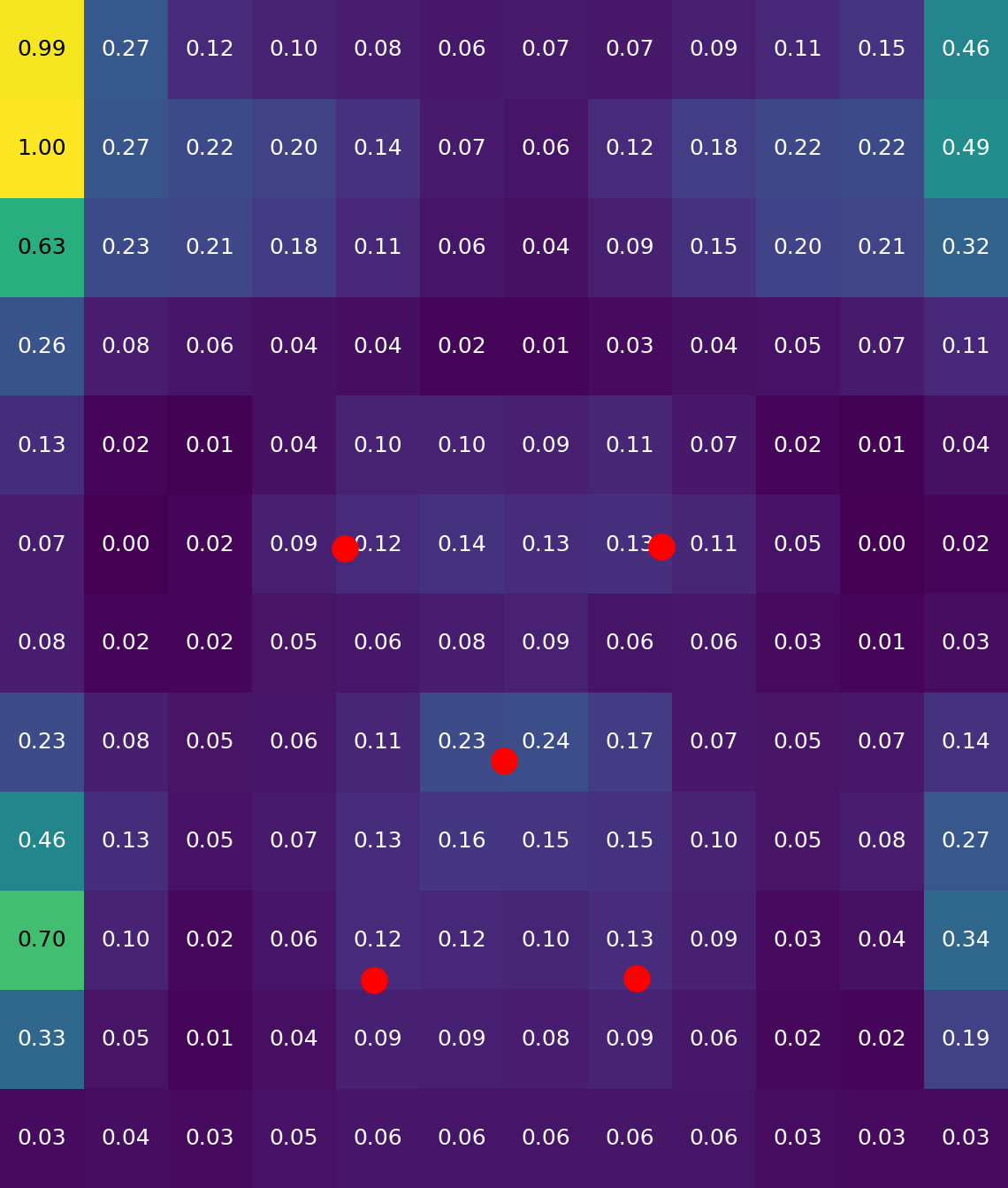}
      \caption{0 Register}
  \end{subfigure}
  \hfill
    \begin{subfigure}[b]{0.11\linewidth}
      \includegraphics[width=\linewidth, height=2cm]{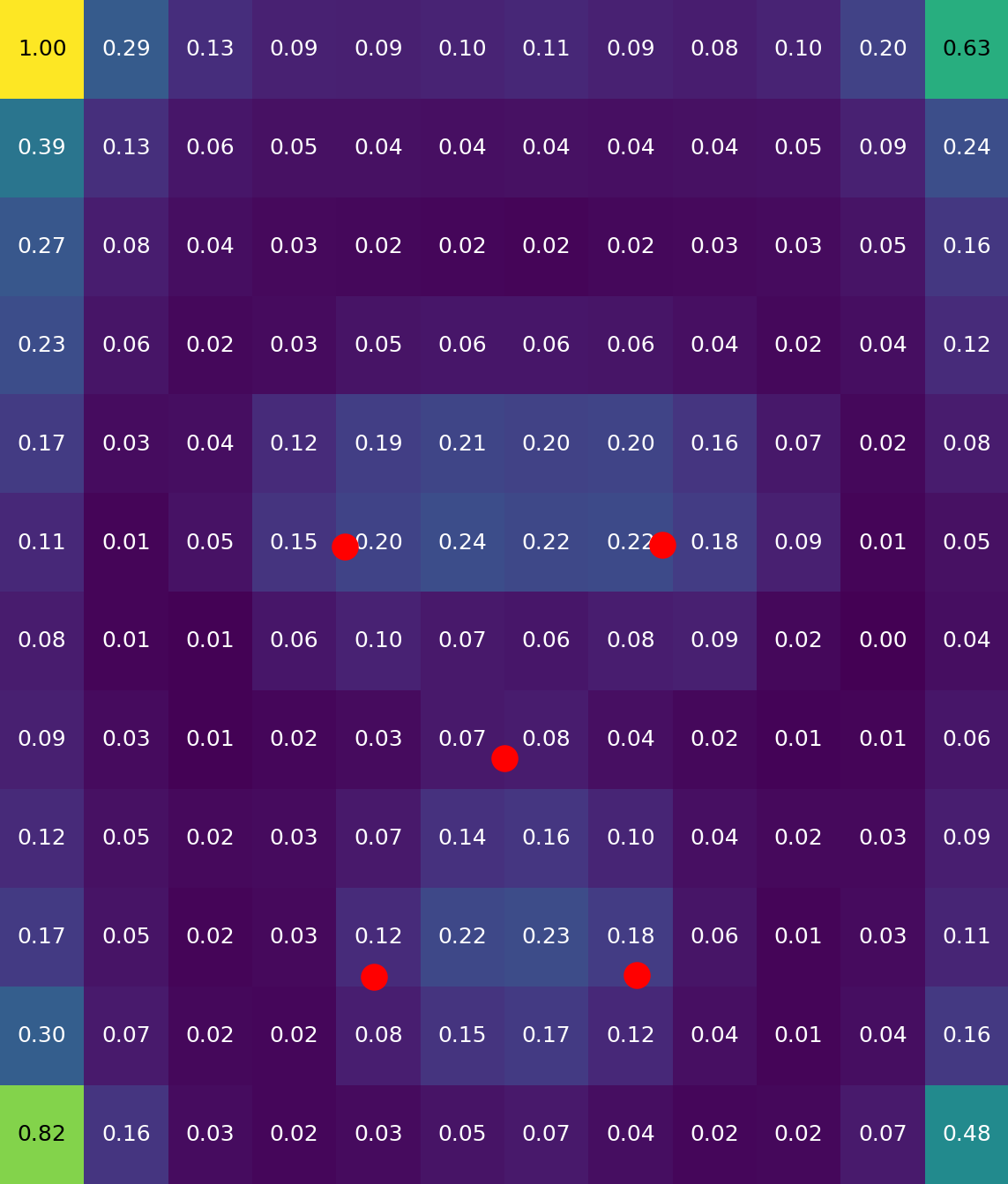}
      \caption{1 Register}
  \end{subfigure}
  \hfill
    \begin{subfigure}[b]{0.11\linewidth}
      \includegraphics[width=\linewidth, height=2cm]{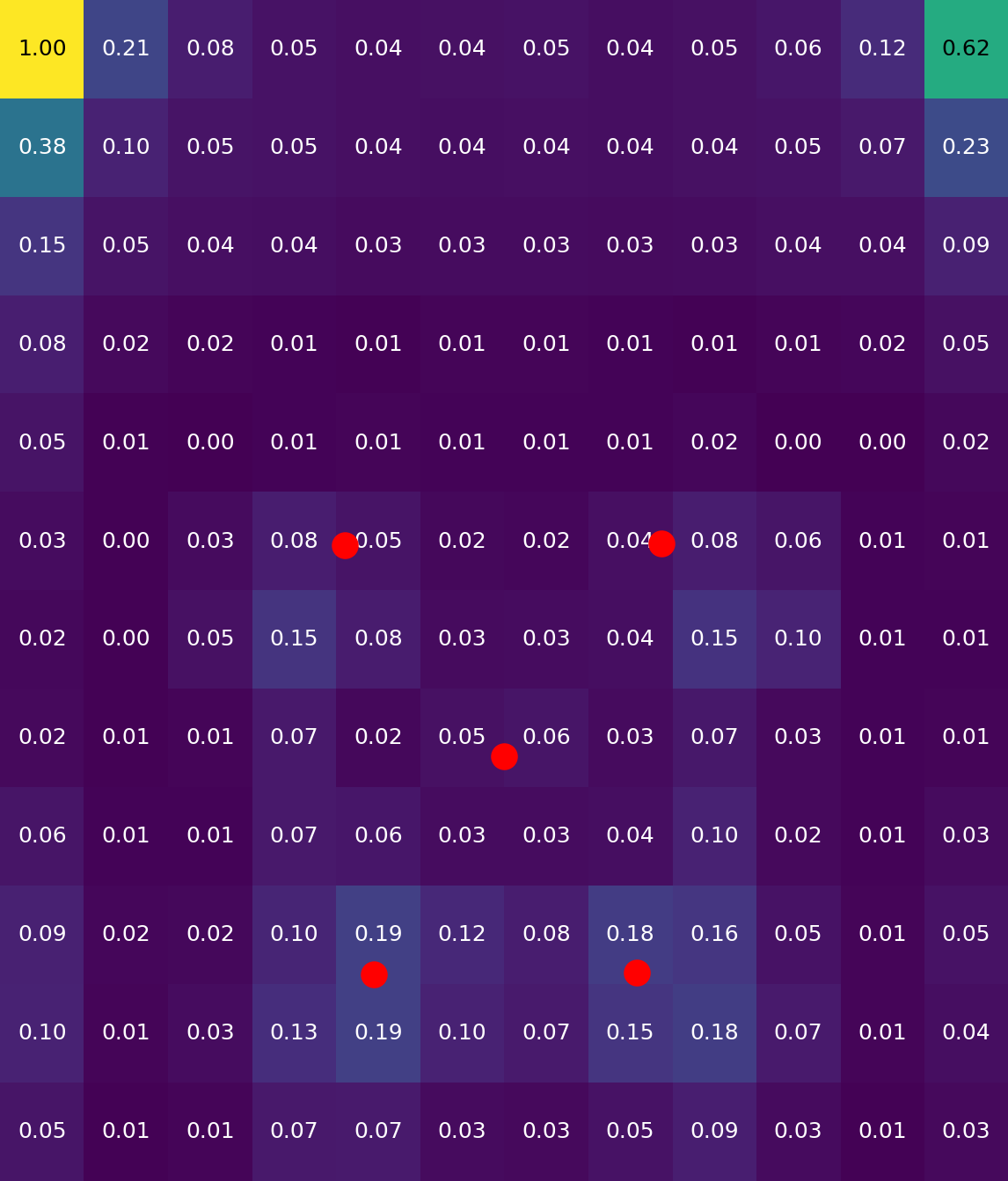}
      \caption{2 Register}
  \end{subfigure}
  \hfill
  \begin{subfigure}[b]{0.11\linewidth}
      \includegraphics[width=\linewidth, height=2cm]{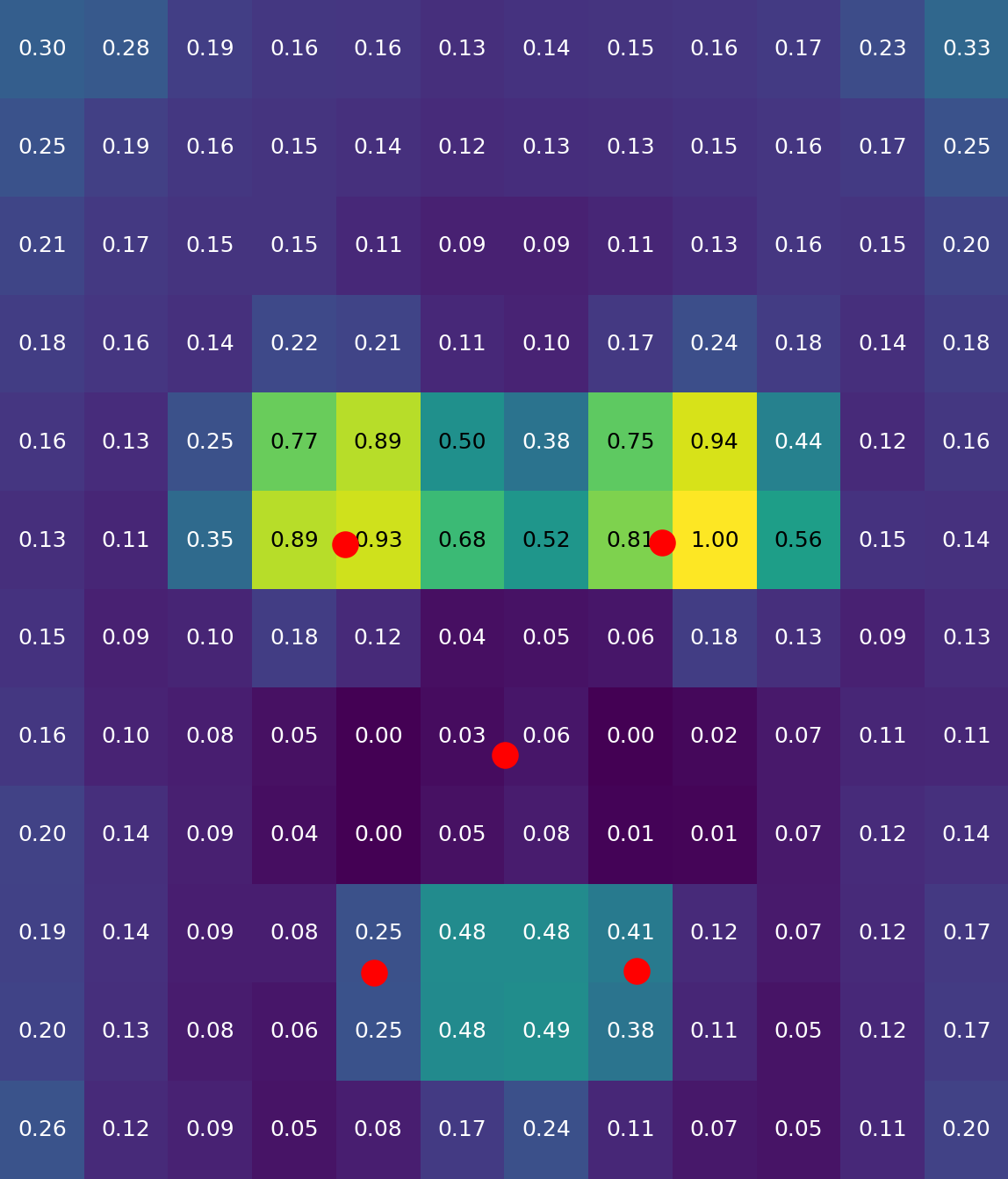}
      \caption{4 Register}
  \end{subfigure}
  \hfill
    \begin{subfigure}[b]{0.11\linewidth}
      \includegraphics[width=\linewidth, height=2cm]{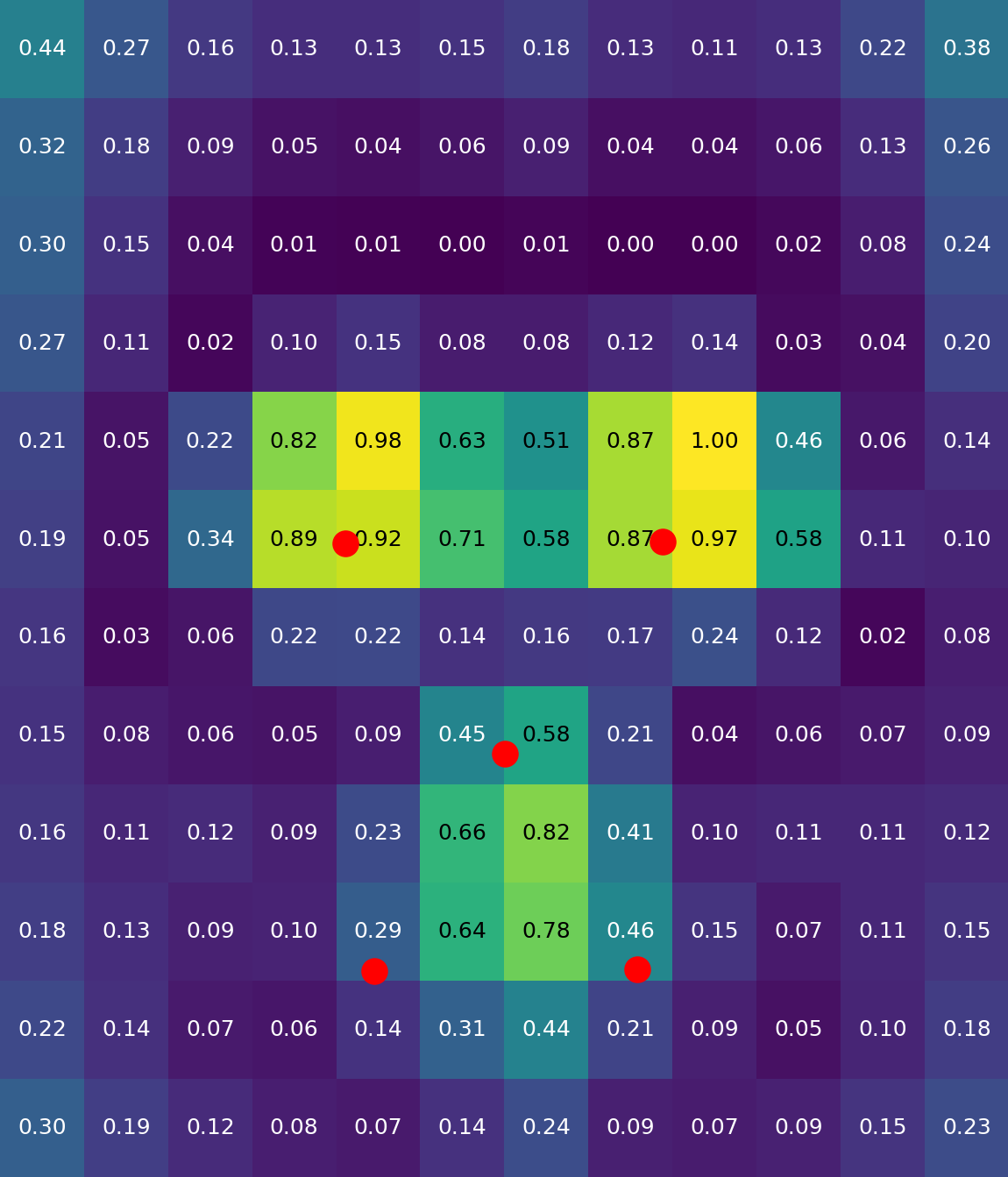}
      \caption{8 Register}
  \end{subfigure}
  \hfill
  \begin{subfigure}[b]{0.11\linewidth}
      \includegraphics[width=\linewidth, height=2cm]{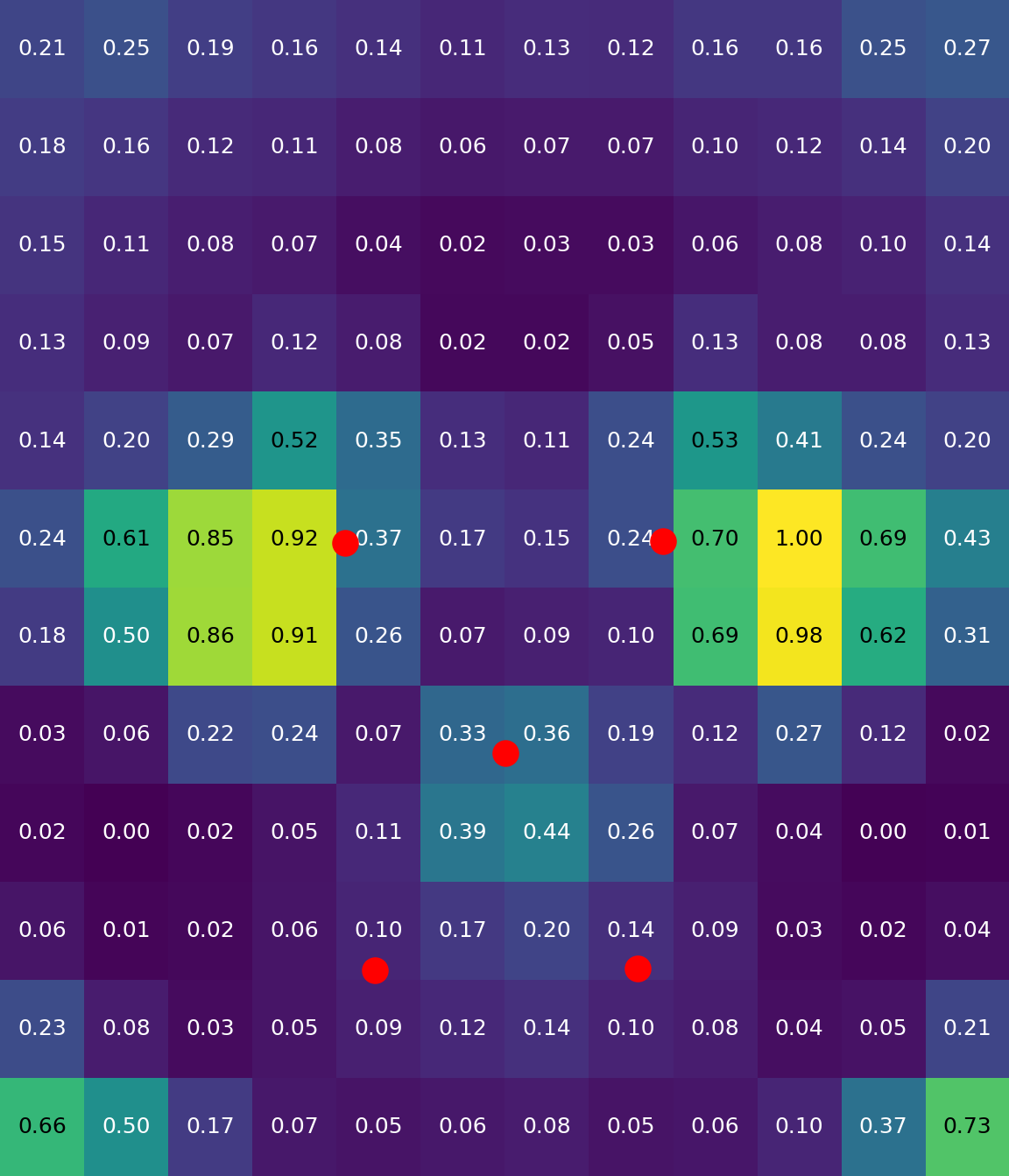}
      \caption{16 Register}
  \end{subfigure}
\end{minipage}
  
  \vspace{0.1cm} 

\begin{minipage}{1\linewidth}
  \centering
    \begin{subfigure}[b]{0.11\linewidth}
      \includegraphics[width=\linewidth, height=2cm]{custom_images/sample_6.png}
      \caption{Sample}
  \end{subfigure}
  \hfill
  \begin{subfigure}[b]{0.11\linewidth}
      \includegraphics[width=\linewidth, height=2cm]{media_new/vitb_cpe_inter.png}
      \caption{0 Register}
  \end{subfigure}
  \hfill
    \begin{subfigure}[b]{0.11\linewidth}
      \includegraphics[width=\linewidth, height=2cm]{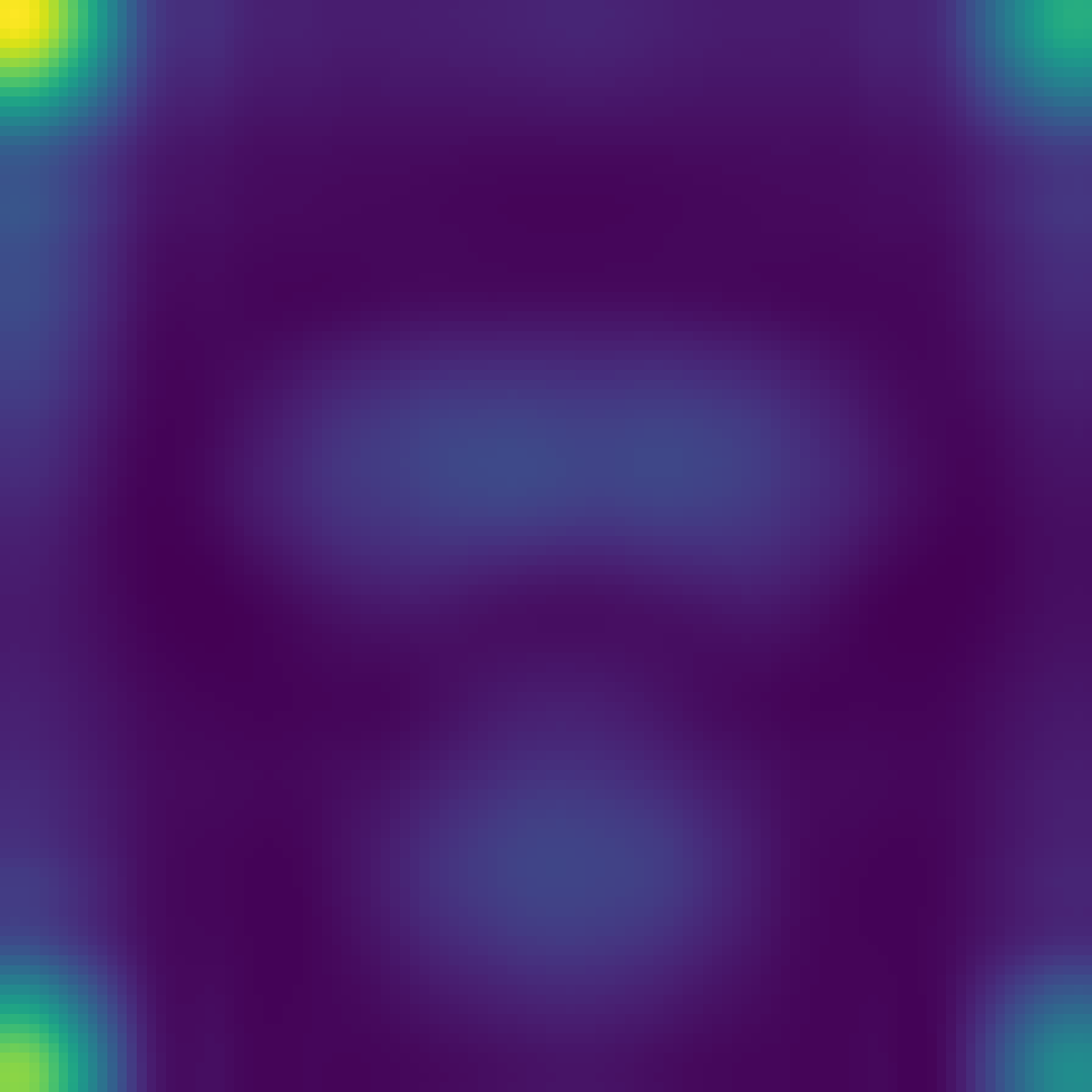}
      \caption{1 Register}
  \end{subfigure}
  \hfill
    \begin{subfigure}[b]{0.11\linewidth}
      \includegraphics[width=\linewidth, height=2cm]{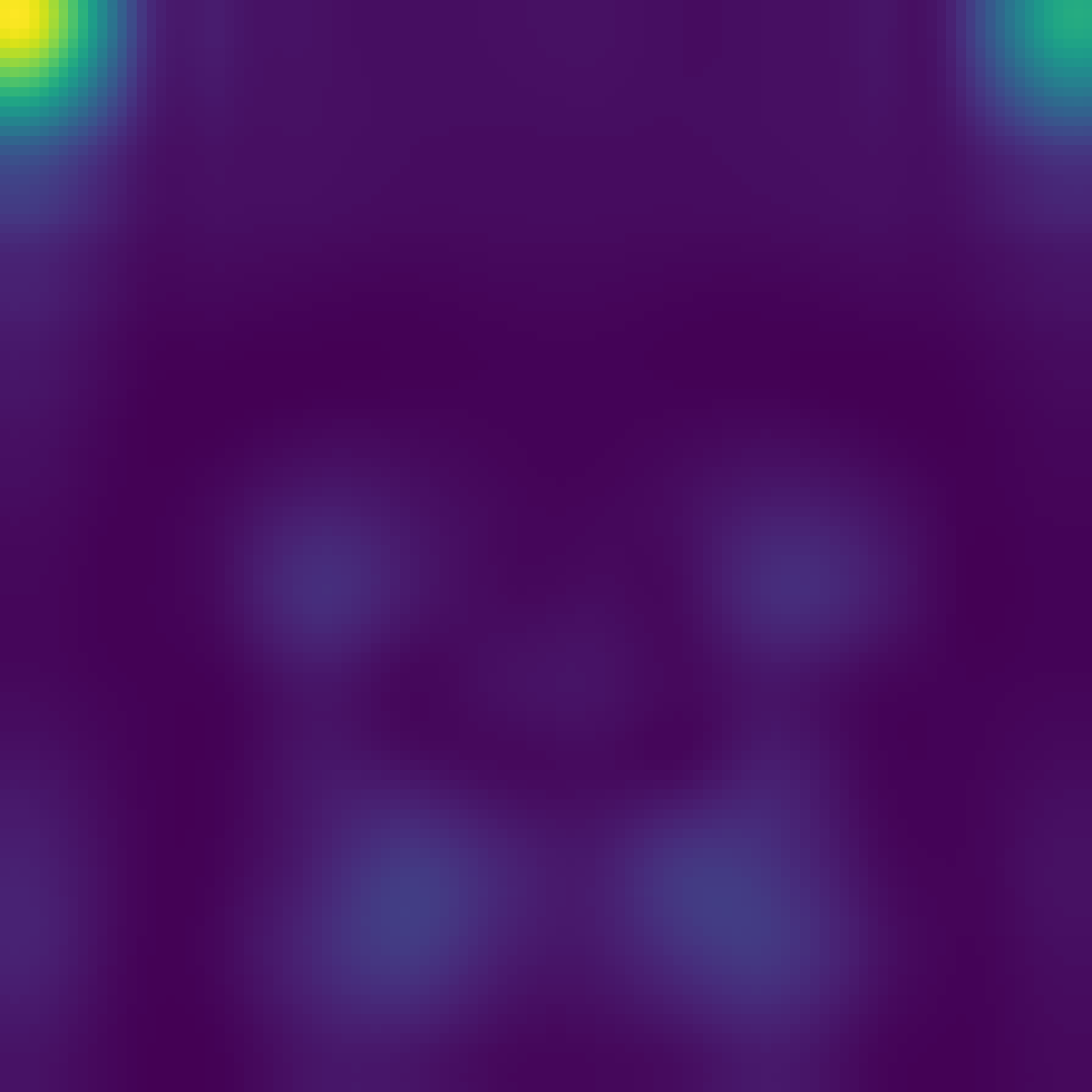}
      \caption{2 Register}
  \end{subfigure}
  \hfill
  \begin{subfigure}[b]{0.11\linewidth}
      \includegraphics[width=\linewidth, height=2cm]{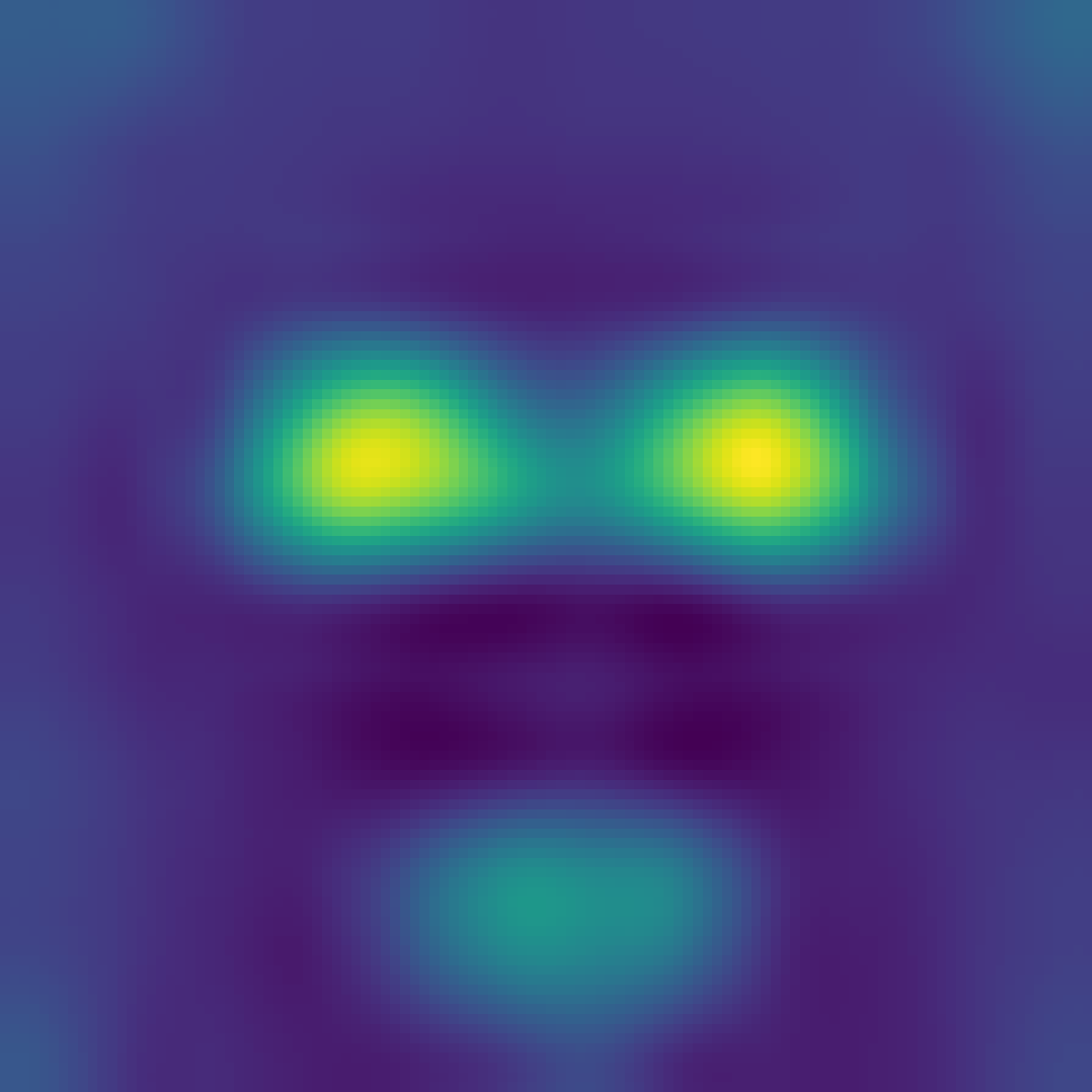}
      \caption{4 Register}
  \end{subfigure}
  \hfill
    \begin{subfigure}[b]{0.11\linewidth}
      \includegraphics[width=\linewidth, height=2cm]{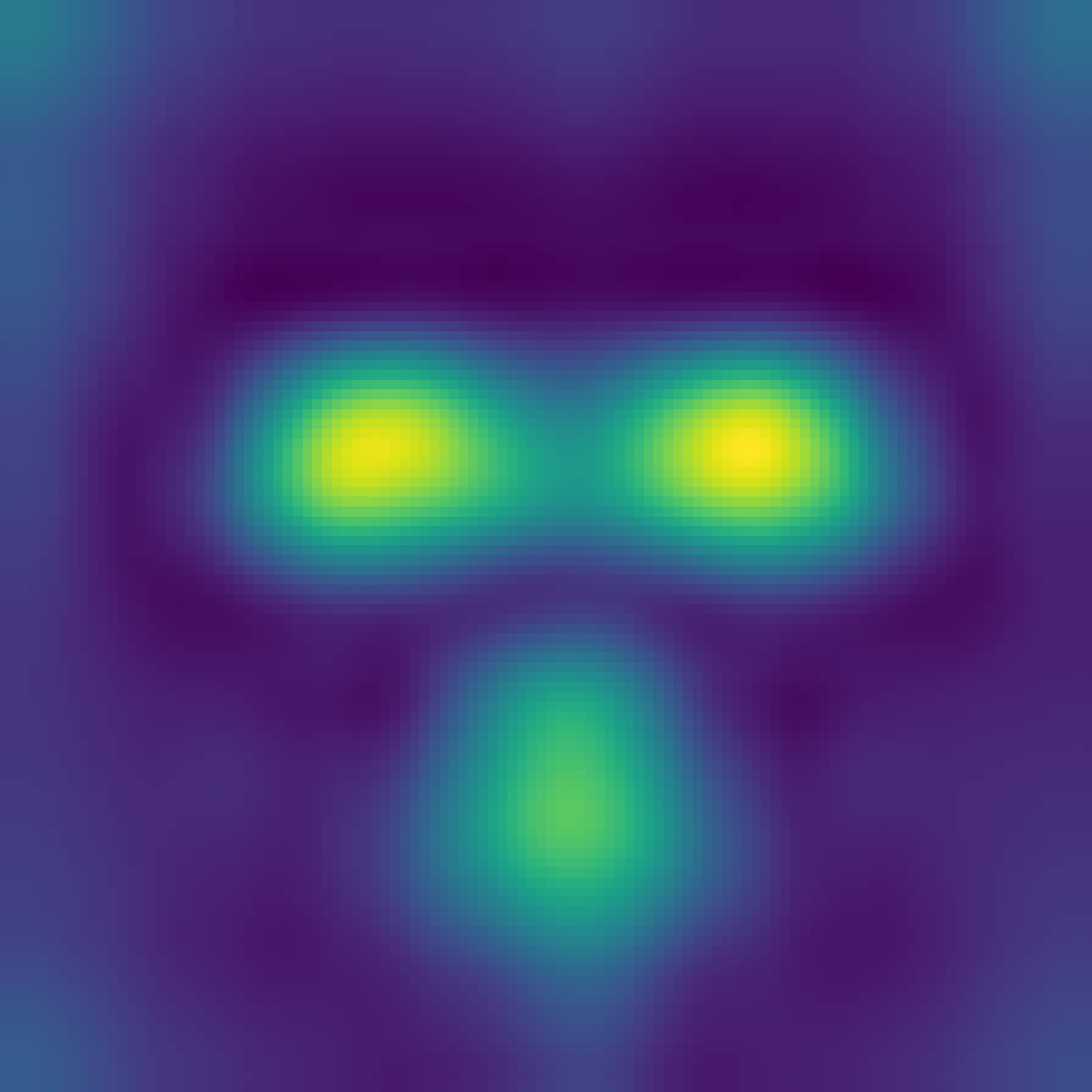}
      \caption{8 Register}
  \end{subfigure}
  \hfill
  \begin{subfigure}[b]{0.11\linewidth}
      \includegraphics[width=\linewidth, height=2cm]{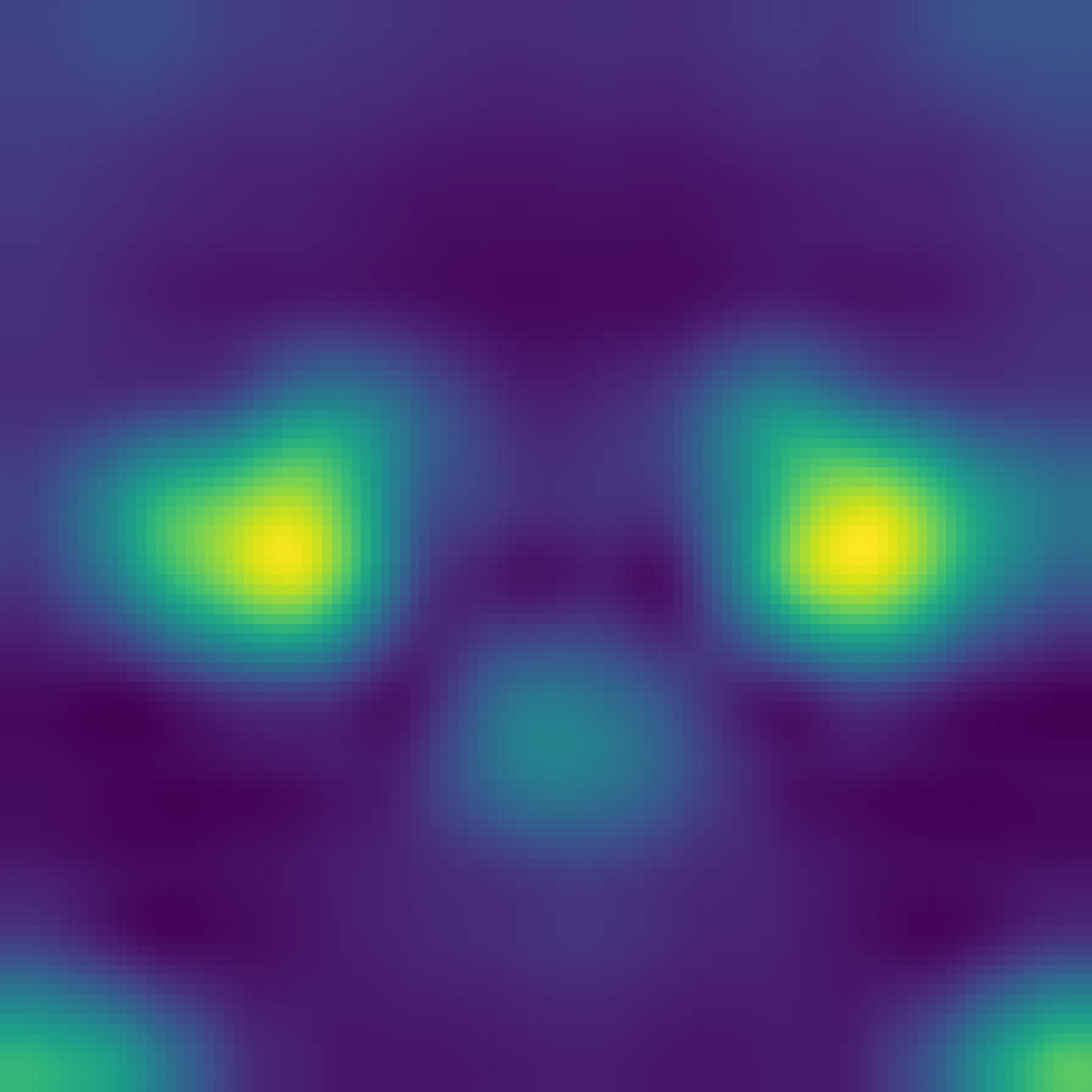}
      \caption{16 Register}
  \end{subfigure}
\end{minipage}
\vspace{-1mm}
\caption{Attention maps of CPE-based ViT-B for FR, \rev{averaged across five benchmarks (see Section \ref{sec:experimentalsetup})}, comparing models trained with 0, 1, 2, 4, 8, and 16 registers. We observe that artifacts are reduced once 4 or 8 registers are added.} 
\label{fig:vit_b_attention_maps}
\vspace{-7mm}
\end{figure*}

\textbf{Training Datasets} 
To train our ViT models,  we employ the MS1MV2 \cite{DBLP:conf/cvpr/DengGXZ19}, MS1MV3 \cite{DBLP:conf/iccvw/DengGZDLS19}, and  WebFace4M \cite{zhu2021webface260mbenchmarkunveilingpower} datasets. These datasets are widely used in FR research, offering a large and diverse set of identities and images. We use them to train our model to enable a wider comparison to SoTA approaches. MS1MV2 is a refined version of the MS-Celeb-1M dataset \cite{guo2016ms}, curated and refined by \cite{Deng_2022}, and comprises 5.8M images of 85K identities. MS1MV3 is a cleaned and updated iteration of MS-Celeb-1M, containing 5.2M images of 96k identities. WebFace4M is a subset of  WebFace260M \cite{zhu2021webface260mbenchmarkunveilingpower}, consisting of 200K identities and 4 million images. All datasets are provided with aligned and resized images (112×112), obtained using five facial landmarks predicted by RetinaFace \cite{DBLP:conf/cvpr/DengGVKZ20}, following \cite{DBLP:conf/cvpr/KimS0JL24}.

\textbf{Evaluation Datasets} 
We evaluate the performance of our models on several FR benchmarks. These include LFW \cite{huang:inria-00321923}, CFP-FP \cite{c3517bca662f4193a58fd8f9e3145c8f}, AgeDB30 \cite{moschoglou2017agedb}, CA-LFW \cite{DBLP:journals/corr/abs-1708-08197}, and CP-LFW \cite{CPLFWTech}. We report verification accuracies (\%) following the official evaluation protocols for each of these benchmarks. In addition, we evaluated on large-scale evaluation benchmarks, IJB-B \cite{inproceedingsijbb}, and IJB-C \cite{DBLP:conf/icb/MazeADKMO0NACG18}. For IJB-C and IJB-B, we used the official 1:1 mixed verification protocol and reported the verification performance as true acceptance rates (TAR) at false acceptance rates (FAR) of $1e-4$ and $1e-5$.

\textbf{Model Architecture:} We employ several ViT backbones of varying sizes, including ViT-L (Large), ViT-B (Base), ViT-S (Small), and ViT-T (Tiny). All models use an input image size of $112 \times 112$ and a patch size of 9 with 8 attention heads. The models differ in their embedding dimensions and the depth of transformer layers. Specifically, ViT-L has an embedding dimension of 768 with 24 transformer layers, while ViT-B has an embedding dimension of 512 and also 24 layers. ViT-S shares the same embedding dimension as ViT-B but has only 12 transformer layers, and ViT-T further reduces the embedding dimension to 256 with 12 layers. 

\textbf{Training Settings:}
Following \cite{DBLP:journals/ivc/ChettaouiDB25}, we employ the CosFace loss function \cite{wang2018cosfacelargemargincosine}. Optimization is carried out using the AdamW optimizer \cite{loshchilov2019decoupledweightdecayregularization} with a weight decay of $0.05$. The model is trained for $40$ epochs using a batch size of $1536$. We set the initial learning rate to $0.001$ and adopt a polynomial learning rate schedule \cite{DBLP:conf/tencon/MishraS19} with a warmup period of $4$ epochs. To improve generalization, we apply a set of data augmentation techniques, following \cite{DBLP:conf/cvpr/KimS0JL24}, including horizontal flipping, brightness, contrast adjustments, scaling, translation, RandAugment \cite{Randaugment_CVPR}, Gaussian blur, cutout, and rotations up to $20^\circ$. Following prior work \cite{Deng_2022, DBLP:conf/cvpr/BoutrosDKK22, DBLP:conf/cvpr/DengGXZ19}, we monitor model convergence after each epoch using several face verification benchmarks, including LFW \cite{huang:inria-00321923}, CALFW \cite{DBLP:journals/corr/abs-1708-08197}, CPLFW \cite{CPLFWTech}, CFP-FP \cite{c3517bca662f4193a58fd8f9e3145c8f}, and AgeDB-30 \cite{moschoglou2017agedb}.

\begin{table*}[!t]
\centering
\resizebox{1\linewidth}{!}{%
\begin{tabular}{cc|cccccc|cc|cc|ccc}
\multirow{2}{*}{\centering Backbone} 
& \multirow{2}{*}{\centering Method} 
& \multirow{2}{*}{\centering LFW} 
& \multirow{2}{*}{\centering CFP-FP} 
& \multirow{2}{*}{\centering AgeDB30} 
& \multirow{2}{*}{\centering CALFW} 
& \multirow{2}{*}{\centering CPLFW} 
& \multirow{2}{*}{\centering Avg.} 
& \multicolumn{2}{|c|}{\rule{0pt}{5.0ex}IJB-B}
& \multicolumn{2}{|c|}{\rule{0pt}{3.0ex}IJB-C} 
& \multicolumn{3}{|c}{\rule{0pt}{3.0ex}Computation Analysis} \\ 
\hhline{~~~~~~~~|--|--|---}
& & & & & & & & $10^{-4}$ & $10^{-5}$ & $10^{-4}$ & $10^{-5}$ & \#Param (M) & GFLOP & $\Delta$ in GFLOP (G) \\ \midrule
\multirow{6}{*}{\centering ViT-B} 
&CPE & 99,82	&98,94	&97,78	&96,00	&94,27	&97,36	&95,53	&87,72	&96,77	&93,32 & 113.834 & 22.89 & \circled{1} \\ 
&CPE + 1-Reg &99,82&	98,83	&97,93	&96,07	&94,53	&97,44&	\textbf{95,76}	&87,87	&\textbf{96,93}	&93,22 & 113.834  & 23.06 & 0.17 + \circled{1} \\ 
&CPE + 2-Reg & 99,80&	98,47&	97,37	&96,13&	94,40	&97,23	&95,42	&87,99&	96.71 &	93.72 & 113.835  & 23.22 & 0.33 + \circled{1} \\ 
&CPE + 4-Reg & 99,75&	98,86&	97,92	&95,98	&94,47&	97,39&	95.52	&88.52&	96,81&	93,89 & 113.835 & 23.55 & 0.66 + \circled{1} \\ 
&CPE + 8-Reg & 99,82	&98,86&	98,00&	96,20&	94,65&	\textbf{97,50}&	95.66& 	\textbf{88.86}&	96.88&	\textbf{94.08} &  113.838 & 24.22 & 1.33 + \circled{1} \\ 
&CPE + 16-Reg & 99,73	&98,40&	97,72	&95,92&	94,47	&97,25&	95.43	&86.57&	96,72	&92,92 & 113.842 &  25.55  & 2.66 + \circled{1} \\ 

\midrule\midrule

\multirow{6}{*}{\centering ViT-S} 
&CPE & 99,83	&98,70	&98,10	&96,12&	94,18&	\textbf{97,39}	&95,39	&87,93	&96.74	&94.14 & \rev{76.023} & \rev{10.99} & \rev{\circled{2}}\\ 
&CPE + 1-Reg & 99,80&	98,56&	97,97&	96,17&	94,23	&97,34	&95.37&	87.77	&96.73	&93.78 & \rev{76.024} & \rev{11.07} & \rev{0.08 + \circled{2}} \\ 
&CPE + 2-Reg & 99,77&	98,69&	97,75	&96,12	&93,95	&97,25	&95.45 &	86.87	&96.73&	93.30	& \rev{76.024} & \rev{11.14} & \rev{0.15 + \circled{2}}\\ 
&CPE + 4-Reg & 99,73	&98,83	&97,97&	96,03	&94,05&	97,32	&\textbf{95,53}	&86,85&	96.75&	93.40& \rev{76.025} & \rev{11.30} & \rev{0.31 + \circled{2}}\\ 
&CPE + 8-Reg & 99,75	&98,79	&97,72	&96,10&	94,17	&97,31&	95,45&	\textbf{89,47}&	\textbf{96,80}&	\textbf{94,59} & \rev{76.027} & \rev{11.60} & \rev{0.61 + \circled{2}}\\ 
&CPE + 16-Reg & 99,80	&98,79	&97,67&	96,15&	94,12&	97,31&	95,37&	87,30&	96,68	&93,57& \rev{76.032} & \rev{12.20} & \rev{1.21 + \circled{2}}\\ 
\hdashline
& \rev{CPE + 8-Reg (w/ Concat)} & \rev{99,75}	& \rev{98,49}	& \rev{98,03}	& \rev{96,15}	& \rev{94,10}	& \rev{97,30}	& \rev{\textbf{95.51}} & \rev{88.34} & \rev{96.79} & \rev{93,77} & \rev{78.124} & \rev{11.60} & \rev{0.61 + \circled{2}}\\ 

\end{tabular}}
\caption{Verification performances by ViT-B and ViT-S for the baseline (CPE) and models with different numbers of registers on several evaluation benchmarks. On IJB-B and IJB-C, the results are reported as TAR at FAR of 1e-4 and 1e-5. The 8-register variant consistently outperforms the baseline on large-scale benchmarks. Additionally, Computation resource comparison in terms of FLOPs and parameter count for CPE-based ViT-B and -S. We measure it as \cite{DBLP:conf/kdd/RasleyRRH20}.}
\label{tab:story_table}
\vspace{-5mm}
\end{table*}

\textbf{Attention Maps Generation:} 
We aim to examine whether the attention patterns differ between the different approaches considered. To this end, we visualize the attention maps extracted from the last attention layer and averaged across five benchmarks: LFW \cite{huang:inria-00321923}, CFP-FP \cite{c3517bca662f4193a58fd8f9e3145c8f}, AgeDB30 \cite{moschoglou2017agedb}, CALFW \cite{DBLP:journals/corr/abs-1708-08197}, and CPLFW \cite{CPLFWTech}. \rev{The self-attention output of the final transformer layer has shape $(H, P, P)$, where $H$ denotes the number of attention heads and $P$ the number of patches. In the original ViT, a \texttt{[CLS]} token is used for classification \cite{dosovitskiy2021imageworth16x16words}, resulting in a self-attention output of shape $(H, P+1, P+1)$. In that case, attention visualization focuses only on the self-attention of the \texttt{[CLS]} token across the heads of the final layer \cite{DBLP:conf/iclr/DarcetOMB24, DBLP:conf/iccv/CaronTMJMBJ21}. Since CPE-based ViT does not employ a \texttt{[CLS]} token and instead processes all patch embeddings jointly, attention visualization is performed by aggregating self-attention across all patch tokens. We first average the attention maps across heads, obtaining a single attention matrix of shape $(P, P)$, which represents how each patch attends to all other patches. Then, we average over the attended-patch dimension, producing a vector of shape $P$ that represents the total attention received by each patch. Finally, this vector is reshaped into a grid to recover the spatial arrangement of patches and is resized to the input image resolution for visualization.} This visualization provides a qualitative view of how the two configurations distribute attention across the image, offering insights into their differing attention behaviors.

\textbf{\rev{Attention Maps Quantitative Analysis: }}
\rev{To quantitatively assess the model's attention behavior, we calculate the Mean Attention-Weighted Similarity (MAWS) to measure the average similarity from each patch, weighted by its attention, to the closest facial landmark. We use the same reference facial landmarks employed to align the training datasets. For both the training and testing sets, face regions and landmarks are detected using RetinaFace \cite{DBLP:conf/cvpr/DengGVKZ20}. The five facial keypoints (two eyes, nose, and two mouth corners) are then used to perform a similarity transformation for alignment. Let each patch $i$ have center coordinates in image space, and $d_i$ is the Euclidean distance from that patch to its nearest facial landmark. The attention weight for patch $i$, denoted $a_i$, is normalized such that $\sum_i a_i = 1$. To make the metric independent of image resolution, we normalize distances by $\sqrt{H^2 + W^2}$, giving
\[
\vspace{-2mm}
\text{MAWS} = 1 - \sum_i a_i \, \frac{d_i}{\sqrt{H^2 + W^2}}.
\vspace{-1mm}
\]
Higher values indicate that attention is concentrated near landmark locations, while lower values indicate more diffuse attention. Besides MAWS, we calculate the average attention in regions surrounding facial landmarks across the considered models, illustrating how registers affect the distribution of attention on critical facial features that hold important cues for FR. We consider four patches covering the regions around key facial landmarks (left eye, right eye, nose tip, and the left and right corners of the mouth), which are highlighted in yellow, with the landmarks indicated as red dots, as shown in Figure \ref{fig:vit_b_attention_maps}-a. We focus on these landmarks because they are the same reference points used to align the training data, where a similarity transformation is performed from landmarks detected by RetinaFace \cite{DBLP:conf/cvpr/DengGVKZ20}.
}

\section{Results} \label{sec:results}

\begin{table*}[!t]
\centering
\resizebox{0.95\linewidth}{!}{%
  \centering
\begin{tabular}{ccc|cccccc|cc}
\multirow{2}{*}{\centering Method} 
& \multirow{2}{*}{\centering Backbone} 
& \multirow{2}{*}{\centering Train data} 
& \multirow{2}{*}{\centering LFW} 
& \multirow{2}{*}{\centering CFP-FP} 
& \multirow{2}{*}{\centering AgeDB30} 
& \multirow{2}{*}{\centering CALFW} 
& \multirow{2}{*}{\centering CPLFW} 
& \multirow{2}{*}{\centering Avg.} 
& \multirow{2}{*}{\centering IJB-B} 
& \multirow{2}{*}{\centering IJB-C}  \\ 
& & & & & & & & & & \\ \midrule 
 
CosFace \cite{DBLP:conf/cvpr/WangWZJGZL018} (CVPR 2018) & ResNet101 & MS1MV2 & 99.81 & 98.12 & 98.11 & 95.76 & 92.28 & 96.82 &  94.80 & 96.37 \\  
ArcFace \cite{DBLP:conf/cvpr/DengGXZ19} (CVPR 2019) & ResNet101 & MS1MV2  & 99.83 & 98.27 & 98.28 & 95.45 & 92.08 & 96.78 & 94.25 &  96.03 \\  
BroadFace \cite{DBLP:conf/eccv/KimPS20} (ECCV 2020) & ResNet101 & MS1MV2  & \textbf{99.85} & 98.63 & \textbf{98.38} & \textbf{96.20} & 93.17 & \textbf{97.25} & 94.97 & 96.38 \\  
CurricularFace \cite{DBLP:conf/cvpr/HuangWT0SLLH20} (CVPR 2020) & ResNet101 & MS1MV2  & 99.80 & 98.37 & 98.32 & \textbf{96.20}  & 93.13 & 97.16 & 94.80 &  96.10 \\  
URL \cite{DBLP:conf/cvpr/Shi0SC020} (CVPR 2020) & ResNet101 & MS1MV2  & 99.78 & 98.64 & - & - & - & - & - & 96.60 \\  
MagFace \cite{DBLP:conf/cvpr/MengZH021} (CVPR 2021) & ResNet101 & MS1MV2  & 99.83 & 98.46 & 98.17 & 96.15 & 92.87 &  97.10 & 94.51 & 95.97 \\  
AdaFace \cite{DBLP:conf/cvpr/Kim0L22} (CVPR 2022) & ResNet101 & MS1MV2  & 99.82 & 98.49 & 98.05 & 96.08 & \textbf{93.53} & 97.19 & \textbf{95.67} &  \textbf{96.89} \\  
ElasticFace-Cos+ \cite{DBLP:conf/cvpr/BoutrosDKK22} (CVPRW 2022) & ResNet101 & MS1MV2  & 99.80 & \textbf{98.73} & 98.28 & 96.18 & 93.23 & - & 95.43 & 96.65 \\  
\hdashline
Face Transformer\cite{DBLP:journals/corr/abs-2103-14803} (2021) & ViT-P12S8  & MS1MV2   & 99.80 & 96.77 & 98.05 & 96.18 & 93.08 & 96.77 & - & 96.31 \\  
TransFace \cite{DBLP:conf/iccv/DanLXD0XS23} (ICCV 2023) & TransFace-B & MS1MV2   & 99.85* & \textbf{99.17*} & \textbf{98.53*} & \textbf{96.20*} & 92.92* & 97.33* & 95.01* & 96.55 \\  
SwinFace \cite{DBLP:journals/tcsv/QinWDWCHD24} (TCSVT 2024) & Swin-T & MS1MV2  & \textbf{99.87} & 98.60 & 98.15 & 96.10 & 93.42 & 97.22 & - & 96.73 \\  
FRoundation \cite{DBLP:journals/ivc/ChettaouiDB25} (IMAVIS 2025) & CLIP ViT-B & MS1MV2   & 99.43 & 93.51 & 92.02 & 93.37 & 90.40 & 93.75 & 82.39 & 86.31 \\  
TransFace++-B \cite{11184862} (TPAMI 2025) & ViT-B & MS1MV2   & 99.85 & 98.53 & 98.34 & - & - & - & 95.45 &  96.67 \\  
ViT* & ViT-B & MS1MV2  & 99.82	&98.94	&97.78	&96.00	&94.27	&97.36	&95.53	&96.77	\\  
ViT-8R (our) & ViT-B + 8 Registers  & MS1MV2  & 99.82	&98.86&	98.00&	\textbf{96.20}&	\textbf{94.65}&	\textbf{97.50}&	\textbf{95.66}&	\textbf{96.88} \\  \hline 

VPL-ArcFace \cite{DBLP:conf/cvpr/DengGYLZ21} (CVPR 2021) & ResNet101 & MS1MV3  & \textbf{99.83} & \textbf{99.11} & \textbf{98.60} & \textbf{96.12} & 93.45 & \textbf{97.42} & 95.56 & 96.76 \\  
AdaFace \cite{DBLP:conf/cvpr/Kim0L22} (CVPR 2022) & ResNet101 & MS1MV3  & \textbf{99.83} & 99.03 & 98.17 & 96.02 & \textbf{93.93} & 97.40 & \textbf{95.84} & \textbf{97.09} \\  
\hdashline
Part-based FR \cite{DBLP:conf/bmvc/SunT22} (BMVC 2022) & Part fViT-B & MS1MV3   & \textbf{99.83} & \textbf{99.21} & 98.29 & - & - & - & \textbf{96.11} & 97.29 \\  
KP-RPE \cite{DBLP:conf/cvpr/KimS0JL24} (CVPR 2024) & ViT-B+KP-RPE & MS1MV3  & - & 99.11 & 97.98 & - & - & - & - & 97.16 \\  
ViT*  & ViT-B & MS1MV3  &  99.78&	99.11	&98.25&	96.12	&94.63	&97.58	&96.05&	97.21 \\  
ViT-8R (our) & ViT-B + 8 Registers & MS1MV3  &  \textbf{99.83}	& 99.17	&\textbf{98.35}	&\textbf{96.18}&	\textbf{94.68}	&\textbf{97.64}&	\textbf{96.11} &	\textbf{97.37} \\  \hline

ArcFace \cite{DBLP:conf/cvpr/DengGXZ19} (CVPR 2019) & ResNet101 & WebFace4M   & \textbf{99.83} & \textbf{99.19} & \textbf{97.95} & 96.00 & 94.35 & 97.46 &  95.75 & 97.16 \\  
AdaFace \cite{DBLP:conf/cvpr/Kim0L22} (CVPR 2022) & ResNet101 & WebFace4M   & 99.80 & 99.17 &  97.90 & \textbf{96.05} & \textbf{94.63}  & \textbf{97.51} & \textbf{96.03} \\  
\hdashline
FRoundation \cite{DBLP:journals/ivc/ChettaouiDB25} (IMAVIS 2025) & CLIP ViT-B  & WebFace4M  &  99.30 & 93.93 & 88.90 & 92.75 & 90.67 & 93.11 & 81.52 & 85.63 \\ 
KP-RPE \cite{DBLP:conf/cvpr/KimS0JL24} (CVPR 2024) & ViT-B+KP-RPE & WebFace4M   & \textbf{99.83}* & 99.01 & \textbf{97.67} & \textbf{96.00}* & \textbf{95.40}* & \textbf{97.58}* & 95.49*  &  97.13 \\  
\rev{KP-RPE*} \cite{DBLP:conf/cvpr/KimS0JL24} (CVPR 2024) & \rev{ViT-B+KP-RPE} & \rev{WebFace4M} &  \rev{99.77} & \rev{98.94} & \rev{97.35} & \rev{95.83} & \rev{95.16} & \rev{97.41} & \rev{95.16} & \rev{96.85} \\ 
\rev{KP-RPE-8R*}  & \rev{ViT-B+KP-RPE + 8 Registers} & \rev{WebFace4M} & \rev{99.85} & \rev{99.03} & \rev{97.47} & \rev{96.00} & \rev{95.08} & \rev{97.49} & \rev{95.44} & \rev{96.96}  \\ 
ViT*  & ViT-B & WebFace4M  &  99,73	&98,97	&97,65&	95,92&	94,97	&97,45& 95.31  & 96.96 \\  
ViT-8R (our) & ViT-B + 8 Registers & WebFace4M &  99,80 &	\textbf{99,03}&	97,40	&95,92&	94,93&	97,42	& \textbf{95,64} & \textbf{97.18} \\   
\end{tabular}}
\caption{Comparison of the ViT-8R (CPE-based ViT with 8 registers) FR performance in comparison to SoTA ViT-based FR solutions. Based on the different training datasets (as reported by these works), the comparison is made under the dashed line in each section (the best performance is in bold). Results on IJB-B and IJB-C are reported as TAR (@FAR=$1e-4$). FR solutions based on non-ViT architectures (ResNet) are provided to put the ViT performance in scope. The results marked with  \textbf{*} were not reported in the original papers and have been reproduced in the paper using the officially released models.} 
\label{tab:sota_table}
\vspace{-5mm}
\end{table*}

\begin{table}[!tp]
\centering
\resizebox{1\linewidth}{!}{%
\begin{tabular}{c|cccccc|c|c}
\multirow{2}{*}{\centering Method} 
& \multirow{2}{*}{\centering LFW} 
& \multirow{2}{*}{\centering CFP-FP} 
& \multirow{2}{*}{\centering AgeDB30} 
& \multirow{2}{*}{\centering CALFW} 
& \multirow{2}{*}{\centering CPLFW} 
& \multirow{2}{*}{\centering Avg.} 
& \multirow{2}{*}{\centering IJBB} 
& \multirow{2}{*}{\centering IJBC} \\ 
& & & & & & & & \\ \midrule 
\texttt{CLS} & 99,77 &	97,67 & 97,83	& 96,05	& 93,50	& 96,96	& 94,75 & 96,03 \\ 
CPE & 99,82 & 98,94 & 97,78 & 96,00 & 94,27 & \textbf{97,36} & \textbf{95,53} & \textbf{96,77} \\ 
\end{tabular}}
\caption{Verification performance of ViT-B models using the \texttt{CLS}-based and CPE-based approaches across multiple benchmarks. On IJB-B and IJB-C, the results are reported as TAR at FAR of 1e-4. The CPE-based solution outperforms the \texttt{CLS}-based approach across most of the benchmarks.} 
\label{tab:cls_cpe}
\vspace{-6mm}
\end{table}

\subsection{Does CPE-based ViT for FR Suffer from Artifacts?} \label{sec:resultscpe_attention}

In this section, we investigate whether the attention maps produced by CPE-based ViT exhibit any artifacts. To this end, we train ViT-L, ViT-B, ViT-S, and ViT-T models on the MS1MV2 dataset and visualize their attention maps using the setting described in Section \ref{sec:experimentalsetup}. From Figure \ref{fig:attention:cpe_vitb_vits}, which illustrates the attention maps of the considered backbones, we draw the following observations:

\textbf{CPE-based ViT for FR exhibits background artifacts.}
All considered, CPE-based backbones suffer from artifacts appearing in the corners of attention maps. 
Given that training and testing images are aligned with similarity transformation with fixed landmark points (Section \ref{sec:experimentalsetup}), 
artifacts are localized in patches corresponding to the background of the face image, which, by comparing the corresponding pixel-level patches, contain less identity-related information compared to the central patches that capture key facial features. As stated in \cite{DBLP:conf/iclr/DarcetOMB24}, artifacts tend to appear on patches that are very similar to their neighbors, indicating redundant information that can be discarded without degrading the quality of the image representation. This observation aligns with our finding that artifacts emerge primarily in regions corresponding to the background at the pixel-level area with low identity relevance. Previous works \cite{DBLP:conf/iclr/DarcetOMB24} utilized ViT for general image classification tasks, reported artifacts in large models, Base, Large, and Huge, while a smaller architecture (ViT-S) exhibits fewer artifacts \cite{DBLP:conf/iclr/DarcetOMB24}. 
Our findings are aligned with previous works \cite{DBLP:conf/iclr/DarcetOMB24} for large models (Vit-L and ViT-B). However, our analysis reveals that small architectures (ViT-S and ViT-T), designed and trained for FR, followed the same trends as large architectures and suffer from similar artifacts in the attention maps. While the \texttt{CLS}-based approach does not exhibit such artifacts, it achieves lower recognition performance compared to CPE\rev{, as shown in Figure \ref{fig:cls_attention} and} reported in Table \ref{tab:cls_cpe}. These observations motivate the need to mitigate artifacts in CPE-based models to improve both interpretability and performance. In the next section, we examine the effect of adding registers to CPE-based ViT as a strategy to minimize these artifacts \cite{DBLP:conf/iclr/DarcetOMB24}.

\vspace{-1mm}
\subsection{Influence of Register on CPE-based ViT for FR} \label{sec:resultsregisters}
\vspace{-1mm}

In this section, we investigate the effect of adding registers to CPE-based ViT to mitigate artifacts observed in attention maps. To this end, we train ViT-B and ViT-S models with 0 (baseline), 1, 2, 4, 8, and 16 registers on the MS1MV2 dataset. From the attention maps, \rev{averaged across five benchmarks (see Section \ref{sec:experimentalsetup})}, visualized in Figure \ref{fig:vit_b_attention_maps}, and from the results summarized in Table \ref{tab:story_table} across a wide range of benchmarks, we make the following observations:

\textbf{Artifacts could be mitigated with four or eight registers in CPE-based ViT.} The attention maps (Figure \ref{fig:vit_b_attention_maps}- b, c, d, e, f and g) and corresponding interpolated ones (Figure \ref{fig:vit_b_attention_maps}- i, j, k, l, m and n) of CPE-based ViT-B with varying numbers of registers are illustrated in Figure \ref{fig:vit_b_attention_maps}, respectively.  
It can be visually observed a clear evolution in focus and artifact suppression by introducing 4 or 8 registers, in comparison to the baseline (without registers). In the baseline model with 0 registers (sub-figures (b, i)), the attention is diffuse and exhibits noticeable artifacts, particularly in peripheral regions of the face. With 1 and 2 registers (sub-figures (c, j) and (d, k)), artifacts are still clearly visible, and interpretability, to a large extent, is limited, especially when comparing the patch-level attention maps with respect to pixel-level areas. A notable improvement occurs at 4 registers (sub-figures (e, l)), where the attention maps become more structured, with clear focus on regions corresponding to pixel-level central facial features, such as the patches corresponding to the areas around the eyes and mouth, and peripheral artifacts are largely mitigated. Increasing to 8 registers (sub-figures (f, m)) further enhances this localization, producing sharper attention on discriminative features, while at 16 registers (sub-figures (g, n)), the maps show slight saturation, indicating diminishing returns from adding more registers.\rev{To validate these observations, Table \ref{tab:maws} reports MAWS and the average attention in regions around key facial landmarks for the 4-patch variant, as defined in Section \ref{sec:experimentalsetup}, across model variants with 0 (baseline), 1, 2, 4, 8, and 16 registers. We first observe that adding registers consistently increases both MAWS and average attention across all five facial landmark regions compared to the baseline, indicating that attention becomes more concentrated around the landmarks. Among the variants, the 4- and 8-register configurations achieve the highest MAWS and average attention, suggesting the strongest landmark-focused attention. Increasing the number of registers to 16 reduces MAWS, indicating that attention becomes more diffuse compared to the 8-register configuration. These results validate the patterns observed in the attention maps.} 
\rev{The observations from Figure \ref{fig:vit_b_attention_maps}, together with the quantitative results in Table \ref{tab:maws}}, demonstrate that adding registers helps the model mitigate artifacts in regions corresponding to background in pixel-level patches while emphasizing identity-relevant facial features. 

\begin{table}[!t]
\centering
\resizebox{1\linewidth}{!}{%
\begin{tabular}{c|cccccc|c}

\multirow{2}{*}{Method} & \multicolumn{6}{|c|}{\rule{0pt}{2.0ex}Landmark 4-Patch Region} & \multirow{2}{*}{MAWS} \\ \cline{2-7}
& Left Eye & Right Eye & Nose & Left Mouth & Right Mouth & Average & \\ \midrule

\rev{CPE} & \rev{0.0056} & \rev{0.0062} & \rev{0.0094} & \rev{0.0062} & \rev{0.0061} & \rev{0.0067} & \rev{0.8086} \\
\rev{CPE + 1-Reg} & \rev{0.0092} & \rev{0.0101} & \rev{0.0072} & \rev{0.0084} & \rev{0.0096} & \rev{0.0089} & \rev{0.8197} \\
\rev{CPE + 2-Reg} & \rev{0.0054} & \rev{0.0052} & \rev{0.0056} & \rev{0.0125} & \rev{0.0108} & \rev{0.0079} & \rev{0.8160} \\
\rev{CPE + 4-Reg} & \rev{0.0188} & \rev{0.0189} & \rev{0.0042} & \rev{0.0098} & \rev{0.0111} & \textbf{\rev{0.0126}} & \rev{0.8492} \\
\rev{CPE + 8-Reg} & \rev{0.0159} & \rev{0.0162} & \rev{0.0123} & \rev{0.0086} & \rev{0.0102} & \textbf{\rev{0.0126}} & \textbf{\rev{0.8523}} \\
\rev{CPE + 16-Reg} & \rev{0.0133} & \rev{0.0112} & \rev{0.0103} & \rev{0.0054} & \rev{0.0058} & \rev{0.0092} & \rev{0.8471} \\

\end{tabular}}
\vspace{-1mm}
\caption{\rev{MAWS and Average attention in regions around key facial landmarks for the 4-patch variant across models with 0, 1, 2, 4, 8, and 16 registers. We observe that both average attention and MAWS are highest for the 4- and 8-register variants, showing that attention is more focused on the facial landmarks in these configurations.}}
\label{tab:maws}
\vspace{-5mm}
\end{table} 

\textbf{Impact of registers on the CPE-based ViT verification accuracies.}
Table \ref{tab:story_table} presents FR verification accuracies of ViT w/o and w/ registers on face benchmarks described in Section \ref{sec:experimentalsetup}. On small evaluation benchmarks LFW, CFP-FP, AgeDB30, CALFW, and CPLFW, ViT-S and ViT-B with registers (1, 2, 4, 8, or 16) achieved competitive results, and even superior ones on most of the considered settings, to the baseline models while providing clear attention maps, especially using 4 or 8 registers. On the large, challenging benchmarks, IJB-B and IJB-C, ViT with 8 registers outperformed the baseline models and the ones with 1, 2, 4, or 16 achieved in general competitive performances to the baseline. Specifically, on IJB-B, the baseline CPE achieves 95.53\% at FAR 1e-4 and 87.72\% at FAR 1e-5, while the 8-register variant improves these to 95.66\% and 88.86\%, respectively. On IJB-C, the baseline reaches 96.77\% at FAR 1e-4 and 93.32\% at FAR 1e-5, compared to 96.88\% and 94.08\% for the 8-register model. Increasing to 16 registers does not yield further gains, with slightly lower performance than the baseline on large-scale benchmarks such as IJB-B and IJB-C, indicating that eight registers provide the optimal balance for enhancing attention and discriminative capability. A similar trend can be observed for the ViT-S variant.

\textbf{\rev{Impact of Concatenating Register Tokens with Patch Tokens.}}
\rev{
To evaluate whether the information captured by the register tokens is beneficial for the downstream task, we conduct an experiment in which the register tokens are concatenated with the output patch tokens before being fed to the MLP to construct the final feature representation. The verification performance of ViT-S with 8 registers with register tokens concatenated across multiple benchmarks is reported in Table \ref{tab:story_table}. This variant achieves performance comparable to the ViT-S with 8 registers evaluated without concatenation, with neither approach consistently outperforming the other.
}

\textbf{\rev{Impact of the Loss Function.}}
\rev{
To assess the sensitivity of our approach to the choice of loss function, we additionally report results with alternative SOTA losses, namely ArcFace \cite{DBLP:conf/cvpr/DengGXZ19} and AdaFace \cite{DBLP:conf/cvpr/Kim0L22}. Their performance is evaluated across multiple benchmarks, presented in Table \ref{tab:loss}. We observe that the model trained with the CosFace loss achieves the highest accuracy on both large-scale benchmarks, IJB-B and IJB-C.
}

\begin{table}[!t]
\centering
\resizebox{1\linewidth}{!}{%
\begin{tabular}{c|cccccc|c|c}
\multirow{2}{*}{\centering Method} 
& \multirow{2}{*}{\centering LFW} 
& \multirow{2}{*}{\centering CFP-FP} 
& \multirow{2}{*}{\centering AgeDB30} 
& \multirow{2}{*}{\centering CALFW} 
& \multirow{2}{*}{\centering CPLFW} 
& \multirow{2}{*}{\centering Avg.} 
& \multirow{2}{*}{\centering IJB-B} 
& \multirow{2}{*}{\centering IJB-C} \\ 
& & & & & & & &  \\ \midrule 
CosFace \cite{DBLP:conf/cvpr/WangWZJGZL018} & 99.80 & 99.03 & 97.40	& 95.92 & 94.93 & 97.42	& \textbf{95.64} & \textbf{97.18} \\ 
\rev{ArcFace} \cite{DBLP:conf/cvpr/DengGXZ19} & \rev{99.77} & \rev{99.04} & \rev{97.58} & \rev{95.98} & \rev{95.03} & \textbf{\rev{97.48}} & \rev{95.57} & \rev{97.13} \\ 
\rev{AdaFace} \cite{DBLP:conf/cvpr/Kim0L22} & \rev{99.77} & \rev{98.97} & \rev{97.42} & \rev{95.93} & \rev{95.12} & \rev{97.44} & \rev{95.43} & \rev{97.08} \\ 
\end{tabular}}
\vspace{-1mm}
\caption{\rev{Verification performance of CPE-based ViT-B + 8 Registers models using different loss functions across multiple evaluation benchmarks. On IJB-B and IJB-C, the results are reported as TAR at FAR of 1e-4. CosFace training yields the highest accuracy on large-scale benchmarks.}}
\label{tab:loss}
\vspace{-5mm}
\end{table}

\vspace{-1mm}
\textbf{Computation Analysis.}
Table \ref{tab:story_table} presents a comparison of computational resources for CPE-based ViT-B and ViT-S with 0 (baseline), 1, 2, 4, 8, and 16 registers, in terms of FLOPs and parameter count. The CPE-based methods concatenate all patch embeddings \( \mathbf{z}_L^i \in \mathbb{R}^D \) for \( i = 1, \dots, N \), which are then projected to a final feature vector. While the register tokens are discarded at the output and not concatenated with the patch embeddings, they pass through all transformer layers, contributing to additional computational load. For instance, the variant with 8 registers requires an additional 1.33 GFLOPs compared to the baseline.

\begin{figure}[!t]
  \centering
  \begin{minipage}{\linewidth}
      \centering
        \begin{subfigure}[b]{0.18\linewidth}
          \includegraphics[width=\linewidth, height=1.6cm]{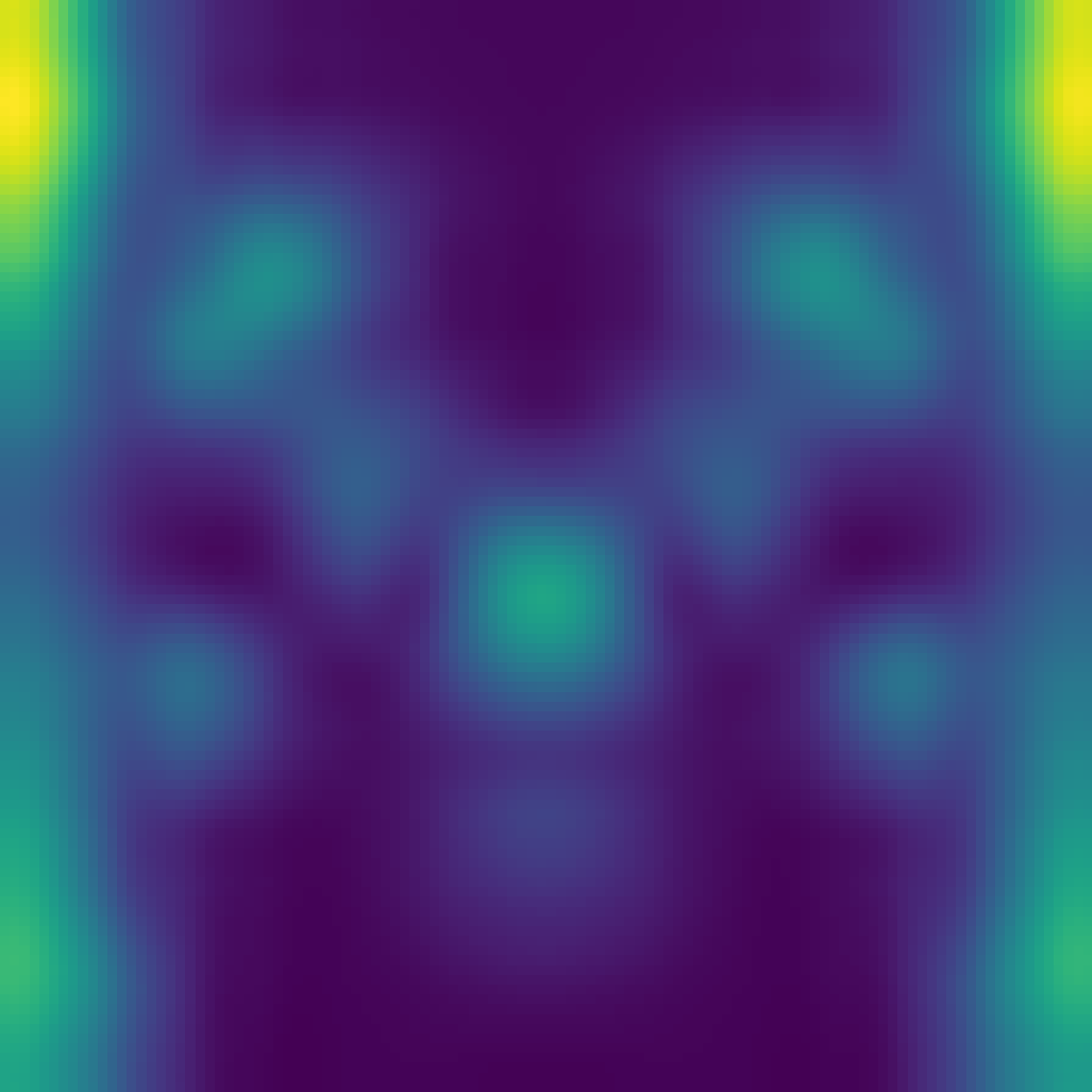}
          \caption{Sample}
      \end{subfigure}
      \hfill
     \begin{subfigure}[b]{0.18\linewidth}
          \includegraphics[width=\linewidth, height=1.6cm]{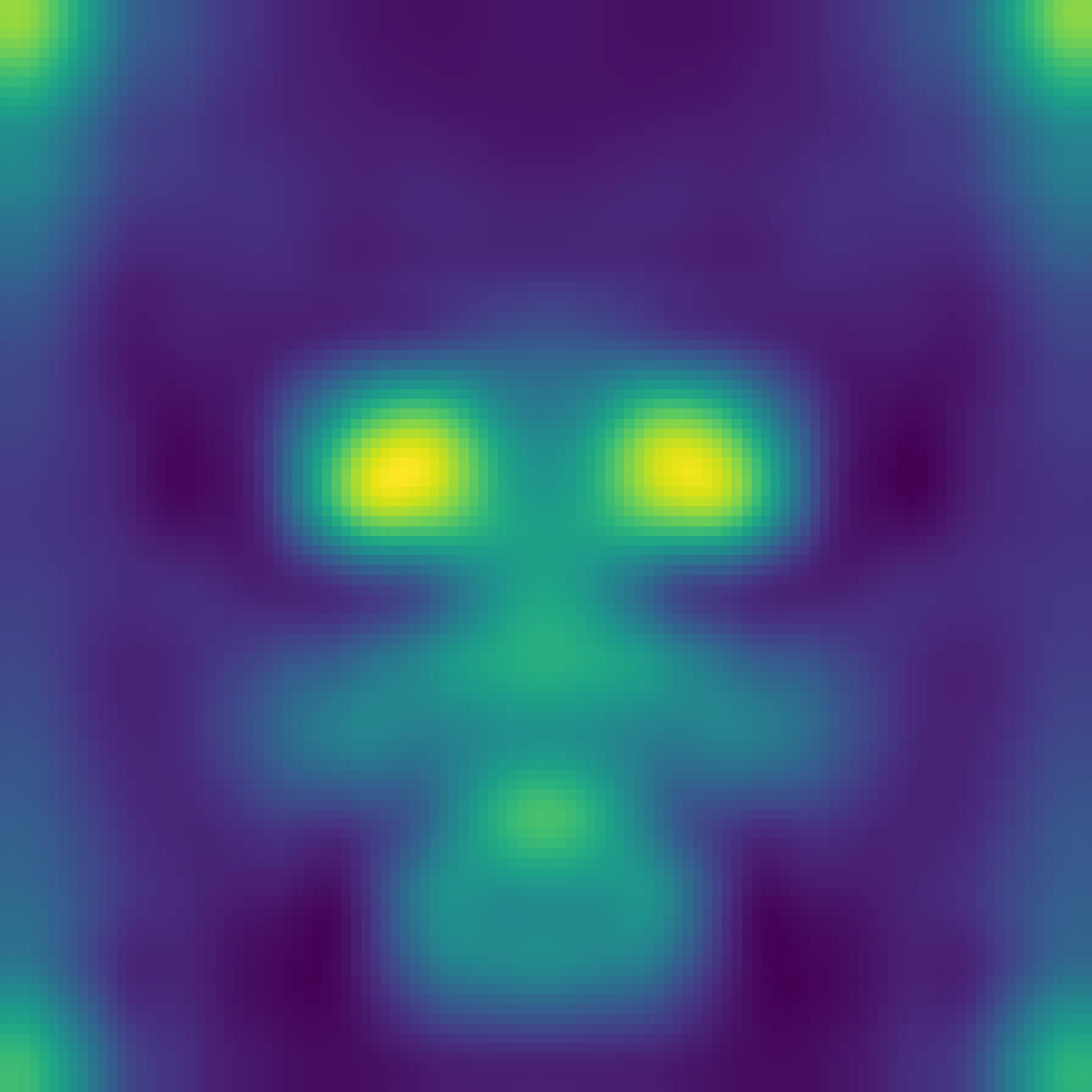}
          \caption{ViT-L}
      \end{subfigure}
      \hfill
      \begin{subfigure}[b]{0.18\linewidth}
          \includegraphics[width=\linewidth, height=1.6cm]{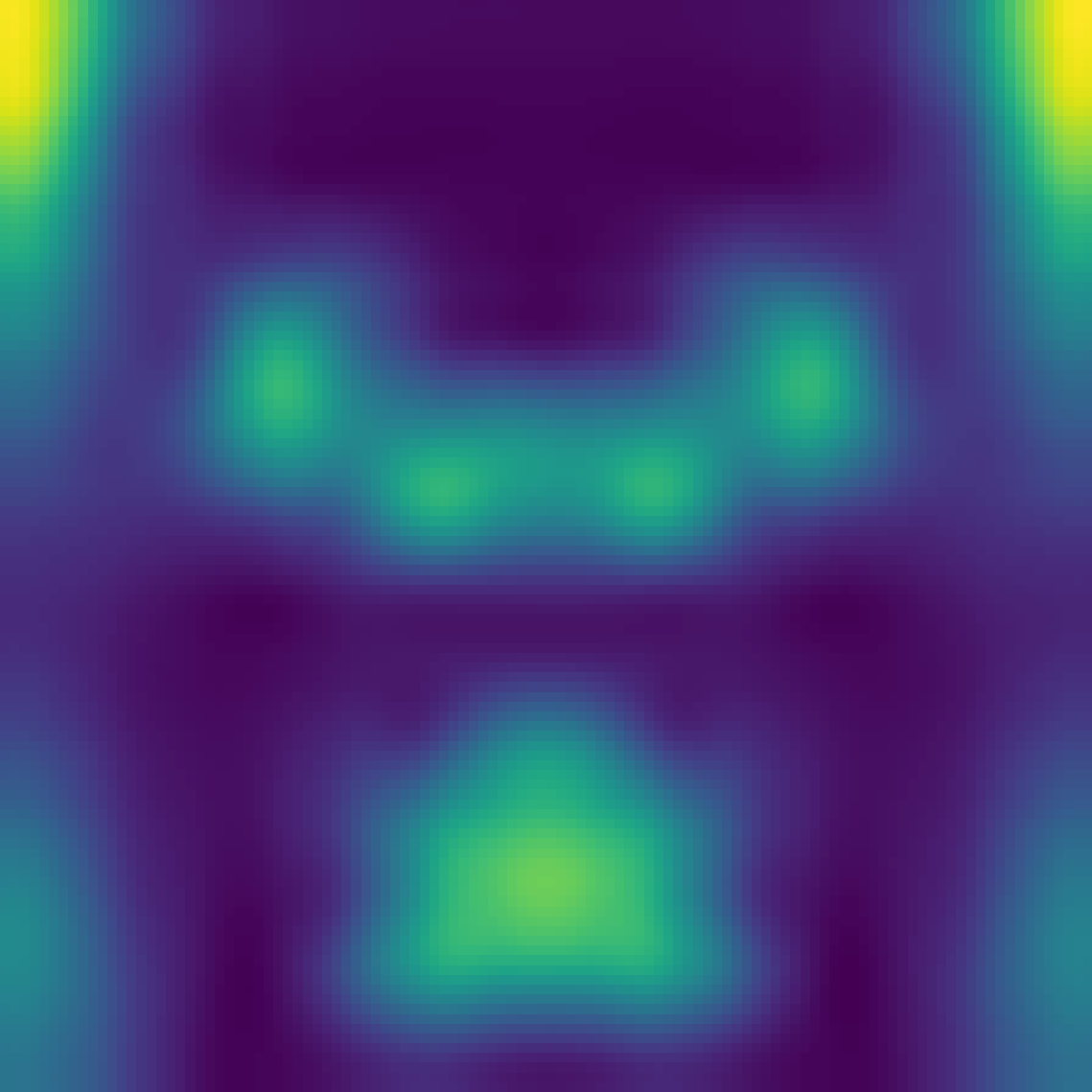}
          \caption{ViT-B}
      \end{subfigure}
      \hfill
      \begin{subfigure}[b]{0.18\linewidth}
          \includegraphics[width=\linewidth, height=1.6cm]{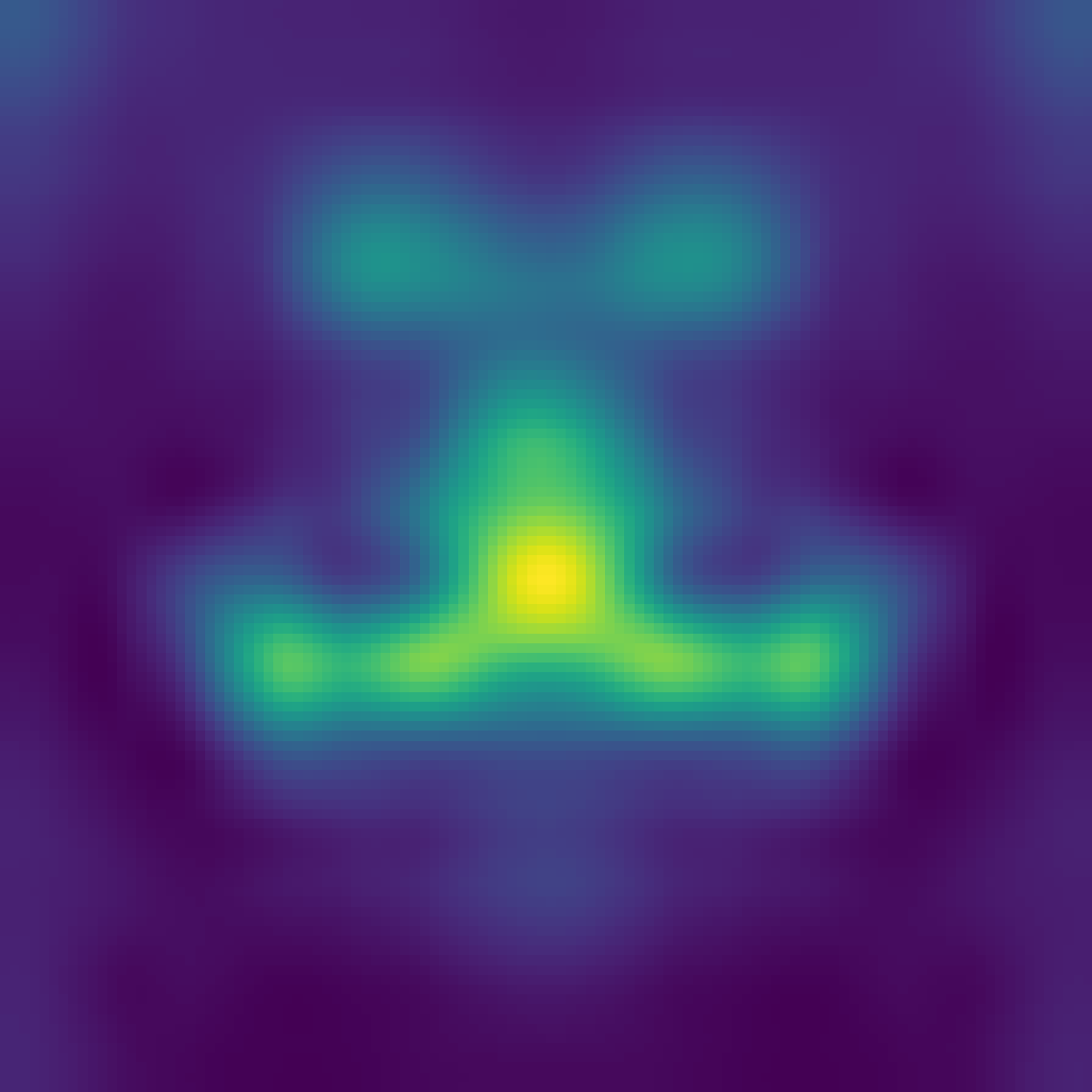}
          \caption{ViT-S}
      \end{subfigure}
      \hfill
        \begin{subfigure}[b]{0.18\linewidth}
          \includegraphics[width=\linewidth, height=1.6cm]{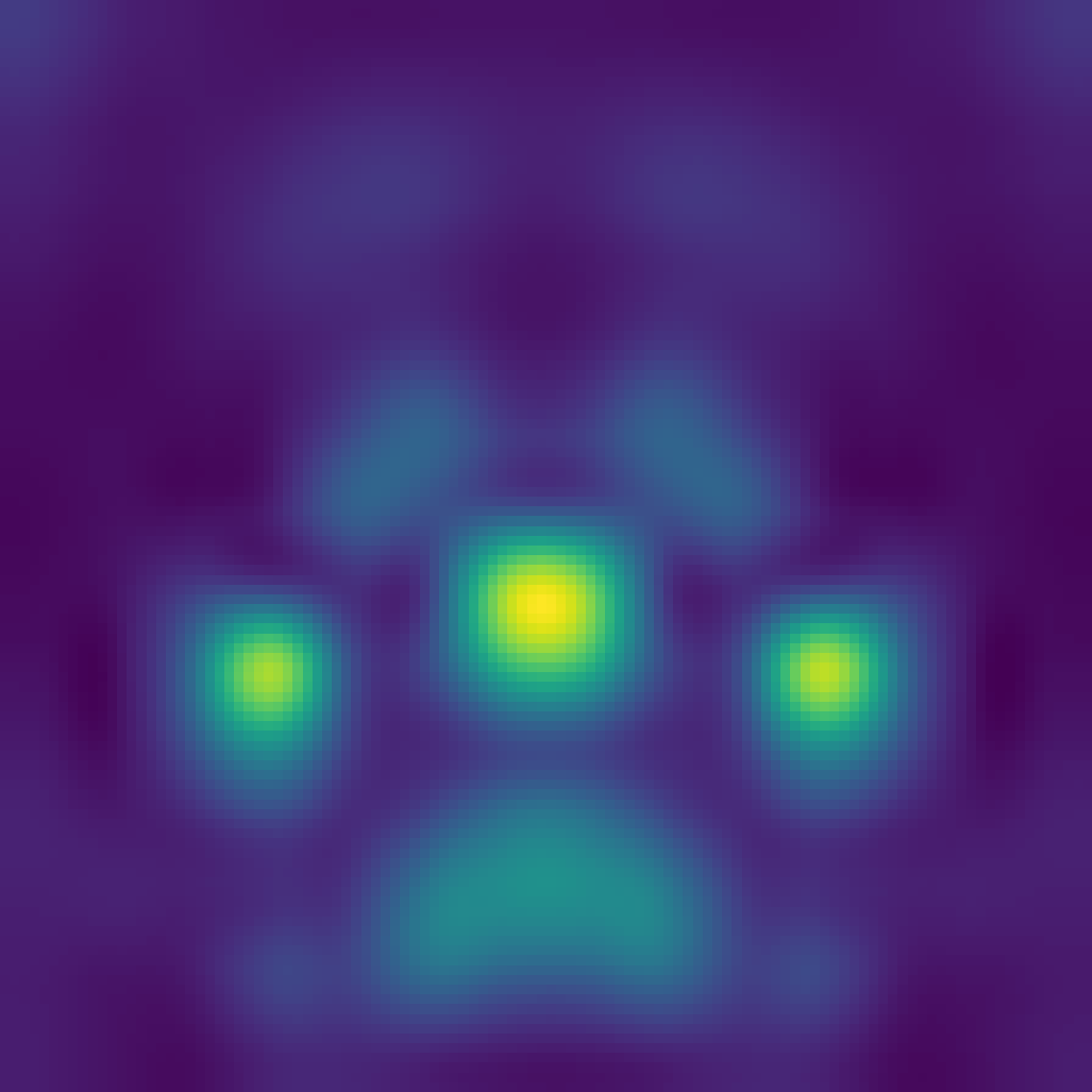}
          \caption{ViT-T}
      \end{subfigure}
  \end{minipage}
  \vspace{-1mm}
  \caption{\rev{Attention maps of KP-RPE \cite{DBLP:conf/cvpr/KimS0JL24}, averaged across five benchmarks (see Section 4), comparing models trained with 0, 1, 2, 4, and 8 registers. Similarly to ViT, we observe that artifacts are reduced once 4 or 8 registers are added.}}
  \label{fig:attn_kprpe}
\vspace{-5mm}
\end{figure}

\textbf{\rev{ViT KP-RPE with Registers.}}
\rev{In addition to enhancing ViT with registers, we also explore ViT KP-RPE \cite{DBLP:conf/cvpr/KimS0JL24} enhanced with registers to evaluate their joint effect. Table \ref{tab:kprpe} presents FR verification accuracies of ViT KP-RPE with 0, 1, 2, 4, and 8 registers on face benchmarks described in Section 4. The model backbone is a ViT-B trained on WebFace4M \cite{zhu2021webface260mbenchmarkunveilingpower}. All models (including the variant without registers) were trained using the officially released code and the same experimental setup as in \cite{DBLP:conf/cvpr/KimS0JL24} for large-scale experiments. On small evaluation benchmarks LFW, CFP-FP, AgeDB30, CALFW, and CPLFW, KP-RPE with registers achieved competitive results, and even superior ones on most of the considered settings, to the baseline models while providing clearer attention maps, especially using 4 or 8 registers, as presented in Figure \ref{fig:attn_kprpe}. On the large and challenging IJB-B benchmark, ViT KP-RPE with 8 register tokens outperforms all other considered models.
}

\vspace{-1mm}
\begin{table}[!t]
\centering
\resizebox{1\linewidth}{!}{%
\begin{tabular}{c|cccccc|c|c}
\multirow{2}{*}{\centering Method} 
& \multirow{2}{*}{\centering LFW} 
& \multirow{2}{*}{\centering CFP-FP} 
& \multirow{2}{*}{\centering AgeDB30} 
& \multirow{2}{*}{\centering CALFW} 
& \multirow{2}{*}{\centering CPLFW} 
& \multirow{2}{*}{\centering Avg.} 
& \multirow{2}{*}{\centering IJB-B} 
& \multirow{2}{*}{\centering IJB-C} \\ 
& & & & & & & & \\ \midrule 

\rev{KP-RPE} & \rev{99.77} & \rev{98.94} & \rev{97.35} & \rev{95.83} & \rev{95.16} & \rev{97.41} & \rev{95.16} & \rev{96.85} \\ 
\rev{KP-RPE + 1 Reg} & \rev{99.77} & \rev{98.89} & \rev{97.40} & \rev{96.05} & \rev{95.17} & \rev{97.45} & \rev{95.40} & \textbf{\rev{97.01}} \\ 
\rev{KP-RPE + 2 Reg} & \rev{99.77} & \rev{99.09} & \rev{97.43} & \rev{96.07} & \rev{95.37} & \textbf{\rev{97.54}} & \rev{95.39} & \rev{96.94} \\ 
\rev{KP-RPE + 4 Reg} & \rev{99.80} & \rev{98.99} & \rev{97.58} & \rev{95.92} & \rev{95.18} & \rev{97.49} & \rev{95.35} & \rev{96.92} \\ 
\rev{KP-RPE + 8 Reg} & \rev{99.85} & \rev{99.03} & \rev{97.47} & \rev{96.00} & \rev{95.08} & \rev{97.49} & \textbf{\rev{95.44}} & \rev{96.96} \\

\end{tabular}}
\caption{\rev{Verification performance of ViT KP-RPE \cite{DBLP:conf/cvpr/KimS0JL24} models approaches across multiple evaluation benchmarks. On IJB-B and IJB-C, the results are reported as TAR at FAR of 1e-4. The 8-register variant consistently outperforms the baseline on large-scale benchmarks.}} 
\label{tab:kprpe}
\vspace{-4mm}
\end{table}

\vspace{-1mm}
\subsection{Comparison with SoTA Methods} \label{sec:sota}
In this section, we compare our CPE-based ViT model with eight registers (ViT-8R) against SoTA ViT-based FR methods trained on various datasets. While most ViT-based approaches adopt the ViT-B architecture, their configurations often differ in terms of attention heads, patch size, embedding dimension, or transformer depth, which can influence both performance and computational efficiency. For instance, our ViT-B backbone uses 8 attention heads and a patch size of 9, whereas KP-RPE \cite{DBLP:conf/cvpr/KimS0JL24} employs 16 attention heads with a patch size of 8. Similarly, FRoundation \cite{DBLP:journals/ivc/ChettaouiDB25} trains models with an image size of 224 compared to 112 in our setup. To provide a broader context, we also include ResNet-based FR methods in our comparison to highlight the advantages of ViT-8R over traditional convolutional architectures.

As shown in Table \ref{tab:sota_table}, our ViT-8R achieves SOTA results on the large-scale benchmarks IJB-B \cite{inproceedingsijbb} and IJB-C \cite{DBLP:conf/icb/MazeADKMO0NACG18} when trained on datasets such as MS1MV2, MS1MV3, and WebFace4M. When trained on MS1MV2, it also achieves the highest average performance across five small benchmarks designed to evaluate FR under diverse conditions.

\vspace{-2mm}
\section{Conclusion}
\vspace{-2mm}
Our study demonstrates that while CPE-based ViTs are strongly promising for FR, their attention maps consistently exhibit artifacts, particularly in regions corresponding to low identity relevance, such as the background in pixel-level patches. We show that these artifacts appear across backbones of various sizes, including both small and large models, highlighting a limitation in interpretability, which has not been addressed by prior work. To address this limitation, we add register tokens to CPE-based ViTs designed and trained for FR, and demonstrate that it effectively mitigates these artifacts, producing more structured and interpretable attention maps. Our experiments reveal that adding four or eight registers provides substantial improvements in attention interpretability, with eight registers offering the clearest focus on identity-relevant facial features. The resulting ViT-8R model, a CPE-based ViT-B with eight registers, not only enhances interpretability but also achieves superior results on large-scale benchmark compared to baseline (0 registers). Additionally, ViT-8R achieves SoTA performance on large-scale benchmarks such as IJB-B and IJB-C, while maintaining competitive results on smaller FR benchmarks. These findings highlight that register tokens provide a simple yet effective mechanism to improve both the interpretability and performance of ViT-based solutions for FR, improving both discriminative capability and interpretability of the model.

\vspace{-2mm}
\section{Ethical Impact Statement}
\vspace{-2mm}
Our research on register tokens for Vit-based face recognition improves both model performance and interpretability, contributing to more accurate and transparent recognition systems. By enhancing the clarity of attention maps, this work supports a deeper understanding of how CPE-based ViTs process facial information, which can facilitate model analysis, debugging, and the development of more trustworthy AI systems. Improved interpretability can also aid practitioners in identifying model behavior and potential limitations. While clearer attention maps provide valuable insights, we emphasize that interpretability does not necessarily guarantee correctness or fairness, and should be complemented with thorough evaluation. We advocate for responsible deployment within established legal and regulatory frameworks (e.g., GDPR), supported by safeguards such as transparency, user consent, and human oversight in sensitive applications. By advancing both performance and interpretability, this work contributes to the development of more reliable and accountable face recognition technologies.


{\small
\bibliographystyle{ieee}
\bibliography{egbib}
}

\end{document}